\documentclass[12pt]{article}
\usepackage{amsgen,amsthm,amsmath,amstext,amsbsy,amsopn,amssymb,latexsym, bbm}
\usepackage{mathtools}
\usepackage[usenames]{color}
\usepackage{array}
\usepackage{cite}
\usepackage{multirow}
\usepackage{bm}
\usepackage{amssymb}
\usepackage{mathabx}
\usepackage{natbib}
\usepackage{wrapfig}
\usepackage[ruled,vlined,linesnumbered,noresetcount]{algorithm2e}
\usepackage{hyperref}
\usepackage{pifont}
\usepackage{float}
\usepackage{tikz}
\usepackage{setspace}
\usepackage{graphicx} 
\usepackage[font=small,labelfont=bf]{caption} 
\usepackage[subrefformat=parens]{subcaption}
\usepackage{enumerate} 
\usepackage{pgfplots}
\usepackage[normalem]{ulem}
\usepackage{titlesec}
\titlespacing\section{0pt}{1pt plus 2pt minus 2pt}{0pt plus 2pt minus 2pt}
\titlespacing\subsection{0pt}{1pt plus 2pt minus 2pt}{0pt plus 2pt minus 2pt}
\titlespacing\subsubsection{0pt}{1pt plus 2pt minus 2pt}{0pt plus 2pt minus 2pt}

\pgfplotsset{width=10cm,compat=1.9}
\usepackage{amsthm}

\newcommand{\innerproduct}[2]{\langle #1, #2 \rangle}

\newtheorem{Corollary}{Corollary}
\newtheorem{Proposition}{Proposition}
\newtheorem{Lemma}{Lemma}

\newtheorem{Theorem}{Theorem}

\newtheorem{Assumption}{Assumption}

\def\trans{^{\scriptscriptstyle \sf T}}
\def\Asc{\mathcal{A}}

\def\Qbb{\mathbb{Q}}
\def\Pbb{\mathbb{P}}

\newcommand{\BB}{{\mathbf{B}}}

\newcommand{\D}{{\mathbf{D}}}

\newcommand{\PP}{{\mathbf{P}}}

\def\YY{\mathbf{Y}}
\def\XX{\mathbf{X}}

\newcommand{\RR}{\mathbf{R}}
\newcommand{\R}{\mathbb{R}}
\newcommand{\E}{\mathbf{E}}

\newcommand{\Tar}{\mathbb{Q}}
\newcommand{\T}{\mathbb{T}}
\def\subT{_{\scriptscriptstyle \T}}
\def\PE{\mathrm{PE}}

\def\bGamma{\boldsymbol{\Gamma}}
\def\BBhat{\widehat{\BB}}
\def\bGammahat{\widehat{\bGamma}}
\def\betahat{\widehat{\beta}}

\def\Csc{\mathcal{C}}
\def\Ssc{\mathcal{S}}

\def\sublinear{_{\scriptscriptstyle \sf linear}}
\def\subinit{_{\scriptscriptstyle \sf init}}
\def\submaximin{_{\scriptscriptstyle \sf maximin}}

\def\subvalid{_{\scriptscriptstyle \sf valid}}

\def\supl{^{\scriptscriptstyle (l)}}
\def\supone{^{\scriptscriptstyle (1)}}
\def\supL{^{\scriptscriptstyle (L)}}

\def\suba{_{\scriptscriptstyle \mathcal{A}_1}}
\def\subaa{_{\scriptscriptstyle \mathcal{A}_2}}

\def\subaj{_{\scriptscriptstyle \mathcal{A}_j}}
\def\subajc{_{\scriptscriptstyle \mathcal{A}_j^c}}

\def\subsource{_{\mathrm{mix}\mathbb{P}}}
\def\supsource{^{\mathrm{mix}\mathbb{P}}}
\def\supconvex{^{\mathrm{convex}}}
\def\suplinear{^{\mathrm{weighted}}}
\def\supmixST{^{\mathrm{mixST}}}
\def\supmixT0{^{\mathrm{mixT0}}}
\def\supweighted{^{\mathrm{weighted}}}
\def\supconvex{^{\mathrm{convex}}}

\def\subX{_{\scriptscriptstyle \sf X}}
\def\subYX{_{\scriptscriptstyle \sf Y|X}}
\def\supQ{^{\mathbb{Q}}}
\def\subQ{_{\mathbb{Q}}}
\def\subP{_{\mathbb{P}}}
\def\subPl{_{\mathbb{P}\supl}}
\def\supT{\trans}
\def\subGART{_{\mathrm{\scriptscriptstyle GART}}}

\def\subUKBWB{_{\scriptscriptstyle \sf UKB,WB}}
\def\subUKBWA{_{\scriptscriptstyle \sf UKB,WA}}
\def\subUKBMIX{_{\scriptscriptstyle \sf UKB,MIX}}
\def\DRE{\Delta_{\scriptscriptstyle \sf RE}}

\newtheorem{remark}{Remark}

\DeclareMathOperator*{\argmax}{arg\,max}
\DeclareMathOperator*{\argmin}{arg\,min}

\let\OLDthebibliography\thebibliography
\renewcommand\thebibliography[1]{
  \OLDthebibliography{#1}
  \setlength{\parskip}{0pt}
  \setlength{\itemsep}{0pt plus 0.3ex}
}

\newcommand{\blind}{1}
\newcounter{auxFootnote}
\allowdisplaybreaks

\usepackage{xr}

\begin{document}
\def\spacingset#1{\renewcommand{\baselinestretch}%
{#1}\small\normalsize} \spacingset{1}

\expandafter\def\expandafter\normalsize\expandafter{%
    \normalsize%
    \setlength\abovedisplayskip{5pt}%
    \setlength\belowdisplayskip{6pt}%
    \setlength\abovedisplayshortskip{-8pt}%
    \setlength\belowdisplayshortskip{2pt}%
}

\if1\blind
{
  \title{\bf Guided Adversarial Robust Transfer Learning with Source Mixing}
  \author{Xin Xiong$^{1}$, 
  Zijian Guo$^{2}$\footnote{Guo and Cai contributed equally}\setcounter{auxFootnote}{\value{footnote}} \footnote{Corresponding author: zijguo@zju.edu.cn}, 
  Tianxi Cai$^{1,3}$\footnotemark[\value{auxFootnote}] \footnote{Corresponding author: tcai@hsph.harvard.edu}
\bigskip \\ 
$^1${Harvard T.H. Chan School of Public Health, Boston, MA, USA} \\
$^2${Center for Data Science, Zhejiang University, Zhejiang, China} \\
$^3${Harvard Medical School, Boston, MA, USA} \\
\date{}
}
  \maketitle
} \fi

\if0\blind
{
  \bigskip
  \bigskip
  \bigskip
  \begin{center}
    {\LARGE\bf Guided Adversarial Robust Transfer Learning with Source Mixing}
\end{center}
  \medskip
} \fi

\bigskip
\begin{abstract}
Transfer learning is a critical technique that enables the application of knowledge gained from existing tasks or domains to improve performance on a new one, reducing the need for extensive data and training in each new context. Many existing transfer learning methods rely on leveraging information from source populations closely resembling the target population. However, this approach often overlooks valuable knowledge that may be present in different yet potentially related auxiliary samples. When dealing with a limited amount of target data and multiple source data, we introduce a novel approach, Guided Adversarial Robust Transfer  (GART) learning, that breaks free from strict similarity constraints. GART is designed to optimize the most adversarial loss with respect to a collection of source mixture distributions that guarantee excellent prediction performances for the target data. We establish the closed form of the population GART and show that the GART estimator achieves a faster convergence rate than the model fitted with the target data. Our simulation studies suggest that GART outperforms existing transfer learning methods, attaining higher robustness and accuracy. We highlight GART's predictiveness and robustness by applying it to form genetic prediction models of high-density lipoprotein cholesterol using multi-institutional biobank-linked electronic health records data.

\end{abstract}

\noindent%
{\it Keywords:} Multi-Source Data; Distributionally Robust Optimization; Heterogeneous Data; High Dimensionality; Electronic Health Record 
\vfill

\newpage
\spacingset{1.9} 

\section{Introduction}
Transfer learning stands as a pivotal concept in the realm of statistical and machine learning due to its immense importance and transformative potential \citep{torreytransfer}. It allows models trained on existing datasets to leverage their knowledge and expertise when confronted with a new target population. Transfer learning enables models to generalize and adapt their acquired knowledge to novel challenges, making them more versatile and generalizable. Transfer learning has the potential to rapidly develop a generalizable model for new target populations, addressing the scarcity of clinical data  (\citep{desautels2017using, gligic2020named, laparra2021review}). An example of such analysis is to develop a rare-disease mortality prediction model in a target hospital with limited labels, with the aid of abundant EHR data from other large hospitals \citep{desautels2017using}. 

Leveraging multiple data sources to derive a precise prediction model for a new target population is challenging in the presence of heterogeneity among the sources and between the target and source populations. To ensure the effective transfer of knowledge from the source samples and avoid negative transfer, one of the key assumptions for traditional transfer learning models is that the target model and some source models must possess a certain level of similarity. For example, both TransLasso \citep{transLasso} and the TransGLM \citep{transglm} models and some other recent works (\citep{bastani2021predicting, zhang2022transfer, he2022transfer, li2023targeting, li2023transfer}) require the recovery of the set of \textit{transferrable} source sites, which share similar parameters with the target site. The distance between the parameters of these transferrable sources and that of the target needs to be sufficiently small so that introducing source data in their algorithms can help improve the estimation error rate. However, informative sources might be dropped because of the stringent requirements of the transferrable set and the natural heterogeneity of data collected from different sources. Given the potential high heterogeneity in the real data, constructing a prediction model without imposing such between-site similarity conditions would be much preferred.  In this paper, we will show that by relaxing the similarity conditions, our proposed model achieves a much-improved performance in theory and experiments compared to traditional transfer learning models. 

A weaker similarity notation is that a mixture of source distributions approximates the target distribution. Under source-mixture assumptions, group distributional robust optimization (DRO)\citep{sagawa2019distributionally,hu2018does,wang2023DRL} or agnostic federated learning \citep{mohri2019agnostic, deng2020distributionally} have been proposed to attain robust prediction performance for a range of target populations. Under the linear model framework, the maximin effect is a group DRO model with the closed form expression \citep{Meinshausen_2015}\citep{guo2020inference}. Particularly,  the group DRO or the maximin literature leverages $L$ source data sets generated from the possibly heterogeneous distributions $\{\mathbb{P}\supl:=(\mathbb{P}\subX\supl, \mathbb{P}\subYX\supl)\}_{1\leq l \leq L}$ to build a robust prediction model for the target domain, where the target model $\Tar\subYX$ is allowed to differ from any of $\{\mathbb{P}\supl\subYX\}_{1\leq l\leq L}$. Given that the target labels are not available, $\Tar\subYX$ is generally not identifiable. Instead of estimating the true $\Tar\subYX$, group DRO models assume $\Tar\subYX$ is a mixture of $\{\mathbb{P}\supl\subYX\}_{1\leq l\leq L}$ and define an uncertainty set $\mathcal{C}_0$ as any mixture of these sources, 
\begin{equation}
\mathcal{C}_0=\left\{\T=(\Tar\subX, \T\subYX): \T\subYX=\sum_{l=1}^{L} q_l \cdot \mathbb{P}\supl\subYX \; \text{with}\; q\in \Delta_{L} \right\}, 
\label{eq: convex hull}
\end{equation}
where $\Delta_L = \left\{q\in\mathbb{R}^L:\sum_{l=1}^Lq_l=1,\min_lq_l\geq 0\right\}$ is the $L$-dimensional simplex. Given a model family $\Theta$ and a loss function $l: \Theta\times (\mathcal{X}\times\mathcal{Y})\to \mathbb{R}_+$, group DRO minimizes the worst-case expected loss over the uncertainty set $\mathcal{C}_0$, that is 
$\min_{\theta\in\Theta} \left\{\sup_{\mathbb{T}\in\mathcal{C}_0} \E_{(X,Y)\sim\mathbb{T}}\left[l(\theta;X, Y)\right] \right\}.$

Group DRO enhances the prediction model's generalizability by choosing a general family $\mathcal{C}_0$ that encodes target distributions with possible distributional shifts from the source distributions. However, group DRO attains the generalizability at the cost of potentially limited performance for any specific target population due to the large uncertainty set $\mathcal{C}_0$. 
In this paper, we shall leverage the observations on a target population $\Tar$ to guide the group DRO to attain higher accuracy for $\Tar$ while maintaining generalizability to future target populations similar to $\Tar$. 

{Another line of multi-task learning research \citep{duan2023adaptive,tian2023learning,tiantowards} analyzes heterogeneous multi-source data, but differs from our setting by aiming to jointly estimate $L$ source-specific models $\Theta \in \mathbb{R}^{p \times L}$, capturing shared and unique components to enhance performance at each site.}

\subsection{Our results and contributions}
This paper considers the high-dimensional transfer learning regime, where we have access to a small amount of target data and abundant yet distinct source data. Inspired by the design of group DRO, the main idea of GART is to incorporate the distributional uncertainty of the future target data into the optimization process. The guidance of the target data is reflected in the construction of the uncertainty class and the loss function. Particularly, we construct a transferable uncertainty class containing a class of conditional outcome distributions which are generated as a mixture of $\Tar\subYX$ and $\{\mathbb{P}\supl\subYX\}_{1\leq l\leq L}$ and achieving excellent prediction accuracy on the observed target data. Such a transferable uncertainty class makes full use of target labels and guides the GART effect toward the interested target problem. To define the loss function, we construct an initial model leveraging the available source and target data. Then we define a new loss function of any prediction model as the difference between its least squares prediction loss and the initial model's prediction loss. We finally define the GART prediction model as the minimizer of the worst-case loss with respect to this transferable uncertainty class. Although GART is defined as a distributionally robust prediction model, the construction of the uncertainty class and loss functions are both novel. Such a careful integration of the target data into constructing the uncertainty set and the loss function enables our GART prediction model to enjoy the benefits of DRO and transfer learning simultaneously.   

GART has three key advantages compared to existing transfer learning methods. Firstly, GART typically has better predictive performance than the existing group DRO or maximin model since, instead of focusing on the whole source mixture, we focus on a small uncertainty class near the observed target. Second, GART enables broader knowledge transfer via source mixing, overcoming the limitation of selecting only target-similar sources. As shown in Proposition \ref{prop: relationship between alpha and h}, source mixing imposes a weaker similarity condition than required in prior transfer learning work \citep{transLasso, transglm}. Lastly, GART is robustness to violation of the source mixing assumption. When the target outcome model is away from the source mixtures (i.e., the source mixing assumption fails), GART is reduced to the model fitted with only target data, which avoids negative transfer issues. When the source mixing assumption holds, we demonstrate that a large amount of the source data can significantly improve the convergence rate of the GART estimator in comparison to the target-only estimator. 

Since GART incorporates labeled target data to attain robust prediction for future target populations, it also relates to the DRO literature \citep{ben2013robust,namkoong2016stochastic,namkoong2017variance, sinha2017certifying,blanchet2019robust,blanchet2019quantifying,gao2023distributionally,DRO_MU}. The DRO method attains robustness by minimizing the worst performance within an uncertainty set defined by the neighborhood of the observed target populations according to specific distance metrics. Under some special configurations, the DRO methods are equivalent to regularized estimation \citep{shafieezadeh2019regularization}. Our proposed GART can also be viewed as a generalization of the DRO methods by further leveraging sources. 

\subsection{Notations}
We use $c$ and $C$ to denote generic positive constants that may vary from place to place. For a vector $x$, we define its $\ell_q$ norm as $\|x\|_{q}=\left(\sum_{l=1}^{p}|x_l|^q\right)^{\frac{1}{q}}$ for $q \geq 0$. For two positive sequences $a_n$ and $b_n$, we denote $a_n \ll b_n$ if $\limsup_{n\rightarrow\infty} {a_n}/{b_n}=0$, and $a_n \lesssim b_n$ if there exists a universal constant $C$ irrelevant to $n$ or any other parameters such that $a_n \leq Cb_n$ for all $n$. For a vector $x\in\mathbb{R}^p$, a matrix $\XX$, and a subset $\mathcal{A}\subset\{1,...,n\}$, $x_{\mathcal{A}}$ is the sub-vector of $x$ with indices in $\mathcal{A}$ and $\XX_{\mathcal{A}}$ denotes the sub-matrix of $\XX$ with row indices belonging to $\mathcal{A}$. For $x\in\mathbb{R}$, $\lceil x \rceil$ denotes the smallest integer that is larger than $x$. We use $\mathbf{1}_{p}$ to denote a $p$-dimension all-ones vector, and $\mathbf{0}_{p}$ a $p$-dimension all-zeros vector, and $\mathbb{I}(\cdot)$ a 0-1 indicator function.
\section{GART: model definition and identification}
We consider training a generalizable prediction model for $Y \mid X$ using a small set of target data and $L$ sets of heterogeneous source data. The $l$-th source training data with $1 \le l \le L$,  $\{X\supl_{i}, Y\supl_{i}\}_{1\leq i\leq N_l}$  are independently generated from 
$\{\mathbb{P}\supl:=(\mathbb{P}\subX\supl, \mathbb{P}\subYX\supl)\}_{1\leq l \leq L}$, where $\mathbb{P}\subX\supl$ and $\mathbb{P}\supl\subYX$ are distributions of $X\supl$ and $Y\supl \mid X\supl$. The target data, $\{X\supQ_{i},Y\supQ_{i}\}_{1\leq i\leq n}$, are generated independently from $\mathbb{Q}:=(\mathbb{Q}\subX, \mathbb{Q}\subYX)$, where $\mathbb{Q}\subX$ and $\mathbb{Q}\subYX$ are distributions of $X\supQ_i$ and $Y\supQ_i \mid X\supQ_i$.
We focus on the regime that the target sample size $n$ is relatively small compared to $\{N_l\}_{l=1}^L$. Although the GART framework can be applied to a broad range of models, we focus on the linear models for $\Pbb\subYX\supl$ as well as the target $\Qbb\subYX$ with
\begin{alignat}{2}
Y_i\supl&=\left(X_i\supl\right)\trans b\supl+\epsilon\supl_i &\quad & \text{where  }\E\left(\epsilon_i\supl|X_i\supl\right)=0,
\label{eq: source model} \\
Y_i\supQ&= \left(X_i\supQ\right)\trans \beta^* + \epsilon_i\supQ 
&\quad& \text{where  }\E\left(\epsilon_i\supQ|X_i\supQ\right)=0. 
\label{eq: target model}
\end{alignat} 
In the following, we first introduce a transferrable uncertainty set in Section \ref{sec: transferrable uncertainty} that narrows down and guides the source convex hull in \eqref{eq: convex hull} to be close to the target population. We then define our GART model on such a set with an adversarial regret function in Section \ref{sec: GART definition}. Section \ref{sec: GART identification} identifies our GART model as a weighted format with a nice interpretation. We shall introduce a data-dependent GART estimator in Section \ref{sec: est} and establish in Section \ref{sec: theory} that our proposed GART estimator accurately estimates the high-dimensional regression parameter $\beta^*$.  
\subsection{Transferrable uncertainty set}
\label{sec: transferrable uncertainty}
\begin{figure}[htp!]
    \centering
    \includegraphics[width=0.65\linewidth, page=1]{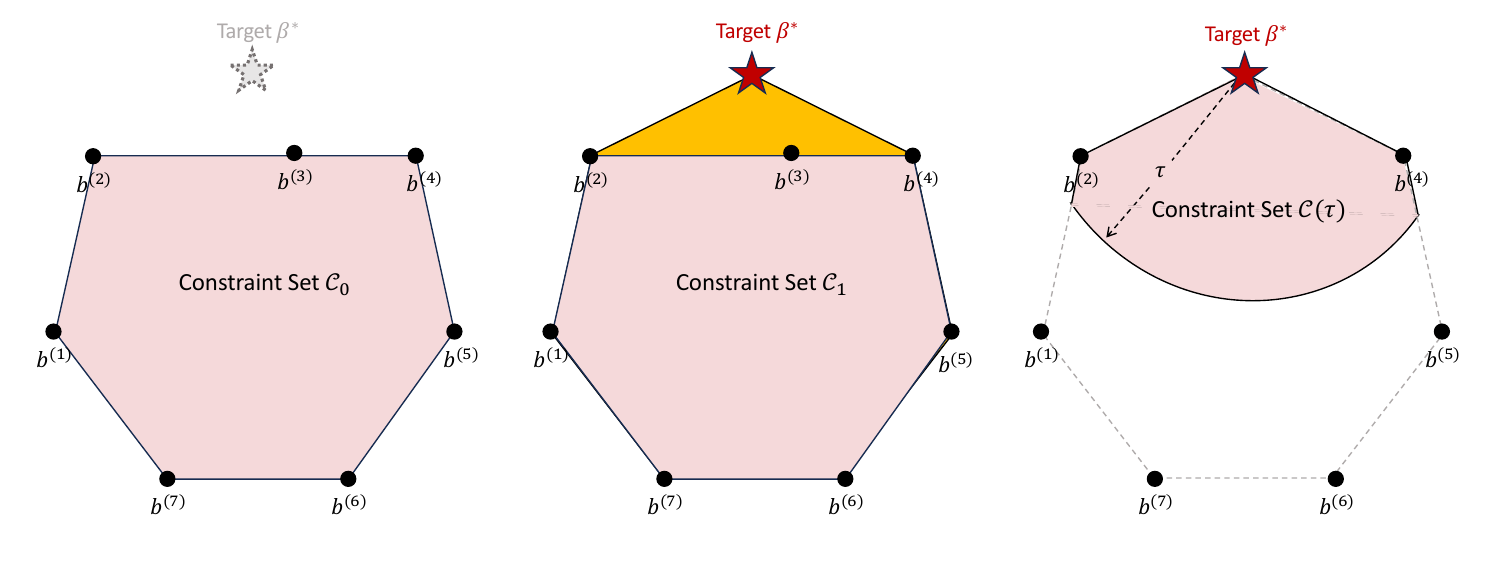}
    \caption{Illustration of the constraint sets $\mathcal{C}_0$ (left), $\mathcal{C}_1$ (center) and $\mathcal{C}(\tau)$ (right) when the true target is out of the convex hull of sources.}
    \label{fig: graph for C0 C1}
\end{figure}

Since some data from the target population is available, we treat it as another potential source and define an expanded uncertainty set $\mathcal{C}_1$ to include the target population:
\begin{equation}
\mathcal{C}_1=\left\{\T=(\Tar\subX, \T\subYX): \T\subYX=q_0 \cdot \Tar\subYX + \sum_{l=1}^{L} q_l \cdot \mathbb{P}\supl\subYX \; \text{with}\; q = (q_0,q_1,...,q_L)\in \Delta_{L_0} \right\},
\label{eq: L_0 convex hull}
\end{equation}
where $L_0=L+1$. As shown in Figure \ref{fig: graph for C0 C1}, even when the true target is away from the convex hull of sources, the uncertainty class $\mathcal{C}_1$ contains the true target while $\mathcal{C}_0$ defined in \eqref{eq: convex hull} does not.

In the presence of heterogeneity, the uncertainty set $\Csc_1$ in \eqref{eq: L_0 convex hull} may be quite large, containing distributions substantially different from the target distribution. Since our goal is to train a prediction model generalizable to future target populations not very different from $\Qbb$, we further refine $\Csc_1$ to a more concise \textit{transferrable} uncertainty set, 
\begin{equation}
        \mathcal{C}(\tau)=\{(\Tar\subX, \T\subYX)\in \mathcal{C}_1: \E_{\Tar}[\E_{\T\subYX}(Y|X)- Y]^2\leq \sigma^2\subQ+\tau \} \text{ with } \sigma^2\subQ=\E_{\Tar}[\E_{\Tar\subYX}(Y|X)- Y]^2, 
\label{eq: def Ctau}
\end{equation}
where $\tau\geq 0$ is a user-specific tolerance parameter with default value $\tau=\sigma^2\subQ/n$. 
As shown in the rightmost panel of Figure \ref{fig: graph for C0 C1}, the uncertainty set $\mathcal{C}(\tau)$ from \eqref{eq: def Ctau} restricts to source mixtures $\T\subYX$ whose prediction on the target is close to that of the true model $\Qbb\subYX$. The parameter $\tau$ controls the size of $\mathcal{C}(\tau)$ and its maximal deviation from $\mathbb{Q}$ in prediction error. As $\tau \to 0$, $\mathcal{C}(\tau)$ shrinks toward $\mathbb{Q}$.

\begin{remark}\rm 
    The idea behind $\mathcal{C}(\tau)$ shares the similarity with the Wasserstein ball that is used to define the ambiguity set in classical DRO literature \citep{shafieezadeh2015distributionally,mohajerin2018data,zhao2018data,shafieezadeh2019regularization} when only the target data is available. The Wasserstein ball contains distributions close to the empirical target distribution in terms of the Wasserstein distance, while our transferable set seeks a mixture of source distributions with similar prediction performance compared to the true target distribution $\Tar$.
\end{remark}

\subsection{GART definition}
\label{sec: GART definition}
In the following, we first introduce the regret function and then introduce our proposed GART population prediction model. 
We use $\beta\subinit$ to denote the parameter of an initial model and define $\Sigma_{\Tar}=\E_{X\sim\Tar\subX}\left[XX\trans\right]$. For any distribution $\T\in \mathcal{C}(\tau)$, we define the regret function as the difference between $\beta$ and the initial model $\beta\subinit$ in terms of prediction error (PE), 
\begin{equation}
\RR_{\T}(\beta, \beta\subinit) := \E_{(X,Y)\sim\mathbb{T}}\left\{\left(Y - X\trans\beta\right)^2 - \left(Y - X\trans\beta\subinit\right)^2\right\} = \PE\subT(\beta) - \PE\subT(\beta\subinit), 
\label{eq: linear GART loss}
\end{equation}
where $\PE\subT(\beta) = \E_{(X,Y)\sim\mathbb{T}}\{\left(Y - X\trans\beta\right)^2\}$.
For a given $\beta\subinit$, we aim for a model $\beta$ with a smaller $\RR_{\T}(\beta, \beta\subinit)$ since
a model $\beta$ with lower $\RR_{\T}(\beta, \beta\subinit)$ attains lower PE in a future target population with distribution $\T$. Generally, $\beta\subinit$ can be any $p$-dimensional vector including $\beta\subinit =0$, which corresponds to the regret function used in the maximin estimation \citep{Meinshausen_2015}. Here, we consider choosing $\beta\subinit$ closer to $\beta^*$, which leads to improved prediction for the target population $\Tar$ as shown in Theorem \ref{thm: second guidance} in Supplementary Materials.

Combining \eqref{eq: def Ctau} and \eqref{eq: linear GART loss}, we define the GART model as the optimizer for the worst-case regret defined with respect to the transferrable uncertainty set:
\begin{equation}
     \beta\subGART(\beta\subinit, \tau) = \argmin_{\beta\in\mathbb{R}^p} \max_{\T\in\mathcal{C}(\tau)}  \RR\subT(\beta, \beta\subinit) = \argmin_{\beta\in\mathbb{R}^p} \max_{\T\in\mathcal{C}(\tau)}  \left\{\PE\subT(\beta) -  \PE\subT(\beta\subinit)\right\}.
    \label{eq: GART problem, linear}
\end{equation}
The definition of $\beta\subGART(\beta\subinit, \tau)$ can be viewed as a two-player game: for each model $\beta$, an adversary selects the most challenging $\T \in \mathcal{C}(\tau)$ that maximizes regret. Unlike traditional group DRO, $\T$ remains close to $\Tar$ due to the prediction error constraint in $\mathcal{C}(\tau)$—the adversary is guided by available target data. Thus, $\beta\subGART(\beta\subinit, \tau)$ ensures both generalizability and optimal accuracy under adversarial perturbation. The parameters $\tau$ and $\beta\subinit$ jointly balance robustness and accuracy. When clear from context, we write $\beta\subGART(\beta\subinit, \tau)$ as $\beta\subGART$.

When a source mixture closely approximates the true target, Theorem \ref{thm: first guidance} below shows that incorporating additional source sites significantly improves estimation accuracy by leveraging their larger sample sizes.
Moreover, optimizing over the rich set of distributions in $\mathcal{C}(\tau)$—rather than solely on the target—enhances model generalizability to distributions near the target.

\subsection{GART identification}
\label{sec: GART identification}
For any $\T \in \mathcal{C}1$ with weight $\gamma \in \Delta{L_0}$, its conditional mean model is:
\begin{equation}
    \E_{\T\subYX}(Y|X) = \gamma_0 \cdot X\trans\beta^* + \sum_{l=1}^L \gamma_l \cdot X\trans b^{(l)} = X\trans\BB_0\gamma \quad \text{with}\quad \BB_0=[\beta^* , b^{(1)}, \cdots , b^{(L)}]_{p \times L_0}.
    \label{eq: transform T and gamma}
\end{equation}
With fixed $\Tar\subX$, the distribution set $\mathcal{C}(\tau)$ corresponds to the weight set $\mathcal{S}(\tau)$:
\begin{equation}
     \mathcal{S}(\tau)=\{\gamma\in\Delta_{L_0}: \E_{(X,Y)\sim\Tar}(X\trans\BB_0\gamma - Y)^2\leq \sigma_{\Tar}^2+\tau\} .
    \label{eq: constraint rar}
\end{equation}
Theorem \ref{thm:eq1} establishes the identification of $\beta\subGART$ as a weighted average of $\beta^*$ and $\{b\supl\}_{1\leq l\leq L}.$ 

\begin{Theorem} Consider the models \eqref{eq: source model} and \eqref{eq: target model}. Given any $\beta\subinit$, $\beta\subGART$ defined in (\ref{eq: GART problem, linear}) has the following closed-form expression, 
\begin{equation}
\beta\subGART  = \gamma\subGART^{(0)} \beta^* + \sum_{l=1}^{L}\gamma\subGART\supl b\supl = \BB_0\gamma\subGART \quad \text{with}\quad \gamma\subGART=\argmin_{\gamma\in \mathcal{S}(\tau)} \gamma\trans\bGamma_0\gamma,
\label{eq: g-maximin step 2}
\end{equation}
where $\bGamma_0=[\BB_0 - \BB\subinit]\trans\Sigma_{\Tar}[\BB_0- \BB\subinit]$ and $\BB\subinit = [\beta\subinit,...,\beta\subinit]_{p\times L_0}$.
\label{thm:eq1}
\end{Theorem}

The optimal weight $\gamma\subGART$ can be obtained by solving a convex optimization problem since $\mathcal{S}(\tau)$ is a convex set for $\gamma$. Minimization of $\gamma\trans\bGamma_0\gamma$ under constraint $\mathcal{S}(\tau)$ is essentially searching for $\gamma\in\mathcal{S}(\tau)$ such that $\BB_0\gamma$ is closest to $\beta\subinit$. In other words, $\beta\subGART$ is the projection of $\beta\subinit$ to the convex subset $\mathcal{B}(\tau):=\{\BB_0\gamma: \gamma\in\mathcal{S}(\tau)\}.$  Although $\beta\subGART$ is not always equal the true $\beta^*$, we derive in the following proposition that $\beta\subGART$ will converge to $\beta^*$ under certain conditions. It is important to note that this convergence is established at the population level and does not involve the sample size $n$ or the dimensionality $p$.

\begin{Proposition}
Consider the models \eqref{eq: source model} and \eqref{eq: target model}. For an invertible $\Sigma_{\Tar}$, $\beta\subGART \to \beta^*$ if either $\tau \to 0$ or $\beta\subinit \to \beta^*$.
    \label{prop: population convergence}
\end{Proposition}

\begin{figure}[htp!]
    \centering
    \includegraphics[width=.7\linewidth, page=1]{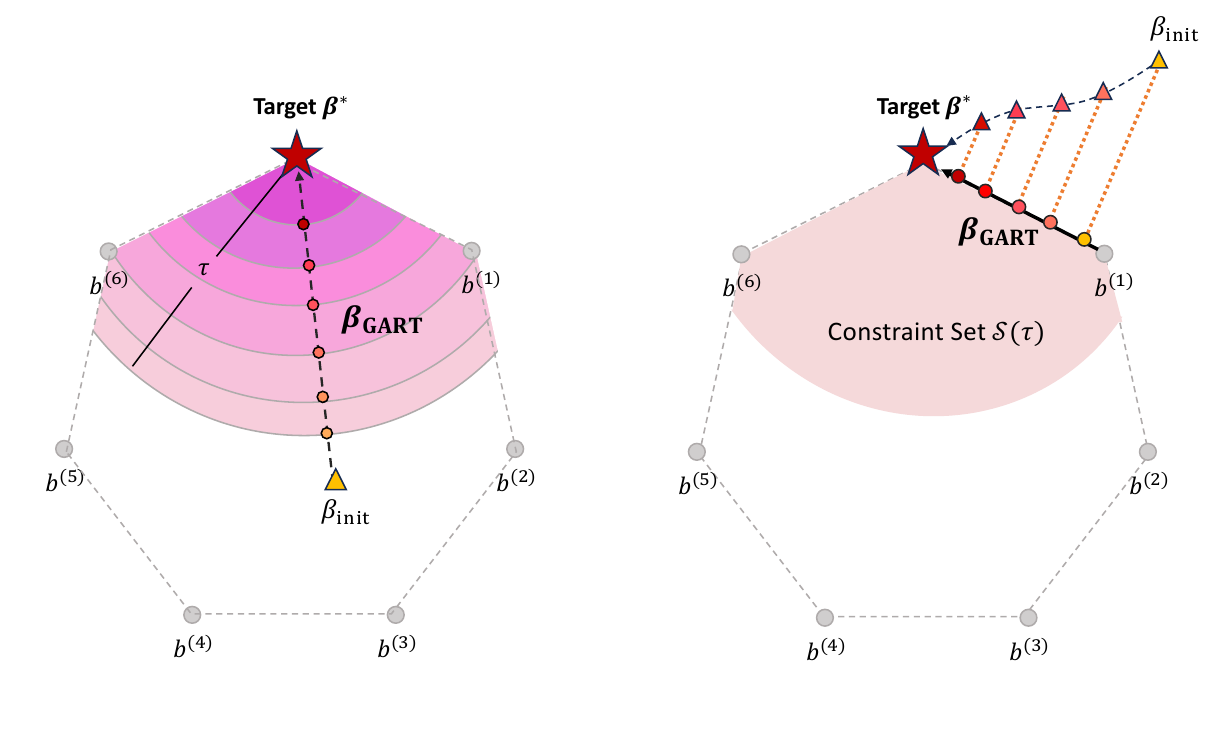}
    \caption{Graphical illustration for the identification of $\beta\subGART$ when $\tau\to 0$ (left) and $\beta\subinit\to\beta^*$ (right). The colored circles represent GART parameters under different $\mathcal{S}(\tau)$ indicated by the shaded areas or different $\beta\subinit$ indicated by triangles. }
    \label{fig: GART graph}
\end{figure}

Figure \ref{fig: GART graph} visually presents the recovery of the true $\beta^*$ when $\tau\to 0$ or $\beta\subinit\to \beta^*$. Firstly, when $\tau\to 0$, the constraint set $\mathcal{B}(\tau)$  shrinks to $\beta^*$ and hence $\beta\subGART$ converges to $\beta^*$. Secondly, if $\beta\subinit\to \beta^*$, the projection $\beta\subGART \to \beta^*$ since the uncertainty set $\mathcal{B}(\tau)$ contains $\beta^*$ by the definition of  $\mathcal{B}(\tau).$  Such a recovery of $\beta^*$ in Proposition \ref{prop: population convergence} cannot be guaranteed in traditional group DRO models due to the lack of target data
\citep{mohri2019agnostic, deng2020distributionally,wang2023DRL,Meinshausen_2015,guo2020inference}.

{As shown in Proposition \ref{prop: population convergence}, when focusing solely on the current target population (i.e., transferability), $\tau$ can be set to zero. To balance accuracy on the current target and robustness to future distribution shifts, $\tau$ can instead be a small positive constant, typically scaled as $c \cdot \frac{L\sigma\subQ^2}{n}$—which, per Theorem \ref{thm: first guidance}, does not affect GART’s convergence rate. The parameter $c$ quantifies the trade-off between generalizability and transferability. If no prior is available, a suggested choice is
$c = \text{Median}\left(\left\{\frac{\|\beta^* - b\supl\|_2}{\|\beta^*\|_2}\right\}\right),$ where the median is computed over all sources $l\in[L]$, under the intuition that if most sources deviate from the current target, future populations may as well. In practice, unknown quantities $\beta^*$, ${b\supl}$, and $\sigma\subQ^2$ can be replaced with plug-in estimators introduced in Section \ref{sec: est}.} 

\begin{remark}[Comparison between GART and group DRO] \rm GART provides a much more accurate prediction model for the target domain than the (unguided) group DRO algorithm such as the maximin \citep{Meinshausen_2015,guo2020inference}, through leveraging the available target data as a guidance. To illustrate this, recall that the maximin estimator is
\begin{equation}
    \beta\submaximin  =  \argmin_{\beta\in\R^p} \left[\max_{\mathbb{T}\in\mathcal{C}_0} \left
    \{\PE\subT(\beta) - \PE\subT(0)\right\} \right] = \sum_{l=1}^{L}\gamma\submaximin\supl b\supl = \BB\gamma\submaximin ,
\end{equation}
with $\gamma\submaximin=\argmin_{\gamma\in \Delta_L} \gamma\trans\bGamma\gamma$, where
$\BB = [b^{(1)},...,b^{(L)}]_{p\times L}$ and $\bGamma=\BB\trans\Sigma_{\Tar}\BB$. 
When a variable has heterogeneous effects scattered around zero across multiple sources, its maximin effect will shrink to the zero initial due to the design of the regret function. Such a shrinkage is often appealing, especially in high dimensional data, since zero is a safe choice as prior knowledge on $\beta$. Yet, under our model setting, the zero initial, ignoring information in the existing target labels, may be a poor choice. Instead, we shall leverage the target data to introduce a more informative initial model; see Section \ref{sec: init}. 
Another distinction between maximin and GART is that the maximin model searches over the entire convex hulls spanned by the support of $\{b\supl\}_{1\leq l \leq L}$ while GART limits its attention to a smaller area (i.e., $\mathcal{C}(\tau)$) close to the target. We will show in the simulation study that the additional guidance in the loss function and in the transferrable constraint set renders GART a more accurate estimation performance than the unguided DRO prediction model. 
\end{remark}

\section{Estimation of the GART model}
\label{sec: est}
We next detail our estimation procedure for $\beta\subGART$. In Section \ref{sec: est of Stau}, we construct estimators for $\beta^*, b\supone,...,b\supl$ and $\sigma_{\Tar}^2$ and $\mathcal{S}(\tau)$. In Section \ref{sec: init}, we present different choices of the initial estimator $\widehat{\beta}\subinit$. Finally, we present a plug-in estimator for $\beta\subGART$ in \eqref{eq: sample version of GART}. Throughout, we let $\YY\supl=[Y_1\supl, ..., Y_{N_l}\supl]\trans$, $\YY\supQ=[Y_1\supQ, ..., Y_n\supQ]\trans$, $\XX\supl=[X_1\supl, ..., X_{N_l}\supl]\trans$, and $\XX\supQ=[X_1\supQ, ..., X_n\supQ]\trans$. For any subset $\Asc \subset \{1,...,n\}$, $\YY_{\Asc}\supQ$ and $\XX_{\Asc}\supQ$ denotes the rows of $\YY\supQ$ and $\XX\supQ$ corresponding to $\Asc$.

\subsection{Construction of $\widehat{\mathcal{S}}(\tau)$}
\label{sec: est of Stau}
With each source data, standard estimators for $b\supl$ can be used. For convenience, we focus on the Lasso estimator to allow for high dimensional $X$ with 
$$
\widehat{b}\supl = \argmin_{\beta\in\R^p}\left[  N_l^{-1} \|\YY\supl - \XX\supl\beta\|_2^2  + \lambda_l\|\beta\|_1\right], 
$$
where $\lambda_l>0$ is the tuning parameter chosen by cross-validation. 
GART only requires each local site to pass $\widehat{b}\supl$ to the target without passing the individual-level source data. 

Since $\mathcal{S}(\tau)$ is intrinsically designed to choose a combination of source and target prediction models that guarantee the predictive performance on the observed target domain. To achieve this, we split the target data into two parts $(\YY\suba\supQ, \XX\suba\supQ)$ and $(\YY\subaa\supQ, \XX\subaa\supQ)$ of approximately equal sample sizes, where $\Asc_1 = \Asc_2^c = \{1,...,n\}\setminus \Asc_2$. We use $n_1$ and $n_2$ to denote the sample sizes of $\Asc_1$ and $\Asc_2$, respectively. The first part is used to train a roughly accurate estimator of $\beta^*$, while the second part is used to evaluate the predictive performance of different models. With the first half of the data $(\YY\suba\supQ, \XX\suba\supQ)$, we estimate $\beta^*$ by
\begin{equation}
\widehat{\beta}\suba = \argmin_{\beta} n_1^{-1}\|\YY\suba\supQ - \XX\suba\supQ\beta\|_2^2+\lambda_0\|\beta\|_1,
\label{eq: estimate b0}
\end{equation}
where $\lambda_0$ is a positive tuning parameter chosen by cross-validation. We evaluate the predictive performance of $\widehat{\beta}\suba$ over the independent second part of data and estimate $\sigma_{\Tar}^2$ by
\begin{equation}
    \widehat{\sigma}_{\Tar}^2 = n_2^{-1}\|\YY\subaa\supQ - \XX\subaa\supQ\widehat{\beta}\suba\|_2^2.
\label{eq: est for sigmaQ, 1}
\end{equation}
Given the limited amount of target data, $\widehat{\sigma}_{\Tar}^2$ might be unstable. As an alternative, when the source mixing assumption holds, we apply the idea of approximating $\Qbb_{\subYX}$ by a mixture of $\{\mathbb{P}\supl\subYX\}_{1\leq l\leq L}$ and construct a source-mixing estimator of $\beta^*$. Specifically, we let 
\begin{equation} 
\widehat{\beta}\suba^{\dag} = \BBhat\widehat{\gamma}\suba \quad \text{with} \quad \widehat{\gamma}\suba = \argmin_{\gamma\in\Delta_L} \left\{n_1^{-1}\|\YY\suba\supQ - \XX\suba\supQ\BBhat\gamma\|_2^2 \right\},\
\label{eq: est for ini gamma}
\end{equation}
where $\BBhat=[\widehat{b}^{(1)},...,\widehat{b}^{(L)}].$ We further estimate $\sigma\subQ^2$ by 
\begin{equation}
\widehat{\sigma}^2\subsource = n_2^{-1}\|\YY\subaa\supQ - \XX\subaa\supQ\BBhat\widehat{\gamma}\suba^{\dag}\|_2^2.
\label{eq: est for sigmaQ, 2}
\end{equation}
When the source-mixing assumption holds, $\widehat{\sigma}^2\subsource$ is a consistent estimator of $\sigma_{\Tar}^2$. 
When the source-mixing assumption is violated, $\widehat{\sigma}^2\subsource$ is larger than $\sigma_{\Tar}^2$ with high probability. 

A robust estimate for $\sigma_{\Tar}^2$ is  constructed as $\min\left\{\widehat{\sigma}_{\Tar}^2, \widehat{\sigma}\subsource^2\right\}$, which leads to  
\begin{equation}
    \widehat{\mathcal{S}}(\tau) = \left\{\gamma\in\Delta_{L_0}: n_2^{-1}\|\YY\subaa - \XX\subaa\BBhat_0\gamma \|_2^2 \leq \min\{\widehat{\sigma}_{\Tar}^2,\widehat{\sigma}\subsource^2\}+\tau\right\}, \quad \text{with}\quad \BBhat_0=[\widehat{\beta}\suba, \BBhat].
    \label{eq: constraint d}
\end{equation} 
The rationale behind using $\min\{\widehat{\sigma}_{\Tar}^2,\widehat{\sigma}\subsource^2\}$ to estimate the target noise level $\sigma\subQ^2$ is to strike a balance between two competing properties: the consistency but high variability of the target-only estimator $\widehat{\beta}\suba^{\dagger}$ in $\widehat{\sigma}_{\Tar}^2$, and the efficiency but potential bias of the source-mixing estimator $\BBhat\widehat{\gamma}\suba^{\dagger}$ as part of $\widehat{\sigma}\subsource^2$. Including the target estimator as a candidate within $\BBhat$ in equation \eqref{eq: est for ini gamma} would introduce its high variance into $\widehat{\sigma}\subsource^2$, thereby undermining the intended balance between consistency and efficiency.

\subsection{Different choices of the initial estimator $\widehat{\beta}\subinit$}
\label{sec: init}
In this section, we leverage the source data to design initial estimators close to the target model. We first consider a convex combination of the source prediction models that is closest to the target:
\begin{equation}
    \widehat{\beta}\subinit\supsource = \BBhat \argmin_{\gamma\in\Delta_L} n^{-1}\|\YY\supQ - \XX\supQ\BBhat\gamma\|_2^2.
    \label{eq: def of source mix init beta}
\end{equation}

Since the source mixing assumption may fail and some source sites may be viewed as adversarial sites with opposite signs of effect compared to $\beta^*$, we also consider a linear combination of sources and the target as an initial estimator via cross-fitting. Specifically, for $j \in \{1,2\}$, we use half of the data $\{\YY\subaj\supQ,\XX\subaj\supQ\}$ to construct $\widehat{\beta}\subaj$ and define $\BBhat_{0\mathcal{A}_j} = (\widehat{\beta}\subaj, \widehat{b}^{(1)},...,\widehat{b}\supL)$. We then use the remaining data to obtain the aggregation weight $\widehat{\gamma}_{\mathcal{A}_j}$, 
\begin{equation}
\begin{aligned}
        \widehat{\gamma}_{\mathcal{A}_j}\supweighted &= \argmin_{\gamma\in\Delta\sublinear} n\subajc^{-1}\|\YY\subajc\supQ - \XX\subajc\supQ\BBhat_{0\mathcal{A}_j}\gamma\|_2^2, 
\end{aligned}
    \label{eq: convex}
\end{equation}
where $\Delta\sublinear = \{\gamma: \gamma_0\in[0,1], \gamma_l\in[-1,1]\text{ for }l=1,...,L\}$. We use the average of the two cross-fitting optimal linear combinations as an initial estimator:
\begin{equation}
    \widehat{\beta}\subinit\supweighted= \left(\BBhat_{0\mathcal{A}_1}\widehat{\gamma}_{\mathcal{A}_1}\supweighted + \BBhat_{0\mathcal{A}_2}\widehat{\gamma}_{\mathcal{A}_2}\supweighted\right) / 2.
    \label{eq: def of weighted init beta}
\end{equation}

{Cross-fitting is crucial to maintain independence between the estimator $\widehat{\beta}\subaj$, a candidate in $\BBhat_{0\mathcal{A}_j}$, and the weight vector $\widehat{\gamma}\subaj\supweighted$, which determines site importance for approximating the target effect. Without cross-fitting, using the same target data ${\YY\supQ, \XX\supQ}$ for both would cause $\widehat{\gamma}\supweighted$ to favor $\widehat{\beta}$, allocating all weight to it. Cross-fitting avoids this circularity by training $\widehat{\beta}\subaj$ and $\widehat{\gamma}\subaj\supweighted$ on disjoint subsets. While the approach can be extended to more folds, doing so reduces per-fold sample size -- undesirable with limited target data -- so we adopt a two-fold split as a practical compromise.} The non-negative condition for $\gamma_0$ in $\Delta\sublinear$ indicates the positive guidance from the existing target model. Including negative weights for sources endows $\widehat{\beta}\subinit$ with more flexibility to capture possible relationships between the target and source models. Similar ideas of negative weights have been applied in the literature of the risk extrapolation model \citep{Krueger2020}. 

\begin{remark} \rm
Our initial estimators' construction resembles model averaging \citep{hansen2007least}, which aims to provide accurate predictions by combining the results from individual models. 
In contrast to averaging multiple candidate estimated models for one true data generation mechanism, we use a linear combination of heterogeneous source and target models to form an initial prediction model on the (future) target domain. {
Simple constructions of $\widehat{\beta}\subinit$ range from a naive zero vector to the Maximin estimator. Theoretically, its choice does not affect the convergence rate in Theorem \ref{thm: first guidance}, provided $\tau = \mathcal{O}(1/n)$. However, when the focus is on the current target population, Theorem \ref{thm: second guidance} shows that a more accurate $\widehat{\beta}\subinit$ -- closer to the true target model -- can reduce estimation error by a constant factor. For generalizability to future populations, which may differ from the current target, we discuss alternative initial estimators in Section \ref{supp: initial estimator} of the Supplementary Materials.
}
\end{remark}

With $\BBhat_0=[\widehat{\beta}\suba, \BBhat]$, $\BBhat\subinit = [\widehat{\beta}\subinit, ..., \widehat{\beta}\subinit]_{p\times L_0}$, and  $\widehat{\Ssc}(\tau)$, we construct a plug-in estimator for $\beta\subGART$ as $\widehat\beta\subGART = \BBhat_0\widehat{\gamma}\subGART$, where $$\widehat{\gamma}\subGART=\argmin_{\gamma\in \widehat{\mathcal{S}}(\tau)} \gamma\trans\bGammahat_0\gamma, \quad 
\bGammahat_0=[\BBhat_0- \BBhat\subinit]\trans\widehat{\Sigma}_{\Tar}[\BBhat_0- \BBhat\subinit],$$  with $\widehat{\Sigma}\subQ = \frac{1}{n}\sum_{i=1}^n X_i\supQ (X_i\supQ)\supT$. 
Algorithm \ref{alg: GART estimation} summarizes the GART estimation procedure.   {Notably, the GART algorithm requires no exchange of raw data between source sites or with the target site. Only summary-level statistics (i.e., $\BBhat$) are needed on the target side, aligning with data privacy regulations.}

\section{Theoretical analysis}
\label{sec: theory}
\subsection{Model assumptions}
Before presenting the main theorems, we introduce the assumptions for GART. For conciseness, we present theoretical results mainly for the more challenging high-dimensional settings but add remarks on cases when the results for the low and high-dimensional settings differ substantially. We assume the high-dimensional vectors satisfy the following capped-$l_1$ sparsity condition:
\begin{equation}
\sum_{j=1}^p \min\bigg\{{|\beta^*_j|}/(\sigma_{\Tar}\lambda_0), 1\bigg\}\leq k, \,\,\,\, \sum_{j=1}^p \min\bigg\{{|b^{(l)}_j|}/(\sigma_{l}\lambda_l), 1\bigg\}\leq k_l \quad \text{ for } \quad l\in\{1,...,L\},
\label{eq: capped l1 condition}
\end{equation}
where $\lambda_0 = \sqrt{2\log p /n}$ and $\lambda_l = \sqrt{2\log p / N_l}$. As noted in \cite{cappedl1}, the capped-$l_1$ condition in \eqref{eq: capped l1 condition} holds if $\beta^*$ is $l_0$ sparse with $\|\beta^*\|_0\leq k$ or $l_q$ sparse with ${\|\beta^*\|_q^q}/{(\sigma_{\Tar}\lambda_0)^q}\leq k$ for $0<q\leq 1$. Hence, the capped-$l_1$ sparsity condition is more relaxed compared to the $l_0$ sparsity assumption.
We introduce the following conditions on the models \eqref{eq: source model} and \eqref{eq: target model}.
\begin{Assumption}
For $1\leq l \leq L$, the regression vector $b\supl$ satisfies \eqref{eq: capped l1 condition} with $k_l \log p/N_l \leq c$ for some small positive constant $c>0$ and 
$\{X_i\supl, Y_i\supl\}_{1\leq i \leq N_l}$ are i.i.d. random variables, where $X_i\supl\in\mathbb{R}^p$ is sub-gaussian with $\Sigma\supl = \E[X_i\supl(X_i\supl)\trans]$  satisfying $c_0\leq \lambda_{\text{min}}(\Sigma\supl)\leq \lambda_{\text{max}}(\Sigma\supl)\leq C_0$ for positive constants $C_0> c_0>0$. For $1\leq l \leq L$, the error $\epsilon_i\supl$ is sub-gaussian with $\E[\epsilon_i\supl|X_i\supl]=0$ and we denote $\sigma_l^2=\E[(\epsilon_i\supl)^2]$. 
\label{assumption: a_1, source}
\end{Assumption} 

\begin{Assumption}
The regression vector $\beta^*$ satisfies \eqref{eq: capped l1 condition} with $k \log p/n \leq c$ for some small positive constant $c>0$ and 
$\{X_i\supQ, Y_i\supQ\}_{1\leq i \leq n}$ are i.i.d samples, where $X_i\supQ\in\mathbb{R}^p$ is sub-gaussian with $\Sigma_{\Tar} = \E[X_i\supQ(X_i\supQ)\trans]$ satisfying $c_1\leq \lambda_{\text{min}}(\Sigma_{\Tar})\leq \lambda_{\text{max}}(\Sigma_{\Tar})\leq C_1$ for positive constants $C_1> c_1>0$. The error $\epsilon_i\supQ$ is sub-gaussian with $\E[\epsilon_i\supQ|X_i\supQ]=0$ and we write $\E[(\epsilon_i\supQ)^2] = \sigma_{\Tar}^2$.
\label{assumption: a_1}
\end{Assumption}

Assumptions \ref{assumption: a_1, source} and \ref{assumption: a_1} are commonly assumed for the theoretical analysis of high-dimensional linear models \citep[e.g.]{buhlmannbook,raskutti2010restricted,zhou2009restricted}. The positive definite $\Sigma\supl$ and the sub-gaussianity of $X_i\supl$ guarantee the restricted eigenvalue condition with a high probability. The sub-gaussian errors are generally required for the theoretical analysis of the Lasso
estimator in high dimensions. 

In the following assumption, we state the convergence rates of site-specific Lasso estimators that have been established in the literature \citep[e.g.]{bickel2009,buhlmannbook,scalelasso2012,lasso2010} under Assumptions \ref{assumption: a_1, source} and \ref{assumption: a_1}.
\begin{Assumption}
For $l=1,...,L$, the source estimator $\widehat{b}^{(l)}$ satisfies that with probability larger than $1 - \delta_l(N_l)$ and $\delta_l(N_l)\to 0$,
$
    \max\left\{\frac{1}{n} \|\XX\supQ(\widehat{b}\supl-b\supl) \|_2^2, ||\widehat{b}\supl - b\supl||_2^2\right\} \lesssim \frac{k_l\log p}{N_l}\sigma_l^2. 
$
Similarly, the estimators $\widehat{\beta}\suba$ and $\widehat{\beta}\subaa$ derived from the target data $(\XX\supQ, \YY\supQ)$ satisfy that with probability larger than $1 - \delta(n)$ with $\delta(n)\to 0$, $\max\left\{\frac{1}{n}\|\XX\supQ(\widehat{\beta}\subaj-\beta^*)\trans\|_2^2, ||\widehat{\beta}\subaj - \beta^*||_2^2\right\} \lesssim \frac{k\log p}{n}\sigma\subQ^2$ for $j=1,2.$
\label{assumption: b_1}
\end{Assumption}
{Although Assumption \ref{assumption: b_1} can be derived for the Lasso estimator under Assumptions \ref{assumption: a_1, source} and \ref{assumption: a_1}, we aim to allow broader applicability beyond Lasso. Therefore, we state it separately. Any estimator satisfying all three assumptions yields a GART estimator with the convergence rate established in the next section.}

\subsection{Theoretical Properties}
We present the theoretical properties of the GART estimator and establish in the following theorem that our GART estimator $\widehat{\beta}\subGART$ converges to $\beta\subGART$ in probability. 
\begin{Theorem}
Consider the linear models \eqref{eq: source model} and \eqref{eq: target model} and suppose that Assumptions \ref{assumption: a_1, source}, \ref{assumption: a_1}, and \ref{assumption: b_1} hold. When $\min_{l} N_l$ and $n$ go to infinity, $\widehat{\beta}\subGART$ in \eqref{eq: sample version of GART} converges in probability to $\beta\subGART$ in \eqref{eq: GART problem, linear}. 
\label{thm: consistency}
\end{Theorem}

Next, we establish the upper bound for the estimation error $\widehat{\beta}\subGART-\beta^*$. The convergence rate depends on the radius $\tau$ of the constraint set $\mathcal{C}(\tau)$ in \eqref{eq: def Ctau} and the distance between $\beta^*$ and its best linear source approximation, defined as
 \begin{equation}
    \alpha = \min_{\gamma\in\Delta_L} (\beta^* - \BB\gamma)\trans \Sigma_{\Tar} (\beta^* - \BB\gamma).
    \label{eq: def of alpha}
    \end{equation} 
When the source mixing assumption holds, then $\alpha=0$. We denote a corresponding set $\mathcal{M}$ that contains any $\gamma$ minimizing $ (\beta^* - \BB\gamma)\trans \Sigma_{\Tar} (\beta^* - \BB\gamma)$:
\begin{equation}
    \mathcal{M}:= \left\{\gamma\in\Delta_L:  (\beta^* - \BB\gamma)\trans \Sigma_{\Tar} (\beta^* - \BB\gamma) = \alpha \right\},
    \label{eq: def of true source gamma}
    \end{equation}
where $\alpha$ is defined as the minimum of source approximation error in \eqref{eq: def of alpha}. When $\mathcal{M}$ includes only one element, $\gamma^*$ is defined as the unique minimizer of $(\beta^* - \BB\gamma)\trans \Sigma_{\Tar} (\beta^* - \BB\gamma)$. When  $\mathcal{M}$ includes multiple elements (e.g., colinearity exists between the columns of $\BB$), $\gamma^*$ is defined as the one minimizing the expected estimation error:
\begin{equation}
\gamma^* = \argmin_{\gamma\in\mathcal{M}}  \E\|(\BB - \BBhat){\gamma}\|_2^2.
\label{eq: best source selection}
\end{equation}

The proposition below, justified in Section \ref{sec: proof of prop 2}, shows that the source mixing assumption is weaker: if a subset of sources lies within $h$-distance of the target, then the distance $\alpha$ between the target and the convex hull of all sources is controlled.
\begin{Proposition}
Let the transferrable set be $\mathcal{A}_h := {1 \leq l \leq L : |\beta^* - b^{(l)}|_1 \leq h}$ for some $h \geq 0$. If $\mathcal{A}_h$ is non-empty, then $\alpha \leq C_1 h^2$.
\label{prop: relationship between alpha and h}
\end{Proposition}

We now establish the finite-sample convergence rate of $\widehat{\beta}\subGART$ that is simultaneously controlled by the rate of the target-data-only estimator and the best source-mixing estimator.
\begin{Theorem}
Consider the linear models \eqref{eq: source model} and \eqref{eq: target model} and suppose that Assumptions \ref{assumption: a_1, source}, \ref{assumption: a_1} and \ref{assumption: b_1} hold and $L\ll n$. 
Then, there exists positive constants $C>0$ and $c>0$ such that, with probability at least {$1-\exp(-c(t^2\cdot L+n))-\delta(n)$} for $t\geq C$, $\widehat{\beta}\subGART$ defined in \eqref{eq: sample version of GART} satisfies 
\begin{align}
     \|\beta^* - \widehat{\beta}\subGART\|_2^2 \lesssim \tau + \frac{t^2L}{n} + \min\bigg\{ \frac{k\log p}{n}\sigma\subQ^2, \,\,\alpha + \|(\BB - \BBhat){\gamma}^*\|_2^2\bigg\},
     \label{eq: thm 3}
\end{align}
where $\alpha$ is defined in \eqref{eq: def of alpha} and $\delta(n)$ is defined in Assumption \ref{assumption: b_1}.
\label{thm: first guidance}
\end{Theorem}

In the following, we provide the intuition for \eqref{eq: thm 3} and discuss the rate improvement of using $\widehat{\beta}\subGART$ compared to the target-only estimator. We consider an important regime with $\tau\lesssim {L}/{n}$ and $L\lesssim k\log p$, under which the estimation error of $\widehat{\beta}\subGART$ becomes 
\[
\|\beta^* - \widehat{\beta}\subGART\|_2^2 \lesssim \min\bigg\{ \frac{k\log p}{n}\sigma\subQ^2, \,\,\frac{L}{n}+\alpha + \|(\BB - \BBhat){\gamma}^*\|_2^2\bigg\}.
\] 
The first term of the right handside is the convergence rate of the Lasso estimator only using target data, and the second term is related to the error for the best source-mixing estimator. Thus, the GART estimator adaptively attains a better convergence rate between the target-only Lasso estimator and the source mixing estimator without explicitly comparing these two estimators. We achieve this data-adaptive selection between the target only and the source mixing estimator by carefully designing the constraint set $\widehat{\mathcal{S}}(\tau)$ in \eqref{eq: constraint d}. Importantly, the GART estimator is robust to the violation of the source-mixing assumption: when the source mixing poorly approximates the target, the approximation error $\alpha$ will be large and the GART estimator is reduced to the target-only estimator.

We now demonstrate the usefulness of Theorem \ref{thm: first guidance} by considering both low- and high-dimensional regimes. 
We start with the low-dimensional regime, where we estimate the regression parameters by OLS instead of the penalized regression. 
\begin{Corollary} 
Under the low-dimensional case with $p\ll \
n$, consider the linear models \eqref{eq: source model} and \eqref{eq: target model} and suppose that Assumptions \ref{assumption: a_1, source} and \ref{assumption: a_1} hold. Suppose that we construct $\widehat{b}^{(l)}$ and $\widehat{\beta}$ as the OLS estimators. 
Then, there exists positive constants $C>0$ and $c>0$ such that, with probability at least {$1-\exp(-c(t^2\cdot L+n))-t^{-1}$} for $t\geq C$, $\widehat{\beta}\subGART$ satisfies 
\begin{align}
     \|\beta^* - \widehat{\beta}\subGART\|_2^2 \lesssim \tau + \frac{t^2L}{n} + \min\bigg\{ \frac{p}{n}\sigma\subQ^2, \,\,\alpha + \sum_{l=1}^L (\gamma_l^*)^2 \frac{tp}{N_l}\sigma_l^2\bigg\}.
     \label{eq: cor low dim rate}
\end{align}
\label{Cor: low-dim}
\end{Corollary}
\vspace{-35pt}
Note that the optimal $\gamma^*$ in \eqref{eq: best source selection} is defined to minimize the variance of $\BBhat{\gamma}^*$ across all convex combinations of the columns of $\BBhat$. 
We may consider a special case where all sites have $b^{(l)}=\beta^*$. In this scenario, $\gamma^*$ is reduced to a particular type of inverse-variance weighting with $\gamma^*_j=1/V_j/(\sum_{l=1}^{L}1/V_l)$ where $V_l = \E\|\widehat{b}^{(l)}-b^{(l)}\|_2^2.$ When $\Sigma^{(l)}=\Sigma$ for $1\leq l\leq L$ and the regression errors are homoscedastic 
 with $\sigma_l^2=\E[(\epsilon\supl)^2|X],$ then we obtain  $\gamma^*_j=N_j/\sigma_j^2/(\sum_{l=1}^{L}N_l/\sigma_l^2)$, which corresponds to the weight of the standard inverse-variance weighting. We now turn to the more challenging high-dimensional regime. 
\begin{Corollary}
Consider the linear models \eqref{eq: source model} and \eqref{eq: target model} and suppose that Assumptions \ref{assumption: a_1, source}, \ref{assumption: a_1} and \ref{assumption: b_1} hold. 
Then, there exists positive constants $C>0$ and $c>0$ such that, with probability at least {$1-\exp(-c(t^2\cdot L+n))-\delta(n)- \sum_{l=1}^L \delta_l(N_l)$} for $t\geq C$, $\widehat{\beta}\subGART$ defined in \eqref{eq: sample version of GART} satisfies 
\begin{align}
     \|\beta^* - \widehat{\beta}\subGART\|_2^2 \lesssim \tau + \frac{t^2L}{n} + \min\bigg\{ \frac{k\log p}{n}\sigma\subQ^2, \,\,\alpha + \left(\sum_{l=1}^L \gamma_l^* \sqrt{\frac{k_l\log p}{N_l}\sigma_l^2}\right)^2\bigg\}.
\end{align}
\label{Cor: high-dim}
\end{Corollary}
\vspace{-35pt}

The above corollary shows that GART efficiently leverages auxiliary source data and improves the convergence rate over the target-only estimator under three conditions:
i) the source sample sizes are much larger than the target's ($n \ll \min_{l \in {1,\dots,L}} N_l$);
ii) the number of sources is small relative to the feature dimension ($L \ll k \log p$); and
iii) the distance $\alpha$ between the target model and the best source mixture vanishes faster than $k \log p / n$.
Under these conditions, GART replaces the target-only rate of $k \log p / n$ with the smaller $L / n$, yielding a faster convergence rate. The $n$ in the denominator reflects the necessary cost of using target labels to construct the transferable uncertainty set—a term that also appears in TransLasso’s rate (Section \ref{sec: compare to translasso}) due to estimating the transferable set from target data.

\begin{remark}[Non-unique source mixing] \rm 
When the columns of $\BB_0$ are colinear, multiple source mixtures of $\{b^{(l)}\}_{1 \leq l \leq L}$ may approximate $\beta^*$ well. As shown in Theorem \ref{thm: first guidance}, GART achieves the rate associated with the best source mixing $\gamma^*$ from \eqref{eq: best source selection}. However, bounding $\|(\BB - \BBhat)\gamma^*\|_2^2$ in the high-dimensional setting (as in \eqref{eq: thm 3}) poses a theoretical challenge. In low dimensions, the OLS estimator satisfies $\E(\widehat{b}^{(l)}) = b^{(l)}$, leading to $\E\|(\BB - \BBhat){\gamma}^*\|_2^2=\sum_{l=1}^{L}(\gamma^*_l)^2 \E \|\widehat{b}^{(l)}-b^{(l)}\|_2^2.$ This equality does not hold in high dimensions, where penalized estimators $\widehat{\beta}^{(l)}$ are biased due to regularization, and averaging them via $\gamma^*$ does not yield the same error reduction seen in the low-dimensional case.
\end{remark} 

\subsection{Comparison to TransLasso}
\label{sec: compare to translasso}

\label{remark: comparison with translasso}
\rm The TransLasso \citep{transLasso} estimator defined a set of transferrable sources $
\mathcal{A}_h = \{1\leq l \leq L: \|\beta^* - b^{(l)}\|_1 \leq h\}$, 
where $h$ controls the `informativeness' of the source model. 
Under the case with an unknown $\mathcal{A}_h$, the TransLasso estimator can achieve the following convergence rate, \begin{equation}
\|\widehat{\beta}^{\rm transLasso}-\beta^*\|_2^2\lesssim \frac{\log L}{n} + \frac{k\log p}{n + \sum_{l\in\mathcal{A}_h} N_l}+\min\left\{\frac{k\log p}{n},h\sqrt{\frac{\log p}{n}}, h^2\right\}.
\label{eq: translasso}
\end{equation}
TransLasso greatly improve the rate of the target-only estimator when $h\ll k\sqrt{{\log p}/{n}}$ and $\mathcal{A}_h$ is non-empty.
That is, only if there exist source models that are sufficiently similar to the target model $\beta^*$, the large sample sizes of the source data sets can be applied to improve the learning performance of the target model significantly. The constraint for $h$ can be easily violated given site heterogeneity, leading to an empty $\mathcal{A}_h$ and degenerates TransLasso into the target-data-only estimator. On the contrary, the GART estimator only requires a relatively close location between $\beta^*$ and the source convex hull (i.e., $\alpha\lesssim {k\log p/n}$) to ensure a successful transfer, which substantially relaxes the constraint for source site similarity.

When $L^*$ large sources exactly match the target, i.e., $b^{(l)} = \beta^*$ for $l \in \mathcal{A}_0$,  GART achieves the same optimal rate as TransLasso when $L$ is bounded as detailed below. 
\begin{Corollary}
    Assume a case when $\alpha = h = 0$, $|\mathcal{A}_0|=L^*>0$,  $\sum_{l\in\mathcal{A}_0}N_l\gg n$, and $\sigma\subQ^2=\sigma_l^2$ and $k = k_l$ for all $l\in \mathcal{A}_0$. When $L$ is bounded, by setting $\tau=\mathcal{O}(1/n)$, TransLasso and GART have the same estimation error rate as $\frac{k\log p}{\sum_{l\in \mathcal{A}_0}N_l} + \frac{1}{n}$. When $L$ grows with $n$ and $p$, GART's rate is upper bounded by $\frac{L}{n} +\frac{p}{\sum_{l\in\mathcal{A}_0} N_l}$ in low dimensions and $\frac{L}{n} +\frac{k\log p}{\frac{1}{L^*}\sum_{l\in\mathcal{A}_0}N_l}$ in high dimensions. 
    \label{corol: ideal case}
\end{Corollary}
When $L$ increases with $n$ and $p$, GART searches over a larger weight space than TransLasso, making it harder to exploit similar source sites under the same similarity condition. In the low-dimensional case, the main difference is the $L/n$ term in GART’s rate versus $\log L / n$ in TransLasso. Closing this gap may require more refined analysis (e.g., \citep{sign-constraint}) that leverages the non-negativity and simplex constraints in GART’s weights. In the high-dimensional case, GART's rate is slightly disadvantaged: it scales with $L^*(\sum_{l\in\mathcal{A}_0} N_l)^{-1}$ instead of $(\sum_{l\in\mathcal{A}_0} N_l)^{-1}$ as in TransLasso. This slower rate arises because simple averaging of Lasso estimators cannot eliminate the individual bias of each. A potential remedy is to use GART’s optimal weights to combine debiased Lasso estimators, reducing bias while maintaining robustness.
Whether GART can match TransLasso’s rate under more general source mixing remains an open question. However, in practice, $L$ is typically small, making this issue less critical.

\section{Simulation}
This section presents simulation results evaluating GART across various settings, compared with two popular transfer learning methods: TransGLM \citep{transglm} and TransLasso \citep{transLasso}.We also include the following benchmarks:
i) a target-only estimator,
ii) the best source mixture: $\argmin_{\beta=\BBhat\gamma,, \gamma \in \triangle_L} \frac{1}{n} |Y\supQ - X\supQ \beta|_2^2$, and
iii) the maximin estimator \citep{Meinshausen_2015}.
To fit the GART model, we consider various initial estimators $\widehat{\beta}\subinit$ constructed as weighted averages of ${\widehat{\beta}^*, \widehat{b}_1, \dots, \widehat{b}_L}$. Unless otherwise specified, we use the optimal linear combination defined in \eqref{eq: def of weighted init beta}, with $\tau = 1/n$ {by default}.
{For each data-generating scenario, we report the mean and the 2.5th and 97.5th percentiles of the mean squared error (MSE) across 100 independent simulations.}

\def\bE{\mathbf{E}}
\subsection{Data generation mechanisms}
Across all simulation studies, we generate $Y \mid X$ from linear models with different choices of $\{\beta^*; b\supl, l=1,...,L\}$ across different settings, $\epsilon_i\supl\sim N(0, 0.5)$, and $\epsilon_i\supQ\sim N(0, 1)$. 
Throughout, we let $N_1=...=N_L=20,000$, and let $n=200$ unless noted otherwise. To evaluate the performance of different estimators,  we generate independent validation datasets with $N\subvalid=5000$, and consider different choices of $\beta^*\subvalid$. 

In Setting 1, we set $p = 30$ and $L = 8$, and define the target and source coefficients as:
\begin{align}
\beta^* & = [0.3, 0.1, 0.5, -0.2, -0.3, 0.2, 0.1, 0.15, 0.03, 0.01, -0.15, -0.03, -0.01, \mathbf{0}{17}\trans]\trans, \nonumber \\
b^{(l)} & = \beta^* + \kappa \times \left[(-1)^l\lceil l/2 \rceil \mathbf{1}{15}\trans, ,,, \epsilon_1,\dots,\epsilon_{15} \right]\trans, \quad \epsilon_k \sim N(0,1), ,,, l \in {1,\dots,8},
\label{eq: sim sd source}
\end{align}
where $\kappa$ controls both site heterogeneity and the violation level of the source-mixing assumption. When $\kappa$ is small, the source coefficients ${b^{(l)}}$ are close to $\beta^*$, favoring traditional transfer learning methods. 
In Setting 2, we generate 4 heterogeneous source coefficients and define $\beta^*$ as a mixture of the first three sources plus a deviation in direction $\bold{a}$, scaled by $\kappa$:
$\beta^* = \frac{1}{3}(b^{(1)} + b^{(2)} + b^{(3)}) + \frac{\kappa}{\sqrt{35}} \bold{a}$.
Setting 3 mirrors Setting 1 but in a high-dimensional regime.
In all Settings 1-3, we vary $\kappa$ while keeping $\beta\subvalid^* = \beta^*$ fixed to assess GART’s robustness under increasing deviation from the source mixing prior.
In Setting 4, we use the same $\mathbf{B}$ and $\beta^*$ as in Setting 2 with $\kappa = 1.2$ and $2.4$, but evaluate performance on varying future target distributions.  Let $\D(\beta^*, \beta\subvalid) := (\beta^* - \beta\subvalid)\supT\Sigma\subQ(\beta^* - \beta\subvalid)$. In Setting 4.1, we shift $\beta\subvalid$ from $\beta^*$ toward the source convex hull under the constraint: $\beta\subvalid = \argmin_{\beta: \D(\beta^*, \beta)\leq \nu} \D(\beta, \beta\subsource)$ where $\beta\subsource = \BB\cdot\argmin_{\gamma\in\Delta_L}\D(\beta^*, \BB\gamma)$. In Setting 4.2, we construct $\beta\subvalid$ by randomly reweighting the sources and training target, subject to the distance constraint: $\beta\subvalid \in \{\beta = \BB_0\gamma\subvalid: \gamma\subvalid\sim \mbox{Dirichlet}(L_0, \mathbf{1}_{L_0}), \D(\beta^*, \beta)\leq \nu\}$ with $\nu$ varying from 0.1 to 0.5.  We assess GART’s robustness to different $\tau$ values under future target shifts of varying degree, indexed by $\nu$.
We further investigate how the choices of the initial estimators impact the GART performance in Appendix \ref{sec: appendix compare ini}. More details about the covariate distribution of the source, target and validation data are given in Appendix \ref{sec: appendix simulation desc}. 
\vspace{-5pt}
\subsection{Result}
\subsubsection{Robustness against source mixing assumption}
\begin{figure}[htbp]
    \centering
    \includegraphics[width=0.7\linewidth]{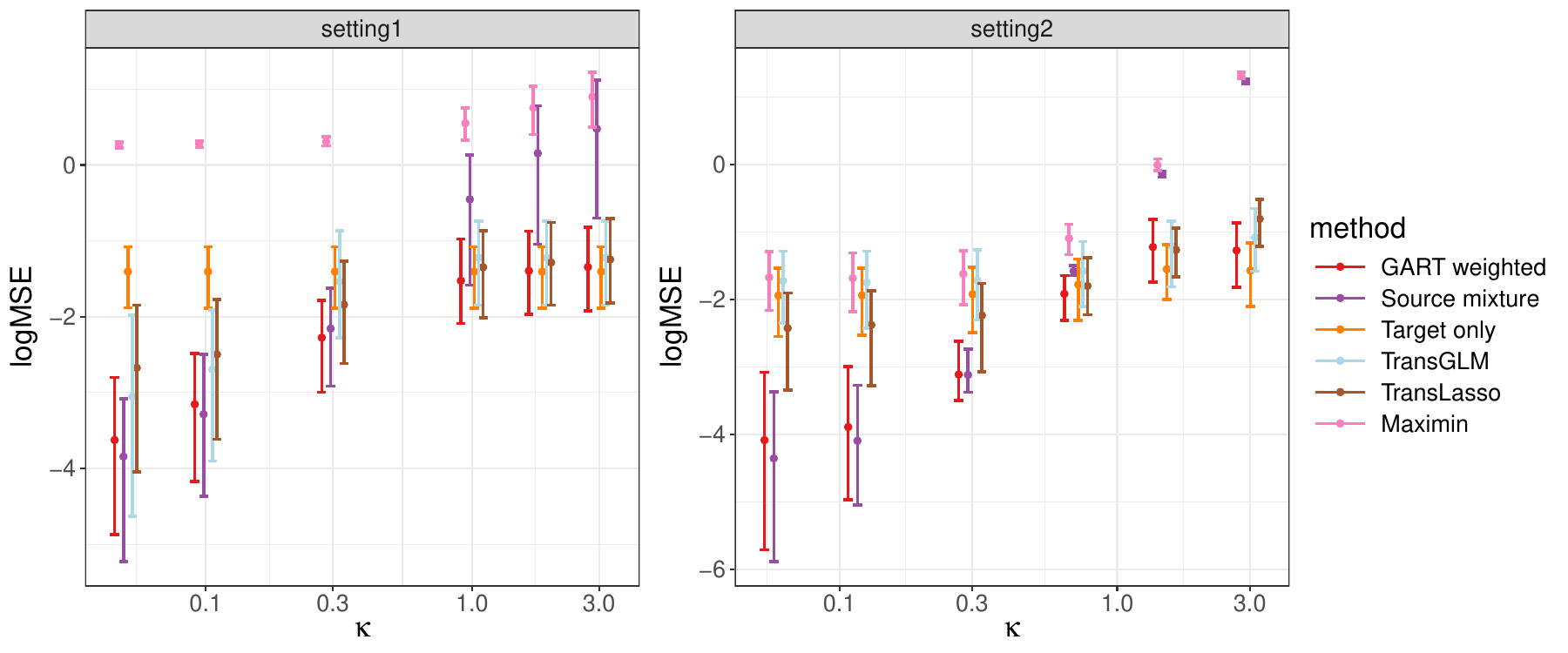}
    \caption{Empirical mean and 95\% confidence interval of $\log \text{MSE}$ when $n=200$ for setting 1 (left) and setting 2 (right). A growing $\kappa$ indicates a larger distance between each source and the target.}
    \label{fig: simulation setting1-2 new}
\end{figure}

Figure \ref{fig: simulation setting1-2 new} summarizes for settings 1 and 2 the mean of log of mean squared error (MSE) and its 95\% confidence interval over increasing $\kappa$ as the source mixing assumption gradually fail, with $\widehat{\beta}$ constructed based on different methods. In setting 1, when all source models resemble the target model (i.e., $\kappa\leq 0.1$), TransLasso and TransGLM exhibit a smaller MSE compared to the target-only estimator, indicating an effective transfer. Yet, the non-zero $\kappa$ indicates the presence of non-negligible heterogeneity between source and target distributions. This can lead the two transfer learning methods to exclude some useful source models from the transferable set, resulting in under-utilization of relevant source information compared to GART. Additionally, the performance gap may also reflect differences in constant factors in the convergence rates of GART, TransLasso and TransGLM, which can be more pronounced in finite-sample settings. When site heterogeneity grows (i.e., $\kappa\geq 0.3$), the performance of TransLasso and TransGLM deteriorates and plateaus slightly higher than the target-only estimator. MSE of the source mixture estimator even explodes when $\kappa\geq 1$, indicating a breakdown of the source mixing assumption. In contrast, GART maintains stable and controlled MSE across varying levels of $\kappa$, owing to its adaptability in leveraging source data selectively when the source mixing prior is valid. The variation of the GART MSE over 100 iterations is comparable to other methods, indicated by the similar CI length. Without the guidance of target labels, Maximin returns a model that is too conservative to work for the current target population even when the source mixing assumption holds.

In setting 2, all source models differ from the target model, leading to the poor performance of traditional transfer learning methods especially TransGLM that performs worse than the target-only estimator across all values of $\kappa$. On the other hand, when source mixing assumption holds approximately with smaller $\kappa$, both GART and source mixing perform well with substantial gain over the target only estimator. When $\kappa$ is large, the source mixing assumption completely fails and the GART estimator tends to resemble the target only estimator as expected while the source mixing estimator eventually performs worse than the target only. The Maximin estimator again performed poorly across all settings especially with larger $\kappa$. {The variation of model MSE does not vary a lot across different benchmarks, choices of $\kappa$ or $\nu$.} The findings from setting 3 indicate similar patterns of relative performance among various methods as in settings 1 and 2, detailed further in the Appendix \ref{sec: additional simu result}.

\subsubsection{Robustness of GART for future target populations}
\begin{figure}[t]
    \centering
    \includegraphics[width=.8\linewidth, page=1]{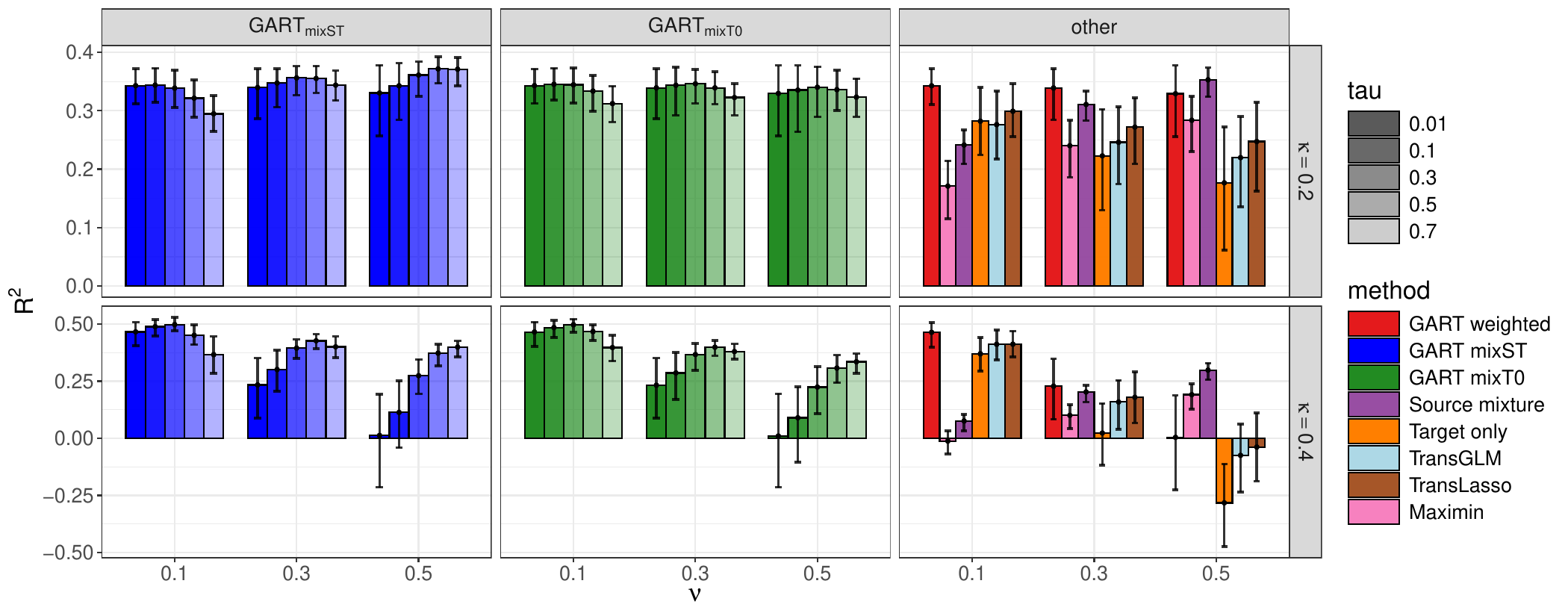}
    \caption{Empirical mean and 95\% confidence interval of $R^2$ of GART estimators and other benchmarks for settings 4 with $\kappa = 0.2$ and $\kappa = 0.4$ on different future target populations (indexed by $\nu$). GART is constructed using different $\tau$ and initial estimators: weighted, $\betahat\subinit\supmixST$ (mixST), $\betahat\subinit\supmixT0$ (mixT0).}
    \label{fig: simulation setting4 tau}
\end{figure}

We next assess the robustness of GART and compare to other estimators with respect to their performance in future target populations with $\beta^*\subvalid$ based on $R^2\subvalid = 1 - \E\{\|\YY\subvalid - \XX\subvalid\widehat{\beta}\|_2^2\}/\E\{\|\YY\subvalid - \E\YY\subvalid\|_2^2\}$. We consider different choices of $\tau \in (0,1)$ and three initial estimators for GART: i) $\betahat\subinit\supweighted$ in \eqref{eq: def of weighted init beta}; ii) $\betahat\supmixST\subinit$ in \eqref{eq: def of mixST init beta} with $\omega = \tau$ and iii) $\betahat\subinit\supmixT0$ with $\omega = \tau$ in \eqref{eq: def of mixT0 init beta}. GART with $\betahat\supweighted$ as initial varies little with $\tau$, so we only present GART with $\betahat\supweighted$ as initial when $\tau = 0.01$. 
Figure \ref{fig: simulation setting4 tau} shows results for setting 4.1, where $\beta^*$ is moderately ($\kappa = 0.2$) or significantly ($\kappa = 0.4$) away from the source convex hull and $\beta\subvalid$ deviates further from $\beta^*$ as $\nu$ increases. When $\kappa=0.2$, GART with $\betahat\supweighted$ as initial attained robust performance with high  $R^2\subvalid$ even if $\beta\subvalid$ deviates from $\beta^*$. GART with $\betahat\subinit\supmixST$ defined in \eqref{eq: def of mixST init beta} or $\betahat\subinit\supmixT0$ defined in \eqref{eq: def of mixT0 init beta} as initials attained comparable performance although choosing a slightly smaller $\tau$ appears to benefit those future target populations more similar to the observed target. In contrast, the target-only estimator trained for $\beta^*$ and the two transfer learning models suffer from a deteriorating $R^2\subvalid$ along with a larger $\nu$, suggesting a poor generalizability compared to GART. When $\kappa=0.4$ with $\beta^*$ further away from the sources, source mixing assumption does not hold for $\beta^*$ while the future target populations are close to the source convex hull. GART with $\betahat\subinit\supmixST$ or $\betahat\subinit\supmixT0$ trained using larger $\tau$ attained more robust performance as $\nu$ increases.  Such improved generalizability comes at the cost of a slightly worse $R^2$ when $\nu$ is small with $\beta\subvalid$ relatively close to $\beta^*$. Results for setting 4.2 show similar patterns except that both maximin and source mixture performed substantially worse than GART estimators with $\betahat\subinit\supmixST$ or $\betahat\subinit\supmixT0$  as initial estimators, as shown in Figure \ref{fig: simulation initial 5} in Appendix. These findings confirm GART's robustness in future target populations that diverge from the observed target. We have also included additional empirical comparisons with three multitask learning methods \citep{huang2025optimal, duan2023adaptive, tian2025learningsimilarlinearrepresentations} especially under settings with small $\ell_1$ discrepancy and nearly zero $\ell_0$ difference, where the multitask methods perform comparably to GART. However, under all other settings considered, they perform substantially worse than GART. See Section \ref{sec: multi-task} in Supplementary Materials for more details.

\section{Application of GART to EHR Biobank Data}

\subsection{Background}
The levels of high density lipoprotein cholesterol (HDL) in the plasma, which are negatively associated with the risk of coronary artery disease, are strongly influenced by genetics, with heritability estimates ranging between 40-60\% \citep{weissglas2010genetic}. The majority of participants in existing genetic studies on HDL are white, leading to little evidence on the effect of genetic markers on HDL for non-white populations \citep{boes2009genetic}. Studies using racially diverse populations, like the Multi-Ethnic Study of Atherosclerosis, have uncovered variations in genetic associations among different racial groups \cite{frazier2013genetic}. However, non-white racial minority groups often have smaller sample sizes, which limits the ability to detect associations and generate precise polygenic risk scores (PRS) for these minority groups. This issue is even more pronounced for individuals of mixed racial backgrounds, a group that is less frequently studied in the existing literature \citep{phillips2007mixed}. We propose to adopt the GART approach to derive accurate HDL polygenic risk scores (PRS) for minority multi-racial subgroups using data from the UK Biobank (UKB) and further externally validated using EHR linked Biobank data from Mass General Brigham (MGB). 

We partition the UKB cohort into different subgroups according to participants' reported race as source populations: 400746 White, 6625 Black, 9440 Asian, and 3802 Others. In addition to these main subgroups, there are three mixed-race groups: 875 White-Black (WB), 701 White-Asian (WA), and 900 reported as Mixed or unknown (MIX). Previous studies have found that mixing source populations with different racial backgrounds can generate PRS more effective for admixed populations, suggesting that source mixing prior is appropriate in this context \citep{marquez2017multiethnic}. Our goal is to develop precise HDL PRS for multi-racial subjects using GART and the source mixing prior. We focus on each of the three mixed-race subgroups individually as the target population. For each subgroup, we randomly select {50 or 100} observations to serve as the target dataset along with the 4 source datasets for training GART, with the remaining target data used for validation. {This small training-to-validation ratio is designed to reflect scenarios where labeled target data is limited. Results for $n=50$ can be found in Section \ref{sec: more real data analysis} in Supplementary Materials.} We focus on 195 SNPs previously reported in \citet{frazier2013genetic} as potentially associated with HDL and construct a PRS for fasting mean HDL diameters, adjusting for age and sex. For each trained model, we evaluate the prediction performance of the estimated PRS based on the percent reduction in prediction error, denoted by $\DRE$, compared to the model trained only using the target observations. To validate model robustness in future potential target populations, we further evaluate the generalizability of the trained models, denoted by GART$\subUKBWB$, GART$\subUKBWA$, and GART$\subUKBMIX$, trained with the 4 main UKB source datasets along with either UKB WB, UKB WA, or UKB MIX as the target data. Specifically, for each of the trained model, we report its prediction performance on the other two admixed populations in UKB as well as in MGB Biboank sub-populations: White ($n=33,017$), Black ($n=2213$), and Asian ($n=900$). {To show the robustness of model performance, we iterate the random selection of training target 50 times (each with 100 subsamples) and report the mean and 95\% empirical interval of $\log\text{1/MSE}$.} We compare GART performance with existing methods including TransLasso, TransGLM, and Maximin. { $\tau$ is set as  $\frac{L\widehat{\sigma}^2\subQ}{n}$ for all the real data analysis.}

\subsection{Derived Models}
\begin{figure}[t]
    \centering
    \includegraphics[width=.85\linewidth, page=2]{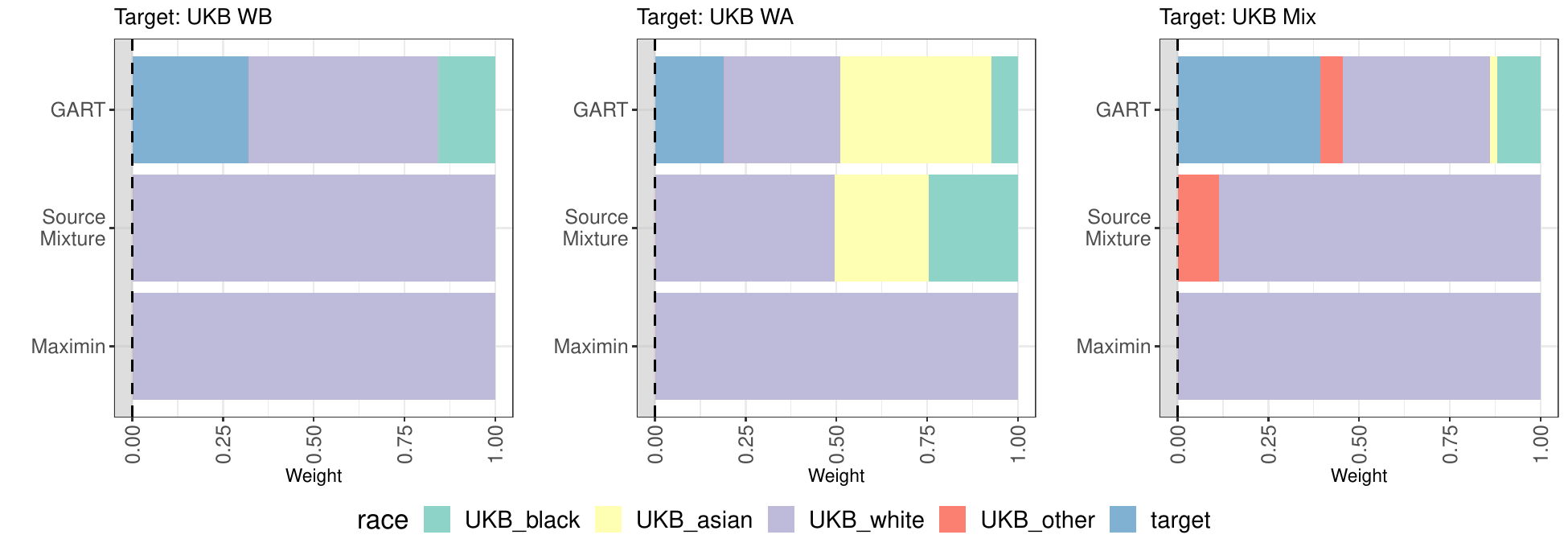}
    \caption{Weight assigned to each UKB source site when trained with different racial groups as the target site (left: white-black; middle: white-asian; right: mixed).}
    \label{fig: weight plot}
\end{figure}

In Figure \ref{fig: weight plot}, we present the mean weights assigned to the source models and the target only model by different methods. Interestingly, both GART$\subUKBWB$ and GART$\subUKBWA$ appropriately assigned comparable weights to the source models trained with the corresponding racial subgroups, consistent with our expectations from prior knowledge. Without guidance from the target label data, Maximin assigns all weights to UKB White. The optimal linear combination of source models assign all weights to the UKB White model when target is UKB WB; and a mixture of White, Asian, and Black when the target is UKB WA. When UKB MIX is the target, the source mixture model assigns around 90\% of weights to UKB White, and the rest all goes to UKB Others. On the other hand, GART$\subUKBMIX$ distributes the weights more evenly across multiple sources. 

\subsubsection{Prediction Performance and Robustness}
\begin{figure}[h]
    \centering
    \includegraphics[width=1\linewidth]{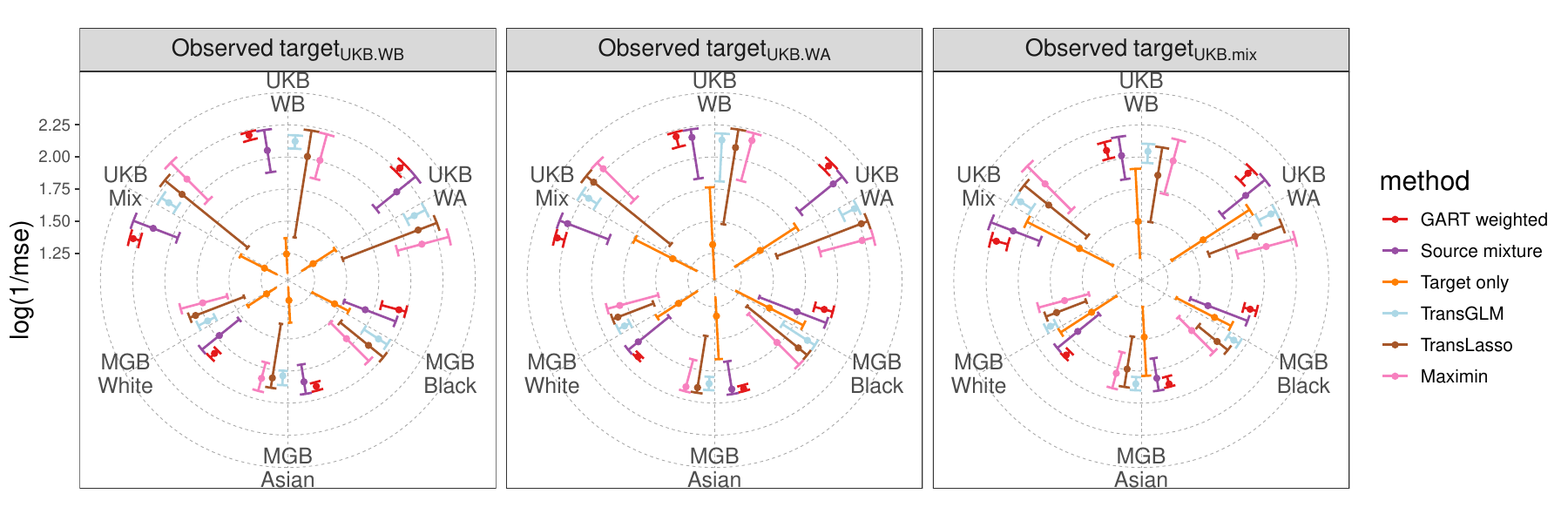}
    \caption{Prediction error $\log (1/\text{mse})$ on different validation populations when trained with UKB WB, UKB WA, or UKB WB being the target population and $n=100$.}
    \label{fig: radar}
\end{figure}

Figure \ref{fig: radar} shows the prediction performance results of GART$\subUKBWB$, GART$\subUKBWA$, and GART$\subUKBMIX$ on the observed target populations in UKB along with other future target populations including MGB sub-populations compared with other competing methods. GART models generally attained the most robust performance when we vary the training target population and future target populations. Compared to target only models, GART-trained models attained a reduction in prediction error, $\DRE$, ranging from 61.1\% to 65.9\%. TransGLM and TransLasso also reduce the prediction errors to a certain extent, but less efficiently compared to the Source Mixture and Maximin when leveraging the source information, especially when validating on MGB. Such a performance gap is as expected, since the target population may possess a certain amount of site heterogeneity compared to the four source sites, making it hard for traditional transfer learning algorithms to detect `transferrable' sources. Additionally, GART estimators attained substantially better generalizability when validated against other target populations, compared to other methods. For example, when the training target population is UKB WB and the validation population switches to MGB White and MGB Black, GART attained $\DRE = 53.0\%$ and $38.3\%$ while the second best approach, Maximin, attained $\DRE = 41.3\%$ and $12.7\%$ and $\DRE = 38.9\%$ and $29.1\%$ for the TransLasso. The higher mean value and the narrower interval of the $-\log\text{MSE}$ of GART compared to the target-only estimator and other benchmarks indicate more accurate prediction and more stable performance over iterations. 

\section{Discussion}
In this paper, we introduce GART which serves as a bridge connecting model transferability with observed target data and generalizability to unobserved different future target populations. Compared to existing transfer learning methods, GART can more flexibly leverage heterogeneous source data that may differ from the target data via source mixing prior. 

One key assumption GART relies on is the closeness between the current target and the source convex hull, which is much weaker than that in classical transfer learning models. While methods like TransLasso and TransGLM perform well when some sources closely match the target, source mixing is often more appropriate in practice. In healthcare, for example, different systems serve diverse populations, making a mixture of sources more representative of a given target group. Similarly, in genetic prediction for individuals of mixed race, genetic profiles often align better with a combination of ancestries, as illustrated in our real data example.
A key strength of GART is its generalizability: by leveraging source mixing, it defines a broader set of potential future targets, enabling better adaptation to populations that differ from the current one.

However, such a source-mixing condition also makes a sharp minimax lower bound for the GART parameter space technically nontrivial. The main challenge is that the mixing condition measures the distance between the target parameter and the convex hull of the source parameters under a Mahalanobis metric. Unlike the coordinate-wise $\ell_1$-type similarity conditions used in transLASSO, this constraint is non-decomposable across coordinates and depends on the joint geometry of all source models. As a result, standard information-theoretic lower-bound arguments do not directly apply and may be quite loose in this setting.
Specifically, since the transLASSO similarity class is contained in the GART source-mixing class, the existing transLASSO minimax lower bound can serve as a valid, although generally loose, lower bound for GART. Obtaining a tight lower bound tailored to the GART parameter space would require new techniques and remains an interesting direction for future work.

GART relies on $\ell_1$ similarity and is not currently designed to leverage $\ell_0$-similarity, which has been explicitly considered in recent multi-task learning methods \citep{xu2025multitask, huang2025optimal}. These models assume an $l_0$ sparse mismatch between target and all sources (i.e., $\|\beta_0-\beta_l\|_0\leq s$ for all $l\in[L]$), and construct robust coordinate-wise aggregation around a shared central parameter, which is then used to shrink the local target estimator. Consequently, their rates depend explicitly on the $l_0$ sparsity level. In contrast, when the mismatched coordinates cannot be absorbed by the source-mixture structure, GART falls back to the target-only benchmark despite the shared structure on the $l_0$ level. The $\ell_0$ similarity naturally motivates a partial-mixing perspective: even if global mixing fails, there may exist a large subset of coordinates on which the source mixture well approximates the target well, while the remaining coordinates require target-specific estimation. The key objective is to identify a mixable subset that captures most of the predictive signal, enabling effective transfer on this subspace while minimizing residual variability that must be learned from the target alone. Principled approaches for incorporating $\ell_0$-level similarity information into the GART framework remains an important direction for further research.

Another scenario in which the global source-mixing assumption may fail is when only part of the model satisfies the mixture condition (partial-mixing assumption). For example, only the first $p_0$ elements of $\beta^*$ satisfy that $\beta^*_{1:p_0}\approx \BB_{1:p_0,}\gamma^*$, while the rest of the coefficients cannot be well approximated by the same source mixture. Currently, GART can be built on the first $p_0$ dimension of feature space and estimate $\beta^*_{p_0+1:p}$ by the target-only model, if there is prior knowledge about those non-mixing dimensions (e.g., the intercept term). When no prior information is available, additional model selection before implementing GART might be beneficial to identify the true partial mixture, instead of making the global mixture driven by the target's non-mixing dimension.

Although GART was developed within a linear model framework, it can be naturally extended to non-linear settings. This extension builds on recent DRO research that leverages machine learning algorithms, such as random forests and neural networks, to approximate the target conditional distribution  \citep{wang2023DRL}. By incorporating a target data-driven constraint set and an initial estimator into their loss function, GART can be seamlessly adapted into a privacy-preserving distributionally robust framework. This adaptation enables the use of machine learning estimators to account for non-linear effects.

\section*{Disclosure statement}
The authors report there are no competing interests to declare.

\bibliographystyle{abbrvnat}
\begingroup
\setstretch{1.3}
\bibliography{main}
\endgroup

\newpage

\appendix
\setcounter{page}{1}
\setcounter{table}{0}
\setcounter{figure}{0}

\begin{center}
{\Large \bf Guided Adversarial Robust Transfer Learning with Source Mixing - Supplementary Materials}
\end{center}
\setstretch{1.65} 

\section{Algorithm Details}
\begin{algorithm}[H]
\caption{Guided Adversarial Robust Transfer Learning (GART)}\label{alg:gart}
\label{alg: GART estimation}
\KwData{Source data $(Y\supl, X\supl)_{l=1,...,L}$, target data $(Y\supQ, X\supQ)$, the tolerance parameter $\tau$, one initial model estimators $\widehat{\beta}\subinit$.}
\KwResult{$\widehat{\beta}\subGART$}
 \textbf{Step 1} (estimate the target and source models locally): \\
\hspace*{5mm} \For{$l=1$ \KwTo $L$}
{\hspace*{5mm} Estimate $\widehat{b}^{(l)}$ by Lasso with data $(Y\supl, X\supl)$; \\
\hspace*{5mm} Upload $\widehat{b}^{(l)}$ to the target site;}
 \hspace*{5mm}Split the target data into $(\YY\suba\supQ, \XX\suba\supQ)$ and $(\YY\subaa\supQ, \XX\subaa\supQ)$\; 
 \hspace*{5mm}Construct $\widehat{\beta}\subaj$ using data $(\YY\subaj\supQ, \XX\subaj\supQ)$ as in \eqref{eq: estimate b0} for $j=1,2$; $\widehat{\sigma}\subQ^2$ as in \eqref{eq: est for sigmaQ, 1} and $\widehat{\Sigma}\subQ = \frac{1}{n}\sum_{i=1}^n X_i\supQ (X_i\supQ)\supT$ \;

 \textbf{Step 2} (estimate $\mathcal{S}(\tau)$): \\
 \hspace*{5mm}With $\{\widehat{b}\supl\}_{l=1,...,L}$ and $(\YY\subaa\supQ, \XX\subaa\supQ)$, construct $\widehat{\sigma}\subsource^2$ as in \eqref{eq: est for sigmaQ, 2}\;
  \hspace*{5mm}With $\widehat{\sigma}\subQ^2$, $\widehat{\sigma}\subsource^2$, $\tau$ and $(\YY\subaa\supQ, \XX\subaa\supQ)$,  construct $\widehat{\mathcal{S}}(\tau)$ as in \eqref{eq: constraint d}\;

 \textbf{Step 3} (estimate $\beta\subGART$): \\
 \hspace*{5mm}Plug in $\widehat{\mathcal{S}}(\tau)$ and $\widehat{\beta}\subinit$ to \eqref{eq: sample version of GART} and obtain the GART estimator $\widehat{\beta}\subGART$.
\begin{equation}
\widehat{\beta}\subGART  = \BBhat_0\widehat{\gamma}\subGART \quad \text{with}\quad \widehat{\gamma}\subGART=\argmin_{\gamma\in \widehat{\mathcal{S}}(\tau)} \gamma\trans\bGammahat_0\gamma 
\label{eq: sample version of GART}
\end{equation}
where $\bGammahat_0=[\BBhat_0- \BBhat\subinit]\trans\widehat{\Sigma}_{\Tar}[\BBhat_0- \BBhat\subinit]$,  $\BBhat_0=[\widehat{\beta}\suba, \widehat{b}^{(1)},...,\widehat{b}^{(L)}]$, $\BBhat\subinit = [\widehat{\beta}\subinit, ..., \widehat{\beta}\subinit]_{p\times L_0}$. 
\end{algorithm}

\newpage

\section{Proofs}
We present the proofs of Theorem \ref{thm:eq1}, \ref{thm: consistency}, \ref{thm: first guidance} and \ref{thm: second guidance} in Sections \ref{sec: thm1}, \ref{sec: proof for thm consistency}, \ref{sec: proof for thm rate} and \ref{sec: proof for thm initial}, respectively. Proofs for related Lemmas and Corollaries have been placed in Section \ref{sec: proof for lemmas}.

\subsection{Proof of Theorem \ref{thm:eq1}}
\label{sec: thm1}
We generalize the proof strategy in \cite{Meinshausen_2015}. However, our proof is more involved since we incorporated different initial estimators and the GART constraint set.

Recall the risk function $\RR_{\T}(\beta, \beta\subinit)$ defined in \eqref{eq: linear GART loss}. We expand the expectation under the linear model assumption \eqref{eq: target model} and rewrite the risk as:
\begin{equation}
\begin{aligned}
\RR_{\T}(\beta, \beta\subinit) &= \E_{X\sim\Tar\subX}\left[\left(X\supT \BB_0\gamma + \epsilon - X\trans\beta\right)^2 - \left(X\supT \BB_0\gamma + \epsilon - X\trans\beta\subinit\right)^2\right]\\
&= \beta\trans\Sigma_{\Tar}\beta + 2\left(\beta\subinit - \beta\right)\trans\Sigma_{\Tar}\BB_0\gamma,
\end{aligned}
\end{equation}
where $B_0$ is defined in \eqref{eq: sample version of GART}. Therefore, GART has another form of expression:
\begin{equation}
\begin{aligned}
     \beta\subGART(\beta\subinit, \tau)= \argmin_{\beta\in\mathbb{R}^p} \max_{\gamma\in\mathcal{S}(\tau)}  \left[\beta\trans\Sigma_{\Tar}\beta + 2\left(\beta\subinit - \beta\right)\trans\Sigma_{\Tar}\BB_0\gamma\right].
\end{aligned}
\end{equation}
The constraint set $\mathcal{S}(\tau)$ can also be simplified to $\{\gamma: (\BB_0\gamma-\beta^*)\trans\Sigma_{\Tar}(\BB_0\gamma-\beta^*)\leq \tau\}$, given the the linear model assumption \eqref{eq: target model} and $\E[\epsilon\supQ|X\supQ]=0$.

Let $C^TC = \Sigma_{\Tar}$ be the Cholesky decomposition of $\Sigma_{\Tar}$. Since we assume $\Sigma_{\Tar}$ to be full rank in assumption \ref{assumption: a_1}, $C$ is invertible. Further, define: $\mathcal{U}_{\tau} := \{C\BB_0\gamma: \gamma\in\mathcal{S}(\tau)\}\subset\R^p$. We can then re-write $\beta\subGART$ as: 
\begin{equation}
    \begin{aligned}
    \beta\subGART &:=   \argmin_{\beta\in\mathbb{R}^p} \max_{\gamma\in\mathcal{S}(\tau)} \left\{\beta\trans\Sigma_{\Tar}\beta + 2\left(\beta\subinit - \beta\right)\trans\Sigma_{\Tar}\BB_0\gamma\right\} \\ 
    &= \argmax_{\beta\in\mathbb{R}^p} \min_{\gamma\in\mathcal{S}(\tau)} \left\{ - \beta\trans\Sigma_{\Tar}\beta - 2\left(\beta\subinit - \beta\right)\trans\Sigma_{\Tar}\BB_0\gamma\right\} \\
    &= C^{-1} \argmax_{\xi_0\in\mathbb{R}^p} \min_{u_0\in\mathcal{U}_{\tau}} \left[-\xi_0\trans\xi_0 + 2\xi_0\trans u_0 - 2\xi\subinit\trans u_0\right] \\
    &= C^{-1} \argmax_{\xi_0\in\mathbb{R}^p} \min_{u_0\in\mathcal{U}_{\tau}} \left[-(\xi_0 - \xi\subinit)\trans(\xi_0 - \xi\subinit) + 2(\xi_0-\xi\subinit)\trans(u_0 - \xi\subinit)\right] 
    \end{aligned}
    \label{eq: thm1 cond1}
\end{equation}
where $\xi\subinit := C\beta\subinit$. 
By defining $\xi = \xi_0 - \xi\subinit$, $u = u_0 - \xi\subinit$, and $\mathcal{U}_{\tau}^{\mathrm{shift}} := \{u_0 - \xi\subinit: u_0\in\mathcal{U}_{\tau}\}$, which is still a convex set, \eqref{eq: thm1 cond1} can be further re-written as:
\[
\beta\subGART = C^{-1} \argmax_{\xi\in\mathbb{R}^p} \min_{u\in\mathcal{U}_{\tau}^{\mathrm{shift}}} \left[-\xi\trans\xi + 2\xi\trans u\right] 
\]

If we can show the `max' and `min' can be exchanged, then the problem is easily solved by:
\begin{align*}
\beta\subGART &= C^{-1}\argmin_{u\in\mathcal{U}_{\tau}^{\mathrm{shift}}} \max_{\xi\in\mathbb{R}^p} \left[-\xi\trans\xi + 2\xi\trans u\right] \\
&= C^{-1}\argmin_{u\in\mathcal{U}_{\tau}^{\mathrm{shift}}}  \|u\|_2^2 \\
&= \BB_0\cdot\argmin_{\gamma\in\mathcal{S}(\tau)} (\BB_0\gamma - \beta\subinit)\trans \Sigma_{\Tar} (\BB_0\gamma - \beta\subinit)
\end{align*}

The second equation holds as the maximum is achieved at $\xi=u$. Therefore,  the problem becomes to show:
\begin{equation}
\begin{aligned}
\xi^* =\argmin_{u\in\mathcal{U}_{\tau}^{\mathrm{shift}}} \|u\|_2^2   =  \argmax_{\xi\in\mathbb{R}^p} \min_{u\in\mathcal{U}_{\tau}^\mathrm{shift}} \left[-\xi\trans\xi + 2\xi\trans u\right],
\end{aligned}
 \label{eq: exchange-max-min}
\end{equation}
the proof of which can be found in theorem 1 in \cite{Meinshausen_2015}. To be self-contained, we provide the proof here.

Given any $\mu\in\mathcal{U}_{\tau}^{\mathrm{shift}}$, let $g(\nu) = \|\xi^* + \nu(\mu - \xi^*)\|_2^2$ where $\nu\in[0,1]$. Since $\mathcal{U}_{\tau}^{\mathrm{shift}}$ is a convex set and $\xi^*$ by definition is inside $\mathcal{U}_{\tau}$, $\xi^* + \nu(\mu - \xi^*)$ must also lie inside $\mathcal{U}_{\tau}^{\mathrm{shift}}$ and
$$g(\nu) \geq \|\xi^*\|_2^2 \text{ for any } \nu\in[0,1].
$$
Because $g(0) = \|\xi^*\|_2^2$ and $g(\cdot)$ is continuous on $[0,1]$, we have:
\begin{align*}
 & g^{\prime}(0) = 2(\mu - \xi^*)\trans\xi^* \geq 0
 \iff  2\mu\xi^* - (\xi^*)\trans\xi^* \geq (\xi^*)\trans\xi^* \text{ for any }\mu\in\mathcal{U}_{\tau}.
\end{align*}
If we take $\xi = \xi^*$, then the right-hand side of \eqref{eq: exchange-max-min} has:
$$
\max_{\xi\in\R^p} \min_{u\in\mathcal{U}_\tau^{\mathrm{shift}}}  2\xi\trans u - \xi\trans\xi \geq \min_{u\in\mathcal{U}_\tau^{\mathrm{shift}}}  2(\xi^*)\trans u - (\xi^*)\trans\xi^* \geq (\xi^*)\trans\xi^*.
$$
On the other hand, if we take $u = \xi^*$:
$$
\max_{\xi\in\R^p} \min_{u\in\mathcal{U}_\tau^{\mathrm{shift}}}  2\xi\trans u - \xi\trans\xi \leq \max_{\xi\in\R^p} 2\xi\trans\xi^* - \xi\trans\xi \leq (\xi^*)\trans\xi^*.
$$
The second equation only holds if $\xi=\xi^*$. In conclusion, $\xi^* = \argmax_{\xi\in\R^p} \min_{u\in\mathcal{U}_\tau^{\mathrm{shift}}}  2\xi\trans u - \xi\trans\xi$.

\subsection{Proof of Theorem \ref{thm: consistency}}
\label{sec: proof for thm consistency}
\begin{proof}
For this proof we assume the invertibility of $\bGamma_0$ and fixed site number $L$ for simplicity. Our proof can be extended to cover a more general setting. 

When $\tau\to 0$, we have shown that $\beta\subGART \to \beta^*$ in Proposition \ref{prop: population convergence} and $\widehat{\beta}\subGART \overset{p}{\to} \beta^*$ in Theorem \ref{thm: first guidance}, which implies that $\widehat{\beta}\subGART \overset{p}{\to} \beta\subGART$. The following proof focuses on the case when $\tau$ is a fixed positive constant.

For notation simplicity, we rewrite  $\gamma\subGART$ in \eqref{eq: g-maximin step 2} and $\widehat{\gamma}\subGART$ in \eqref{eq: sample version of GART} as $\gamma_{\mathrm{G}}$ and $\widehat{\gamma}$. Additionally, we define 
    \begin{equation}
       g(\gamma) = \gamma\supT\bGamma_0\gamma,\,\,\, \widetilde{\gamma} = \argmin_{\gamma\in \widehat{\mathcal{S}}(\tau)} g(\gamma),\,\,\, \widebar{\gamma} = \argmin_{\gamma\in \mathcal{S}(\tau) \cap \widehat{\mathcal{S}}(\tau)} g(\gamma).
        \label{eq: thm 2 gamma def}
    \end{equation}

The following Lemma first aims to show that 
under an identical convex constraint set for $\gamma$, the distance between the minimizer of $\gamma\supT\bGamma_0\gamma$ and that of $\gamma\supT\bGammahat_0\gamma$ is well controlled.
    \begin{Lemma}
        For the convex subset $\mathcal{H}$ of $\Delta_{L_0}$, given the semi-positive definite matrix $\bGamma_0$, $\widehat{\bGamma}_0$, we define $q^* = \argmin_{q\in\mathcal{H}} q\trans\bGamma_0 q$ and $\widehat{q} = \argmin_{q\in\mathcal{H}}q\trans\widehat{\bGamma}_0q$. If $\lambda_{\min}(\bGamma_0)>0$, then
        \begin{equation}
            \|\widehat{q} - q^*\|_2 \leq \frac{L_0\|\widehat{\bGamma}_0 - \bGamma_0\|_2}{\lambda_{\min}(\bGamma_0)}.
        \label{eq: thm2 condition1}
        \end{equation}
    \end{Lemma}
    The detailed proof can be found in Lemma 6 of \cite{maximin-infer}. By taking $\mathcal{H} = \widehat{\mathcal{S}}(\tau)$, we derive that $\|\widehat{\gamma} - \widetilde{\gamma}\|_2$ is controlled by the right hand side of \eqref{eq: thm2 condition1}. Since $\widehat{\bGamma}_0$ is a plug-in estimator containing all consistent estimators (i.e., $\widehat{b}^{(1)},...,\widehat{b}^{(L)}$, $\widehat{\beta}^*$, $\widehat{\Sigma}_{\Tar}$), $\|\widehat{\bGamma}_0 - \bGamma_0\|_2$ will converge to 0 in probability, which implies $\|\widehat{\gamma} - \widetilde{\gamma}\|_2 \overset{p}{\to} 0$. Next, we will prove that $\|\widetilde{\gamma} - \widebar{\gamma}\|_2 \overset{p}{\to} 0$ and $\|\gamma_{\mathrm{G}} - \widebar{\gamma}\|_2 \overset{p}{\to} 0$, which closes the proof given that
    \[\|\widehat{\gamma} - \gamma_{\mathrm{G}}\|_2 \leq \|\widehat{\gamma} - \widetilde{\gamma}\|_2 + \|\widetilde{\gamma} - \widebar{\gamma}\|_2 + \|\widebar{\gamma} - \gamma_{\mathrm{G}}\|_2.\]

We define two convex functions $f(\gamma)$ and $\widehat{f}_n(\gamma)$ of $\gamma$ for any $\gamma\in\Delta_{L_0}$:
    \begin{equation}
        f(\gamma) = \E\left[(Y_i\supQ - (X_i\supQ)\trans\BB_0\gamma)^2\right] - \sigma_{\Tar}^2 - \tau \text{, and}
        \label{eq: def of fgamma}
    \end{equation}    
    \begin{equation}
        \widehat{f}_{n}(\gamma) = n_2^{-1}\|\YY\subaa\supQ - \XX\subaa\supQ\BBhat_0\gamma\|_2^2 - \min\{\widehat{\sigma}_{\Tar}^2, \widehat{\sigma}\subsource^2\} - \tau.
        \label{eq: def of fngamma}
    \end{equation}
 Note that $\gamma\in \mathcal{S}(\tau)$ is equivalent to $f(\gamma)\leq 0$ and $\gamma\in \widehat{\mathcal{S}}(\tau)$ is equivalent to $\widehat{f}_n(\gamma)\leq 0$.
 The following lemma illustrates that $\widehat{f}_{n}(\gamma)$ converges uniformly to $f(\gamma)$ in probability, where the proof can be found in Section \ref{sec: proof for lemmas}.
\begin{Lemma} For a fixed $L_0=L+1$, we have
  \[
  \sup_{\gamma\in\Delta_{L_0}} \bigg|\widehat{f}_n(\gamma) - f(\gamma)\bigg|  \overset{p}{\to} 0,
  \]
  where $f(\gamma)$ and $\widehat{f}_n(\gamma)$ are defined in \eqref{eq: def of fgamma} and \eqref{eq: def of fngamma}, respectively.
  \label{lemma: consistency them 1}
\end{Lemma}

With the uniform convergence for the function defined by the constraint set, we then show in the following lemma the convergence of the objective function $g(\cdot)$ evaluated at $\widetilde{\gamma}$, $\widebar{\gamma}$ and $\gamma_{\mathrm{G}}$ before proving the convergence of $\widetilde{\gamma} -\widebar{\gamma}$ and $\gamma_{\mathrm{G}} - \widebar{\gamma}$. 

\begin{Lemma} For the convex function $g(\cdot)$, $\widetilde{\gamma}$ and $\widebar{\gamma}$ defined in \eqref{eq: thm 2 gamma def}, and $\gamma_{\mathrm{G}}$ (i.e., $\gamma\subGART$) defined in \eqref{eq: g-maximin step 2}, $g(\widetilde{\gamma}) - g(\widebar{\gamma}) \overset{p}{\to} 0; \; g(\gamma_{\mathrm{G}}) - g(\widebar{\gamma}) \overset{p}{\to} 0$.
\label{lemma: thm2 assist 2}
\end{Lemma}

Recall that both $\widetilde{\gamma}$ and $\widebar{\gamma}$ belong to $\widehat{\mathcal{S}}(\tau)$. Since $\widehat{\mathcal{S}}(\tau)$ is a convex constraint set, and $\widetilde{\gamma}$ is the minimizer of $g(\gamma)$ over all $\gamma \in \widehat{\mathcal{S}}(\tau)$, we have that for all $t\in(0,1)$, $(1-t)\widetilde{\gamma} + t\widebar{\gamma} \in \widehat{\mathcal{S}}(\tau)$ and
\begin{equation*}
    \begin{aligned}
&\widetilde{\gamma}\supT\bGamma_0\widetilde{\gamma} \leq ((1-t)\widetilde{\gamma} + t\widebar{\gamma})\supT\bGamma_0((1-t)\widetilde{\gamma} + t\widebar{\gamma}) \\
&\iff t^2 (\widebar{\gamma} - \widetilde{\gamma})\supT\bGamma_0(\widebar{\gamma} - \widetilde{\gamma}) + 2t\widetilde{\gamma}\supT\bGamma_0(\widebar{\gamma} - \widetilde{\gamma}) \geq 0.
    \end{aligned}
\end{equation*}

By letting $t\to 0$, we derive that $\widetilde{\gamma}\supT\bGamma_0(\widebar{\gamma} - \widetilde{\gamma}) \geq 0$. We then write out $g(\widetilde{\gamma}) - g(\widebar{\gamma})$ as:
\begin{equation}
    \begin{aligned}
    g(\widetilde{\gamma}) - g(\widebar{\gamma}) &= \widetilde{\gamma}\supT\bGamma_0\widetilde{\gamma} - \widebar{\gamma}\supT\bGamma_0\widebar{\gamma} \\
    &= \widetilde{\gamma}\supT\bGamma_0\widetilde{\gamma} - (\widebar{\gamma} - \widetilde{\gamma} + \widetilde{\gamma})\supT\bGamma_0(\widebar{\gamma} - \widetilde{\gamma} + \widetilde{\gamma}) \\
    &=  -(\widebar{\gamma} - \widetilde{\gamma})\supT\bGamma_0(\widebar{\gamma} - \widetilde{\gamma}) -2\widetilde{\gamma}\supT\bGamma_0(\widebar{\gamma} - \widetilde{\gamma}).
    \end{aligned}
    \label{eq: thm2 cond2}
\end{equation}
Since $0 > g(\widetilde{\gamma}) - g(\widebar{\gamma}) \overset{p}{\to} 0$ shown in Lemma \ref{lemma: thm2 assist 2}, there exists $N>0$ such that for all $n>N$, $g(\widetilde{\gamma}) - g(\widebar{\gamma}) \geq -\frac{1}{n}$ with high probability. Therefore, combining \eqref{eq: thm2 cond2} with $\widetilde{\gamma}\supT\bGamma_0(\widebar{\gamma} - \widetilde{\gamma}) \geq 0$,
\[
(\widebar{\gamma} - \widetilde{\gamma})\supT\bGamma_0(\widebar{\gamma} - \widetilde{\gamma})  \leq -2\widetilde{\gamma}\supT\bGamma_0(\widebar{\gamma} - \widetilde{\gamma}) + \frac{1}{n} \leq \frac{1}{n},
\]
which implies $\|\widebar{\gamma} - \widetilde{\gamma}\|_2 \leq \frac{1}{n \lambda_{\min}(\bGamma_0)} \to 0$. 

A similar proof applies to$\| \widebar{\gamma} - \gamma_{\mathrm{G}} \|_2$ by treating $\gamma_{\mathrm{G}}$ as $\widetilde{\gamma}$ and switching the role of $\widehat{f}_n(\cdot)$ (i.e., $\widehat{\mathcal{S}}(\tau)$) and $f(\cdot)$ (i.e., $\mathcal{S}(\tau)$).

\end{proof}

\subsection{Proof of Theorem \ref{thm: first guidance}}
\label{sec: proof for thm rate}
We introduce the following events to facilitate the proofs:
\begin{equation*}
    \begin{aligned}
        G_1 &= \left\{n_2^{-1}\|\XX\subaa\supQ(\beta^* - \widehat{\beta}\suba)\|_2^2 \lesssim \frac{k\log p}{n_1}\sigma_{\Tar}^2\right\}, \\
        G_2(t) &= \left\{n_2^{-1}\|\XX\subaa\supQ(\beta^* - \BBhat\widehat{\gamma}\suba)\|_2^2 \lesssim 4\alpha + 2\|(\BBhat - \BB)\gamma^*\|_2^2 + \frac{t^2L}{n_1}\right\},\\
        G_3(t) &= \left\{\max_{\delta:= \gamma_1 - \gamma_2, \gamma_1,\gamma_2\in\Delta_{L_0}}\frac{\bigg|n_2^{-1}\innerproduct{\epsilon\subaa\supQ}{X\subaa\supQ \BBhat_0\delta}\bigg|}{n_2^{-1}\|X\subaa\supQ\BBhat_0 \delta\|_2} \lesssim t\sqrt{L}\right\}
    \end{aligned}
\end{equation*}
where $t>0$, $\alpha$ is defined in \eqref{eq: def of alpha} and $\gamma^*$ is defined in \eqref{eq: best source selection}.
Lemma \ref{lemma: summary} establishes that the above events happen with a high probability.

\begin{Lemma}
Suppose that $\widehat{\beta}\suba$ and $\BBhat = (\widehat{b}^{(1)},...,\widehat{b}^{(L)})$ satisfy Assumption \ref{assumption: b_1}. For $\widehat{\gamma}\suba$ defined in \eqref{eq: est for ini gamma} and $\widehat{\BB}_0$ defined in \eqref{eq: convex}, there exists $c>0$ such that 
    \begin{equation*}
        \begin{aligned}
            \PP(G_1) &\geq 1 - \delta(n_1), \,\,\,\,   \PP\left(G_3(t)\right) \geq 1 - a_0\exp\left(-\frac{c\cdot c_1^2t^2L_0}{16} \right) - a_0\exp\left(-\frac{cn_2}{4}\right) \\
            \PP(G_2(t)) &\geq  1 - a\exp\left(-\frac{c\cdot c_1^2t^2L}{16} \right) - a\exp\left(-\frac{cn_1}{4}\right) - a\exp\left(-\frac{cn_2}{4}\right) - \sum_{l=1}^L \delta_l(N_l),
        \end{aligned}
    \end{equation*}
where $a = 3^L$ and $a_0=3^{L_0}$.
    \label{lemma: summary}
\end{Lemma}
In the following analysis, we condition on the events $G_1$, $G_2(t)$ and $G_3(t)$. By the construction of $\widehat{\mathcal{S}}(\tau)$, any $\gamma\in\widehat{\mathcal{S}}(\tau)$ satisfies
\[
n_2^{-1} \|\YY\subaa\supQ - \XX\subaa\supQ\BBhat_0\gamma\|_2^2 \leq \min\{\widehat{\sigma}^2\subQ, \widehat{\sigma}^2\subsource\} + \tau.
\]
The above inequality first implies 
\begin{equation}
\begin{aligned}
    &n_2^{-1} \|\YY\subaa\supQ - \XX\subaa\supQ\BBhat_0\gamma\|_2^2 \leq \widehat{\sigma}_{\Tar}^2 + \tau = n_2^{-1} \|\YY\subaa\supQ - \XX\subaa\supQ\widehat{\beta}\suba\|_2^2 + \tau \\
    \iff &n_2^{-1} \|\XX\subaa\supQ\beta^* - \XX\subaa\supQ\BBhat_0\gamma + \epsilon\subaa \|_2^2 \leq n_2^{-1} \|\XX\subaa\supQ\beta^* - \XX\subaa\supQ\widehat{\beta}\suba + \epsilon\subaa \|_2^2 + \tau\\
    \iff &\underbrace{n_2^{-1} \|\XX\subaa\supQ(\beta^* - \BBhat_0\gamma)\|_2^2}_{:=V^2} \leq \underbrace{n_2^{-1} \|\XX\subaa\supQ(\beta^* - \widehat{\beta}\suba)\|_2^2}_{:=U_1^2} + 2n_2^{-1}\innerproduct{\epsilon\subaa }{\XX\subaa\supQ(\BBhat_0\gamma - \widehat{\beta}\suba)} + \tau,
\end{aligned}    
\label{eq: thm3 decomp 1}
\end{equation}
where $V = \frac{1}{\sqrt{n_2}} \|\XX\subaa\supQ(\beta^* - \BBhat_0\gamma)\|_2$ and $U_1 = \frac{1}{\sqrt{n_2}} \|\XX\subaa\supQ(\beta^* - \widehat{\beta}\suba)\|_2.$

On the event $G_3(t)$, we take $\delta = \gamma - (1,0,...,0)\trans$ and upper bound $n_2^{-1}\innerproduct{\epsilon\subaa }{\XX\subaa\supQ(\BBhat_0\gamma - \widehat{\beta}\suba)}$ as follows, 
\begin{equation}
    \begin{aligned}
    n_2^{-1}\innerproduct{\epsilon\subaa }{\XX\subaa\supQ(\BBhat_0\gamma - \widehat{\beta}\suba)} &\lesssim
       \frac{t\sqrt{L_0}}{n_2}\|\XX\subaa\supQ(\BBhat_0\gamma - \widehat{\beta}\suba)\|_2 \\
        & \leq t\sqrt{\frac{L_0}{n_2}}\left(\underbrace{n_2^{-1/2}\|X\supQ\subaa(\BBhat_0\gamma - \beta^*)\|_2}_{V} + \underbrace{n_2^{-1/2}\|X\supQ\subaa(\widehat{\beta}\suba - \beta^*)\|_2}_{U_1}\right).
    \end{aligned}
    \label{eq: thm3 decomp 1 II}
\end{equation}
Combining \eqref{eq: thm3 decomp 1}, \eqref{eq: thm3 decomp 1 II}, we then re-write the last inequality of \eqref{eq: thm3 decomp 1} as
\begin{equation}
    \begin{aligned}
        V^2 &= n_2^{-1} \|\XX\subaa\supQ(\beta^* - \BBhat_0\gamma)\|_2^2 \lesssim U_1^2 + 2t\sqrt{\frac{L_0}{n_2}} (V + U_1)+\tau.
    \end{aligned}
    \label{eq: thm3 V2 1}
\end{equation}
Since $2t\sqrt{L_0/n_2}V \leq 100t^2L_0/n_2 + 1/100V^2$, \eqref{eq: thm3 V2 1} implies that
\begin{align*}
    &\frac{99}{100}V^2 \lesssim \tau + U_1^2 + 2t\sqrt{\frac{L_0}{n_2}}U_1 + \frac{100 L_0t^2}{n_2} \\
    &\implies V^2  \lesssim \tau + \left(U_1 + 10t\sqrt{\frac{L_0}{n_2}}\right)^2 \\
    &\implies V^2 \lesssim \tau + U_1^2 + \frac{t^2L_0}{n_2}.
\end{align*}
According to event $G_1$ in Lemma \ref{lemma: summary}, $U_1$ can be bounded by $\frac{k\log p}{n_1}\sigma\subQ^2$ with probability at least $1-\delta(n_1)$. Therefore,
\begin{equation}
    \PP\left(V^2 \lesssim \tau + \frac{k\log p}{n_1}\sigma\subQ^2 + \frac{t^2L_0}{n_2} \right) \geq 1 - \delta(n_1) - a_0\exp\left(-\frac{c\cdot c_1^2t^2L_0}{16} \right) - a_0\exp\left(-\frac{cn_2}{4}\right)
    \label{eq: thm3 decomp 1 final}
\end{equation}

Similar to \eqref{eq: thm3 decomp 1}, by denoting $U_2 = \frac{1}{\sqrt{n_2}}\|\XX\subaa\supQ(\beta^* - \BBhat\widehat{\gamma}\suba)\|_2$, our aim has another bound by merit of the existence of `min' in the construction of $\widehat{\mathcal{S}}(\tau)$:
\begin{align*}
    &n_2^{-1} \|\YY\subaa\supQ - \XX\subaa\supQ\BBhat_0\gamma\|_2^2 \leq \widehat{\sigma}^2\subsource + \tau = n_2^{-1} \|\YY\subaa\supQ - \XX\subaa\supQ\BBhat\widehat{\gamma}\suba\|_2^2 + \tau \\
    \iff &\underbrace{n_2^{-1} \|\XX\subaa\supQ(\beta^* - \BBhat_0\gamma)\|_2^2}_{V^2} \leq \underbrace{n_2^{-1} \|\XX\subaa\supQ(\beta^* - \BBhat\widehat{\gamma}\suba)\|_2^2}_{:=U_2^2} + 2n_2^{-1}\innerproduct{\epsilon\subaa }{\XX\subaa\supQ(\BBhat_0\gamma - \BBhat\widehat{\gamma}\suba)} + \tau
\end{align*}
Control of $n_2^{-1}\innerproduct{\epsilon\subaa }{\XX\subaa\supQ(\BBhat_0\gamma - \BBhat\widehat{\gamma}\suba)}$ is similar to that in \eqref{eq: thm3 decomp 1 II} by replacing $\widehat{\beta}\suba$ with $\BBhat\widehat{\gamma}\suba$, and $\delta = \gamma - (1,0,...,0)\trans$ with $\delta = \gamma - (0,(\widehat{\gamma}\suba)\trans)\trans$. As a result, $V^2$ has another bound as
\[
V^2 \lesssim \tau + U_2^2 + \frac{t^2L_0}{n_2}.
\]
According to event $G_2(t)$ in Lemma \ref{lemma: summary} which specifies the upper bound for $U_2$, we can derive that
\begin{equation}
\begin{aligned}
    &\PP\left(V^2 \lesssim \tau + \alpha + \|(\BBhat - \BB)\gamma^*\|_2^2 + \frac{t^2L}{n_1} + \frac{t^2L_0}{n_2}\right) \\
    &\geq 1 - a\exp\left(-\frac{c\cdot c_1^2t^2L}{16} \right) - a\exp\left(-\frac{cn_1}{4}\right) - a\exp\left(-\frac{cn_2}{4}\right).
\end{aligned}
    \label{eq: thm3 decomp 2}
\end{equation}

Combining \eqref{eq: thm3 decomp 1 final} and \eqref{eq: thm3 decomp 2}, when $n_1 \asymp n_2 \asymp n$ and $L_0 = L+1\asymp L$, we show that with probability at least $1 - a\exp\left(-\frac{c\cdot c_1^2t^2L}{16} \right) - 2a\exp\left(-\frac{cn}{4}\right) - \delta(n)$, 
 \begin{equation}
     n_2^{-1} \|\XX\subaa\supQ(\beta^* - \BBhat_0\gamma)\|_2^2 \lesssim \tau + \frac{t^2L}{n} + \min\bigg\{ \frac{k\log p}{n}\sigma\subQ^2, \,\,\alpha + \|(\BBhat - \BB)\gamma^*\|_2^2\bigg\}.
     \label{eq: lemma summary end}
 \end{equation} 
 By definition of $\widehat{\gamma}\subGART\in\widehat{\mathcal{S}}(\tau)$, $\widehat{\beta}\subGART = \BBhat_0 \widehat{\gamma}\subGART$ shares the same convergence rate in terms of estimation error as in \eqref{eq: lemma summary end}. 

 In terms of the estimation error of the parameter, we leverage the control for event $G_6$ which has been proved in Lemma \ref{lemma for lemma}. 
\begin{Lemma}
For $\BBhat_{\mathsf{double}}:=(\BBhat_0, \BB_0)\in\R^{p\times2L_0}$ and $\Delta_{\mathsf{double}} := \{\gamma_{1}-\gamma_{2}$: $\gamma_{1}, \gamma_{2}\in\Delta_{2L_0}\}$, we have 
\[
\PP\left(\max_{\delta_{\mathsf{double}}\in\Delta_{\mathsf{double}}}\frac{\|\BBhat_{\mathsf{double}} \delta_{\mathsf{double}}\|_2^2}{\frac{1}{n\subaa}\|\XX\supQ\subaa\BBhat_{\mathsf{double}} \delta_{\mathsf{double}}\|_2^2} \lesssim \frac{2}{c_1}\right) \geq 1 - 3^{2L_0}\exp\left(-\frac{cn}{4}\right)
\]
\end{Lemma}
We apply the above lemma with $\widehat{\delta}_{\mathsf{double}} = (\widehat{\gamma}\subGART\supT, \mathbf{0}_{L_0}\supT)\supT - (\mathbf{0}_{L_0}\supT, 1, \mathbf{0}_L\supT)\supT$ and obtain 
\begin{equation*}
    \begin{aligned}
        \|\widehat{\beta}\subGART - \beta^*\|_2^2 &= \|\BBhat_0\widehat{\gamma}\subGART - \BB_0\cdot (1,\mathbf{0}_L\supT)\supT\|_2^2 = \|\BBhat_{\mathsf{double}}\widehat{\delta}_{\mathsf{double}}\|_2^2  \lesssim n_2^{-1}\|\XX\supQ\subaa\BBhat_{\mathsf{double}} \widehat{\delta}_{\mathsf{double}}\|_2^2.
    \end{aligned}
\end{equation*}
The right hand side of the above inequation is equal to $n_2^{-1}\|\XX\supQ\subaa(\widehat{\beta}\subGART - \beta^*)\|_2^2$. Together with \eqref{eq: lemma summary end}, we establish the bound for $\widehat{\beta}\subGART$.

\subsection{Proof of Theorem \ref{thm: second guidance}}
\label{sec: proof for thm initial}
Define $\widehat{\beta}_G^0 := \widehat{\beta}\subGART(\widehat{\beta}^0\subinit, \tau)$ and $\widehat{\beta}_G^1 := \widehat{\beta}\subGART(\widehat{\beta}^1\subinit, \tau)$. Define $\widehat{\mathcal{B}}(\tau) := \{\widehat{B}_0 \gamma: \gamma\in\widehat{\mathcal{S}}(\tau)\}$. With slight abuse of notation, we rewrite $\XX\subaa\supQ$ as $\XX$ and $n_2$ as $n$ in the current proof. We first rewrite the difference between the estimation error of $\widehat{\beta}_G^0$ and $\widehat{\beta}_G^1$ into:
\begin{equation*}
    \begin{aligned}
    \delta &:= \frac{1}{n}\|\XX(\widehat{\beta}_G^0-\beta^*)\|_2^2 - \frac{1}{n}\|\XX(\widehat{\beta}_G^1 -\beta^*)\|_2^2 \\
    &= \frac{1}{n}\|\XX(\widehat{\beta}_G^0-\widehat{\beta}\subinit^1)\|_2^2 - \frac{1}{n}\|\XX(\widehat{\beta}_G^1-\widehat{\beta}\subinit^1)\|_2^2 + \frac{2}{n}\innerproduct{\XX(\widehat{\beta}_G^0-\widehat{\beta}_G^1)}{\XX(\widehat{\beta}\subinit^1-\beta^*)}.
    \end{aligned}
\end{equation*}
Recall that $\widehat{\beta}_G^1 = \argmin_{b \in \widehat{\mathcal{B}}(\tau)}\frac{1}{n}\|\XX(b - \widehat{\beta}\subinit^1)\|_2^2$ and $\widehat{\beta}_G^0 \in \widehat{\mathcal{B}}(\tau)$. Therefore, we derive that $\frac{1}{n}\|\XX(\widehat{\beta}_G^1-\widehat{\beta}\subinit^1)\|_2^2 \leq \frac{1}{n}\|\XX(\widehat{\beta}_G^0-\widehat{\beta}\subinit^1)\|_2^2$ and 
\begin{equation}
\delta \geq \frac{2}{n}\innerproduct{\XX(\widehat{\beta}_G^0-\widehat{\beta}_G^1)}{\XX(\widehat{\beta}\subinit^1-\beta^*)} \geq -\frac{1}{\sqrt{n}}\|\XX(\widehat{\beta}_G^0-\widehat{\beta}_G^1)\|_2 \cdot \frac{1}{\sqrt{n}}\|\XX(\widehat{\beta}\subinit^1-\beta^*)\|_2.
\label{eq: thm4 cond1}
\end{equation}
Similarly, $\delta$ can also be re-written as:
\[\delta = \frac{1}{n}\|\XX(\widehat{\beta}_G^0-\widehat{\beta}\subinit^0)\|_2^2 - \frac{1}{n}\|\XX(\widehat{\beta}_G^1-\widehat{\beta}\subinit^0)\|_2^2 + \frac{2}{n}\innerproduct{\XX(\widehat{\beta}_G^0-\widehat{\beta}_G^1)}{\XX(\widehat{\beta}\subinit^0-\beta^*)}. 
\]
Since $\widehat{\beta}_G^0 = \argmin_{b = \BBhat_0\gamma, \gamma\in\widehat{\mathcal{S}}(\tau)} \frac{1}{n}\|\XX(b - \widehat{\beta}\subinit^0)\|_2^2$ and $\widehat{\beta}_G^0 = \BBhat_0\widehat{\gamma}_G^0$ where $\widehat{\gamma}_G^0 \in\widehat{\mathcal{S}}(\tau)$, we derive that $\frac{1}{n}\|\XX(\widehat{\beta}_G^0-\widehat{\beta}\subinit^0)\|_2^2 \leq \frac{1}{n}\|\XX(\widehat{\beta}_G^1-\widehat{\beta}\subinit^0)\|_2^2$ and 
\begin{equation}
\delta \leq \frac{2}{n}\innerproduct{\XX(\widehat{\beta}_G^0-\widehat{\beta}_G^1)}{\XX(\widehat{\beta}\subinit^0 -\beta^*)} \leq \frac{1}{\sqrt{n}}\|\XX(\widehat{\beta}_G^0-\widehat{\beta}_G^1)\|_2 \cdot \frac{1}{\sqrt{n}}\|\XX(\widehat{\beta}\subinit^0-\beta^*)\|_2.
\label{eq: thm4 cond2}
\end{equation}

When $\tau\geq\min\{\frac{k\log p}{n}\sigma\subQ^2, \alpha + \left(\sum_{l=1}^L \gamma_l^* \sqrt{\frac{k_l\log p}{N_l}\sigma_l^2}\right)^2\}$, Theorem \ref{thm: first guidance} implies the following inequality,
\begin{equation}
    \frac{1}{n}\|\XX(\widehat{\beta}_G^0-\widehat{\beta}_G^1)\|_2^2 \leq (\frac{1}{\sqrt{n}}\|\XX(\widehat{\beta}_G^0-\beta^*)\|_2 + \frac{1}{\sqrt{n}}\|\XX(\widehat{\beta}_G^1 - \beta^*)\|_2)^2 \lesssim \tau.
    \label{eq: thm4 cond3}
\end{equation}
Therefore, we first derive that $\delta$ lies on the following interval with a high probability:
\begin{equation}
-\frac{\sqrt{\tau}}{\sqrt{n}}\|\XX(\widehat{\beta}\subinit^1-\beta^*)\|_2 \lesssim \delta \lesssim \frac{\sqrt{\tau}}{\sqrt{n}}\|\XX(\widehat{\beta}\subinit^0-\beta^*)\|_2.
\label{eq: thm4 delta range}
\end{equation}

To show the scenario when $\delta\geq 0$, we further denote $\widehat{\beta}^*_G := \widehat{\beta}\subGART(\beta^*, \tau)$. Based on the triangle inequality, $\frac{1}{\sqrt{n}}\|\XX(\widehat{\beta}_G^0-\beta^*)\|_2 \geq \frac{1}{\sqrt{n}}\|\XX(\widehat{\beta}_G^0-\widehat{\beta}_G^1)\|_2 - \frac{1}{\sqrt{n}}\|\XX(\widehat{\beta}_G^1-\beta^*)\|_2$. Therefore, 
\begin{equation}
    \frac{1}{\sqrt{n}}\|\XX(\widehat{\beta}_G^0-\beta^*)\|_2 - \frac{1}{\sqrt{n}}\|\XX(\widehat{\beta}_G^1-\beta^*)\|_2 \geq \frac{1}{\sqrt{n}}\|\XX(\widehat{\beta}_G^0-\widehat{\beta}_G^1)\|_2  - \frac{2}{\sqrt{n}}\|\XX(\widehat{\beta}_G^1-\beta^*)\|_2.
    \label{eq: cor 1 cond4}
\end{equation}
Additionally, since $
\frac{1}{\sqrt{n}}\|\XX(\widehat{\beta}_G^1-\beta^*)\|_2 \leq \frac{1}{\sqrt{n}}\|\XX(\widehat{\beta}_G^1-\widehat{\beta}_G^*)\|_2 + \frac{1}{\sqrt{n}}\|\XX(\widehat{\beta}_G^*-\beta^*)\|_2$, \eqref{eq: cor 1 cond4} can be further bounded by
\begin{equation}
\begin{aligned}
    &\frac{1}{\sqrt{n}}\|\XX(\widehat{\beta}_G^0-\beta^*)\|_2 - \frac{1}{\sqrt{n}}\|\XX(\widehat{\beta}_G^1-\beta^*)\|_2 \\
    &\geq \frac{1}{\sqrt{n}}\|\XX(\widehat{\beta}_G^0-\widehat{\beta}_G^1)\|_2  - 2[\underbrace{\frac{1}{\sqrt{n}}\|\XX(\widehat{\beta}_G^1-\widehat{\beta}_G^*)\|_2}_{\text{(I)}} + \underbrace{\frac{1}{\sqrt{n}}\|\XX(\widehat{\beta}_G^*-\beta^*)\|_2}_{\text{(II)}}] 
\end{aligned}
    \label{thm4: cond3}
\end{equation}
Since $\widehat{\beta}_G^* = \argmin_{b = \widehat{\mathcal{B}}(\tau)}\frac{1}{n}\|\XX(b - \beta^*)\|_2^2$ and $\widehat{\beta}\suba\in\widehat{\mathcal{B}}(\tau)$, the estimation error of $\widehat{\beta}_G^*$ satisfies
\begin{equation}
    \text{(II)} = \frac{1}{\sqrt{n}}\|\XX(\widehat{\beta}_G^*-\beta^*)\|_2 \leq \frac{1}{\sqrt{n}}\|\XX(\widehat{\beta}\suba-\beta^*)\|_2. 
    \label{thm4: cond5}
\end{equation}
To find an upper bound for (I), recall the definition of $\widehat{\beta}_G^1$ as the projection of $\widehat{\beta}\subinit^1$ onto the convex set $\widehat{\mathcal{B}}(\tau)$, and also $\widehat{\mathcal{B}}(\tau)$ is a convex set. Thus, for all $\beta\in\widehat{\mathcal{B}}(\tau)$ and all $\lambda\in(0,1)$, we have
\begin{align*}
    \|\XX(\widehat{\beta}\subinit^1 - \widehat{\beta}_G^1)\|_2^2 &\leq \|\XX\left(\widehat{\beta}\subinit^1- \lambda\beta-(1-\lambda)\widehat{\beta}_G^1\right)\|_2^2 \\
    &= \|\XX(\widehat{\beta}\subinit^1 - \widehat{\beta}_G^1)\|_2^2 - 2\lambda\innerproduct{\XX(\beta - \widehat{\beta}_G^1)}{\XX(\widehat{\beta}\subinit^1 - \widehat{\beta}_G^1)} + \lambda^2\|\XX(\beta - \widehat{\beta}_G^1)\|_2^2 \\
    \iff \lambda^2\|\XX(\beta &- \widehat{\beta}_G^1)\|_2^2 - 2\lambda\innerproduct{\XX(\beta - \widehat{\beta}_G^1)}{\XX(\widehat{\beta}\subinit^1 - \widehat{\beta}_G^1)} \geq 0
\end{align*}

By letting $\lambda\to 0$, we conclude that \begin{equation}
    \innerproduct{\XX(\beta - \widehat{\beta}_G^1)}{\XX(\widehat{\beta}\subinit^1 - \widehat{\beta}_G^1)}\leq 0 \text{ for any }\beta\in\widehat{\mathcal{B}}(\tau).
    \label{eq: thm4 cond7}
\end{equation}
Similarly, since $\widehat{\beta}^*_G$ is the projection of $\beta^*$ onto $\widehat{\mathcal{B}}(\tau)$, we obtain that 
\begin{equation}
    \innerproduct{\XX(\beta - \widehat{\beta}_G^*)}{\XX(\beta^* - \widehat{\beta}_G^*)}\leq 0 \text{ for any }\beta\in\widehat{\mathcal{B}}(\tau).
    \label{eq: thm4 cond8}
\end{equation}

By taking $\beta = \widehat{\beta}_G^*$ in \eqref{eq: thm4 cond7} and $\beta = \widehat{\beta}^1_G$ in \eqref{eq: thm4 cond8}, we then sum up the two inequalities to derive 
\begin{equation}
    \begin{aligned}
        &\innerproduct{\XX(\widehat{\beta}_G^* - \widehat{\beta}_G^1)}{\XX(\widehat{\beta}\subinit^1 - \widehat{\beta}_G^1)} + \innerproduct{\XX(\widehat{\beta}_G^1 - \widehat{\beta}_G^*)}{\XX(\beta^* - \widehat{\beta}_G^*)} \leq 0 \\
        &\implies \innerproduct{\XX(\widehat{\beta}_G^* - \widehat{\beta}_G^1)}{\XX(\beta^* - \widehat{\beta}\subinit^1)} \geq \|\XX(\widehat{\beta}_G^* - \widehat{\beta}_G^1)\|_2^2 \\
        &\implies \text{(I)} = \frac{1}{\sqrt{n}}\|\XX(\widehat{\beta}_G^* - \widehat{\beta}_G^1)\|_2^2 \leq \frac{1}{\sqrt{n}}\|\XX(\beta^* - \widehat{\beta}\subinit^1)\|_2^2 
    \end{aligned}
        \label{thm4: cond4}
\end{equation}

Combining \eqref{thm4: cond5} and \eqref{thm4: cond4}, we derive an upper bound for \eqref{thm4: cond3}:
\begin{equation*}
    \begin{aligned}
        &\frac{1}{\sqrt{n}}\|\XX(\widehat{\beta}_G^0-\beta^*)\|_2 - \frac{1}{\sqrt{n}}\|\XX(\widehat{\beta}_G^1-\beta^*)\|_2 \\
        &\geq \frac{1}{\sqrt{n}}\|\XX(\widehat{\beta}_G^0-\widehat{\beta}_G^1)\|_2 - 2\left(\frac{1}{\sqrt{n}}\|\XX(\widehat{\beta}\suba-\beta^*)\|_2  + \frac{1}{\sqrt{n}}\|\XX(\widehat{\beta}\subinit^1-\beta^*)\|_2\right)
    \end{aligned}
\end{equation*}

When $\frac{1}{\sqrt{n}}\|\XX(\widehat{\beta}_G^0-\widehat{\beta}_G^1)\|_2 > 2\left(\frac{1}{\sqrt{n}}\|\XX(\widehat{\beta}\suba-\beta^*)\|_2  + \frac{1}{\sqrt{n}}\|\XX(\widehat{\beta}\subinit^1-\beta^*)\|_2\right)$, we have 
\[
\delta = \left(\frac{1}{\sqrt{n}}\|\XX(\widehat{\beta}_G^0-\beta^*)\|_2 - \frac{1}{\sqrt{n}}\|\XX(\widehat{\beta}_G^1-\beta^*)\|_2\right) \cdot \left(\frac{1}{\sqrt{n}}\|\XX(\widehat{\beta}_G^0-\beta^*)\|_2 + \frac{1}{\sqrt{n}}\|\XX(\widehat{\beta}_G^1-\beta^*)\|_2\right) > 0.
\]

\subsection{Proof of Lemmas, Corollaries and Propositions}
\label{sec: proof for lemmas}
\subsubsection{Proof of Lemma \ref{lemma: consistency them 1}}

\begin{proof}
    We first write out $|\widehat{f}_n(\gamma) - f(\gamma)|$ according to their definitions and decompose it into three different types of error:
        \begin{align*}
             \bigg|\widehat{f}_n(\gamma) - f(\gamma)\bigg| &=  \bigg|n_2^{-1}\|\YY\subaa\supQ - \XX\subaa\supQ\BBhat_0\gamma\|_2^2 - \min\{\widehat{\sigma}_{\Tar}^2, \widehat{\sigma}\subsource^2\} - \E\left[(Y_i\supQ - (X_i\supQ)\trans\BB_0\gamma)^2\right] - \sigma_{\Tar}^2 \bigg|\\
             &\leq \underbrace{\bigg|\E\left[(Y_i\supQ - (X_i\supQ)\trans\BBhat_0\gamma)^2  \right] - \E\left[(Y_i\supQ - (X_i\supQ)\trans\BB_0\gamma)^2  \right]\bigg|}_{\text{(I)}}  +  \\
             & \underbrace{\bigg|n_2^{-1}\|\YY\subaa\supQ - \XX\subaa\supQ\BBhat_0\gamma\|_2^2 - \E\left[(Y_i\supQ - (X_i\supQ)\trans\BBhat_0\gamma)^2 \right]\bigg|}_{\text{(II)}} + \underbrace{\bigg| \min\{\widehat{\sigma}_{\Tar}^2, \widehat{\sigma}\subsource^2\} - \sigma_{\Tar}^2\bigg|}_{\text{(III)}}
             \stepcounter{equation}\tag{\theequation}\label{eq: thm consistency 1}
        \end{align*}
    (I) measures the relative expected prediction errors with coefficient $\BB_0\gamma$ versus $\BBhat_0\gamma$, which can be further decomposed into:
    \begin{align*}
        & \E\left[(Y_i\supQ - (X_i\supQ)\trans\BBhat_0\gamma)^2 -  \left(Y_i\supQ - (X_i\supQ)\trans\BB_0\gamma\right)^2\right] \\
        &= \E\left[\left((X_i\supQ)\trans(\BB_0\gamma - \BBhat_0\gamma)\right)\cdot
        \left(2Y_i\supQ - (X_i\supQ)\trans(\BBhat_0\gamma + \BB_0\gamma)\right)\right] \\
        &= \E\left[\left((X_i\supQ)\trans(\BB_0\gamma - \BBhat_0\gamma)\right)\cdot
        \left(2(X_i\supQ)\trans\beta^* + 2\epsilon\supQ_i - (X_i\supQ)\trans(\BBhat_0\gamma - \BB_0\gamma) - 2(X_i\supQ)\trans\BB_0\gamma\right)\right] \\
        &= \underbrace{\E\left[(\BB_0\gamma - \BBhat_0\gamma)\trans X_i\supQ(X_i\supQ)\trans(\BB_0\gamma - \BBhat_0\gamma) \right]}_{\text{(I,1)}} \\
         & \,\,\,\,\,\,\,\, +  \underbrace{\E\left[2(\BB_0\gamma - \BBhat_0\gamma)\trans X_i\supQ(X_i\supQ)\supT(\beta^* - \BB_0\gamma)\right]}_{\text{(I,2)}} + \underbrace{\E\left[2\epsilon\supQ_i\cdot (X_i\supQ)\supT(\BB_0\gamma - \BBhat_0\gamma)\right]}_{\text{(I,3)}}.
    \end{align*}
    The second equation is because $Y_i\supQ$ is sampled from the target population in \eqref{eq: target model}.
    Since $\BBhat_0$ is an unbiased estimator of $\BB_0$, and $\BBhat_0$ does not contain data from $X\subaa\supQ$ that $X_i\supQ$ comes from, we have:
    \begin{align*}
        \text{(I, 2)} &=     \E_{X_i\supQ}\E_{\BBhat_0|X_i\supQ}\left[(\BB_0\gamma - \BBhat_0\gamma)\trans X_i\supQ(X_i\supQ)\trans\left(\beta^* - \BB_0\gamma\right)\right] = 0 \\
         \text{(I, 3)} &= \E_{X_i\supQ}\E_{\epsilon_i|X_i}\E_{\BBhat_0}\left[(\BB_0\gamma - \BBhat_0\gamma)\trans X_i\supQ\epsilon_i\right] = 0
    \end{align*}
    In addition, according to Assumption \ref{assumption: b_1}, for a fixed $L$, 
    \begin{equation*}
    \begin{aligned}
        \text{(I, 1)} &= \E\left[\left(\gamma_0(X_i\supQ)\trans(\beta^* - \widehat{\beta}\suba) + \sum_{l=1}^L \gamma_l(X_i\supQ)\trans(b^{(l)} - \widehat{b}^{(l)})\right)^2 \right] \\
        &\leq \max\bigg\{\E\left[\left((X_i\supQ)\trans(\beta^* - \widehat{\beta}\suba)\right)^2\right], \max_{l\in\{1,...,L\}}\E\left[\left( (X_i\supQ)\trans(b^{(l)} - \widehat{b}^{(l)})\right)^2 \right] \bigg\} \to 0.
    \end{aligned}
    \end{equation*}
  Therefore, $\text{(I)} \to 0$ regardless of the choice of $\gamma$. 

  For (II), given that $(\YY\subaa\supQ, \XX\subaa\supQ)$ is independent of $\BBhat_0$, we can first condition on $\BBhat_0$ when considering the estimation error of $n_2^{-1}\|\YY\subaa\supQ - \XX\subaa\supQ\BBhat_0\gamma\|$. Treating $\BBhat_0$ as a constant, we define
  \begin{equation}
      \begin{aligned}
            h(\gamma, Y_i\supQ, X_i\supQ) &= \left(Y_i\supQ - (X_i\supQ)\supT\BBhat_0\gamma\right)^2 \\
            &\leq \max\left\{\left(Y_i\supQ - (X_i\supQ)\supT\widehat{\beta}\suba\right)^2, \max_{l\in\{1,...,L\}}\left(Y_i\supQ - (X_i\supQ)\supT\widehat{b}^{(l)}\right)^2 \right\} := d(Y_i\supQ, X_i\supQ),
      \end{aligned}
      \label{eq: lemma1 cond1}
  \end{equation}
  for any $\gamma\in\Delta_{L_0}$. Since the variance of $Y_i\supQ$ and $Y_i\supl$ is bounded from $\infty$ for any $l\in\{1,...,L\}$, we have
  \begin{equation}
      \begin{aligned}
          \max\left\{\E\left(Y_i\supQ - (X_i\supQ)\supT\widehat{\beta}\suba\right)^2, \max_{l\in\{1,...,L\}}\E\left(Y_i\supQ - (X_i\supQ)\supT\widehat{b}^{(l)}\right)^2 \right\} < \infty.
      \end{aligned}
      \label{eq: lemma1 cond2}
  \end{equation}
  As a result, combining \eqref{eq: lemma1 cond1} and \eqref{eq: lemma1 cond2}, given a constant $L$, we can derive that 
  \begin{equation}
      h(\gamma, Y_i\supQ, X_i\supQ) \leq d(Y_i\supQ, X_i\supQ) \;\;\;\text{with} \;\;\; \E\left[d(Y_i\supQ, X_i\supQ)\right]<\infty
      \label{eq: lemma1 cond3}
   \end{equation}
  Additionally, $\Delta_{L_0}$ that $\gamma$ belongs to is compact, and $h(\gamma, Y_i\supQ, X_i\supQ)$ is continuous at each $\gamma\in\Delta_{L_0}$. Then by applying the uniform law of large numbers, for any $\epsilon > 0$,
  \[
  \lim_{n_2\to\infty} \PP\left(\sup_{\gamma\in\Delta_{L_0}} \bigg|n_2^{-1}\sum_{i=1}^{n_2}\left(Y_i\supQ - (X_i\supQ)\supT\BBhat_0\gamma\right)^2 - \E\left(Y_i\supQ - X_i\supQ\BBhat_0\gamma\right)^2\bigg| \leq \epsilon \bigg| \BBhat_0 \right)= 0
  \]
  Therefore,
  \begin{equation}
      \begin{aligned}
          \lim_{n_2 \to \infty} \PP\left(\text{(II)} \leq \epsilon\right) &= \lim_{n_2 \to \infty} \E_{\BBhat_0}\PP\left(\text{(II)} \leq \epsilon|\BBhat_0\right)= \E_{\BBhat_0} \lim_{n_2 \to \infty} \PP\left(\text{(II)} \leq \epsilon|\BBhat_0\right) =0 \\
      \end{aligned}
  \end{equation}
  We can interchange the expectation and limit in the second equation since $\PP(\text{(II)}\leq\epsilon|\BBhat_0)$ is bounded by 1. In other words, when $n_2\to\infty$:
    \begin{equation}
        \sup_{\gamma\in\Delta_{L_0}} \bigg|n_2^{-1}\|\YY\subaa\supQ - \XX\subaa\supQ\BBhat_0\gamma\|_2^2 - \E\left[(Y_i - (X_i\supQ)\trans\BBhat_0\gamma)^2\right]\bigg| \overset{p}{\to} 0
        \label{eq: ulln 1}
    \end{equation}
  
   For (III), based on the design of $\widehat{\sigma}_{\Tar}^2$ in \eqref{eq: est for sigmaQ, 1} and $\widehat{\sigma}\subsource^2$ in \eqref{eq: est for sigmaQ, 2}, we have 
    $$\widehat{\sigma}_{\Tar}^2 \overset{p}{\to} \sigma_{\Tar}^2 \text{ and } \widehat{\sigma}\subsource^2 \overset{p}{\to} \sigma_{\Tar}^2 + \argmin_{\gamma\in\Delta_{L}} (\beta^* - \BB\gamma)\trans\Sigma_{\Tar} (\beta^* - \BB\gamma) \geq \sigma_{\Tar}^2.$$
    Therefore, $\min\{\widehat{\sigma}_{\Tar}^2, \widehat{\sigma}\subsource^2\} - \sigma_{\Tar}^2 \overset{p}{\to} 0$ when $n_1, n_2\to \infty$. 

    Combining the convergence of (I), (II) and (III), we derive that
    \[
    \sup_{\gamma\in\Delta_{L_0}} \bigg|\widehat{f}_n(\gamma) - f(\gamma)\bigg| = \sup_{\gamma\in\Delta_{L_0}} \text{(I) + (II) + (III)}  \overset{p}{\to} 0.
    \]
    
\end{proof}

\subsubsection{Proof of Lemma \ref{lemma: thm2 assist 2}}
\begin{proof}
We first show that $g(\widetilde{\gamma}) - g(\widebar{\gamma}) \overset{p}{\to} 0$. When $f(\widetilde{\gamma})\leq 0$ and $\widehat{f}_n(\widetilde{\gamma}) \leq 0$, it is obvious that $\widetilde{\gamma} = \widebar{\gamma}$ and therefore $g(\widetilde{\gamma}) = g(\widebar{\gamma})$. When $f(\widetilde{\gamma})>0$ and $\widehat{f}_n(\widetilde{\gamma})\leq 0$, note that there exists $\gamma^0 = (1,0,...,0)\in\Delta_{L_0}$ such that $f(\gamma^0) = -\tau <0$. Accordingly, we construct 
\begin{equation}
    \widetilde{\gamma}^0 = \lambda\widetilde{\gamma} + (1-\lambda)\gamma^0
    \label{eq: lemma thm2 assist 2 cond0}
\end{equation}
where $\lambda := \frac{f(\gamma^0)}{f(\gamma^0) - f(\widetilde{\gamma})} = \frac{-\tau}{-\tau - f(\widetilde{\gamma})} \in (0,1)$. Since $f(\gamma)$ is a convex function on $\Delta_{L_0}$ and $\widetilde{\gamma}, \gamma^0 \in \Delta_{L_0}$,
\[
f(\widetilde{\gamma}^0) \leq \lambda f(\widetilde{\gamma}) + (1-\lambda)f(\gamma^0) = 0.
\]
Additionally, given that $\widehat{f}_n(\gamma)$ converges uniformly to $f(\gamma)$ in probability for all $\gamma\in\Delta_{L+1}$, we have $\widehat{f}_n(\widetilde{\gamma}^0) - f(\widetilde{\gamma}^0)  \overset{p}{\to} 0$, which means that $\widehat{f}_n(\widetilde{\gamma}^0)\leq 0$ with high probability when $n$ is large.
In other words, $\widetilde{\gamma}^0 \in \{\gamma\in\Delta_{L_0}:  \widehat{f}_n(\gamma)\leq 0, f(\gamma)\leq 0\}$ when $n$ is large. Recall the definition of $\widebar{\gamma}$ and $\widetilde{\gamma}$, we then have
\begin{equation}
    g(\widetilde{\gamma}^0)\geq g(\widebar{\gamma}) \geq g(\widetilde{\gamma}).
    \label{eq: lemma thm2 assist 2 cond1}
\end{equation}
Given $\widetilde{\gamma}^0$ is a linear interpolation of $\widetilde{\gamma}$ and $\gamma^0$, we then bound the distance between $\widetilde{\gamma}^0$ and $\widetilde{\gamma}$:
\begin{equation}
    \|\widetilde{\gamma}^0 - \widetilde{\gamma}\|_2 =  \frac{f(\widetilde{\gamma})}{f(\widetilde{\gamma}) - \delta}\|\widetilde{\gamma} - \widebar{\gamma}\|_2 \leq \frac{f(\widetilde{\gamma})}{f(\widetilde{\gamma}) - \delta}(\|\widetilde{\gamma}\|_2 + \|\widebar{\gamma}\|_2) \leq \frac{2f(\widetilde{\gamma})}{f(\widetilde{\gamma}) - \delta}
    \label{eq: lemma thm2 assist 2 cond2}
\end{equation}
The last inequality results from the Cauchy–Schwarz inequality as $\widetilde{\gamma}, \widebar{\gamma}\in\Delta_{L_0}$. We then further control $f(\widetilde{\gamma})$. Due to the uniform convergence of $\widehat{f}_n(\gamma)$ to $f(\gamma)$, $\widehat{f}_n(\widetilde{\gamma}) -
f(\widetilde{\gamma}) \overset{p}{\to} 0$, which means for any $\xi>0, \xi_0>0$, there exists $N>0$ such that for any $n>N$, $\PP\left( \big|\widehat{f}_n(\widetilde{\gamma}) - f(\widetilde{\gamma})\big| \leq \xi_0\right) \geq 1 - \xi$. Since $\widehat{f}_n(\widetilde{\gamma}) \leq 0$ and $f(\widetilde{\gamma}) > 0$, $\{\big|\widehat{f}_n(\widetilde{\gamma}) - f(\widetilde{\gamma})\big| \leq \xi_0\} \subseteq \{f(\widetilde{\gamma})\leq \xi_0\}$ and
\begin{equation}
 \PP\left( f(\widetilde{\gamma})\leq \xi_0\right) \geq 1 - \xi  \implies f(\widetilde{\gamma})\overset{p}{\to} 0.
    \label{eq: lemma thm2 assist 2 cond3}
\end{equation}
Combining \eqref{eq: lemma thm2 assist 2 cond2} and \eqref{eq: lemma thm2 assist 2 cond3}, $\|\widetilde{\gamma}^0 - \widetilde{\gamma}\|_2\overset{p}{\to} 0$. Due to the continuity of $g(\cdot)$, we then have $g(\widetilde{\gamma}^0) - g(\widetilde{\gamma}) \overset{p}{\to} 0$. Recall in \eqref{eq: lemma thm2 assist 2 cond1}, $g(\widebar{\gamma})$ is bounded by $g(\widetilde{\gamma})$ and $g(\widetilde{\gamma}^0)$. As a result, $g(\widebar{\gamma}) - g(\widetilde{\gamma}) \overset{p}{\to} 0$.

The proof for the convergence of $g(\gamma_{\mathrm{G}}) - g(\widebar{\gamma})$ is similar to that of $g(\widetilde{\gamma}) - g(\widebar{\gamma})$, since there exists $\gamma^0_n = \argmin_{\gamma \in \{\gamma^0, \widehat{\gamma}\suba\}} \widehat{f}_{n}(\gamma) $ such that $\widehat{f}_n(\gamma^0_n) = -\tau <0$, where $\widehat{\gamma}\suba$ defined in \eqref{eq: est for ini gamma}. Therefore, we can construct $\widetilde{\gamma}^0 = \lambda\gamma_{\mathrm{G}} + (1-\lambda)\gamma^0_n$ where $\gamma_{\mathrm{G}}$ serves like $\widetilde{\gamma}$ and $\gamma_n^0$ like $\gamma_0$ in \eqref{eq: lemma thm2 assist 2 cond0}.

\end{proof}

\subsubsection{Proof of Lemma \ref{lemma: summary}}
\begin{proof}
 The control of the event $G_1$ directly follows from assumption \ref{assumption: b_1}. 

To facilitate the proof for the event $G_2(t)$ and $G_3(t)$ in Lemma \ref{lemma: summary}, we further introduce the following events:

    \begin{align*}
        G_4^j(w_1, w_2, t) &= \left\{\bigg|w_1\supT\left(\frac{1}{n_j}(\XX\subaj\supQ)\supT \XX\subaj\supQ - \Sigma\subQ\right)w_2 \bigg|\lesssim \frac{t}{\sqrt{n_j}}\|\Sigma\subQ^{1/2} w_1\|_2\|\Sigma\subQ^{1/2} w_2\|_2\right\} \text{  for } j=1,2, \\
        G_5(t) &= \left\{\max_{\delta:= \gamma_1 - \gamma_2, \gamma_1,\gamma_2\in\Delta_{L_0}}\frac{\bigg|n_2^{-1}\innerproduct{\epsilon\subaa\supQ}{\XX\subaa\supQ \BBhat_0\delta}\bigg|}{\frac{1}{\sqrt{n_2}}\|\BBhat_0 \delta\|_2} \lesssim t\sqrt{L}\right\} \\
        G_6 &= \left\{\max_{\delta:= \gamma_1 - \gamma_2, \gamma_1,\gamma_2\in\Delta_{L_0}}\frac{\|\BBhat_0 \delta\|_2}{\frac{1}{n_2}\|\XX\supQ\subaa\BBhat_0 \delta\|_2^2} \lesssim \frac{2}{c_1}\right\} \\
        G_7(t) &= \left\{\max_{\delta:= \gamma_1 - \gamma_2, \gamma_1,\gamma_2\in\Delta_{L}}\frac{\bigg|\frac{1}{n_1}\innerproduct{\epsilon\suba\supQ}{\XX\suba\supQ \BBhat\delta}\bigg|}{\frac{1}{n_1}\|\XX\suba\supQ\BBhat \delta\|_2} \lesssim t\sqrt{L}\right\}, \\
        G_8 &= \left\{\max_{\delta:= \gamma_1 - \gamma_2, \gamma_1,\gamma_2\in\Delta_{L}}\frac{n_2^{-1}\|\XX\subaa\supQ\BBhat \delta\|_2^2}{\frac{1}{n_1}\|\XX\suba\supQ\BBhat \delta\|_2^2}\lesssim \frac{3C_1}{c_1}\right\}, \\
        G_9 &= \left\{    \max\bigg\{\frac{n_1^{-1}\|\XX\suba\supQ(\BB - \BBhat){\gamma}^*\|_2^2}{\|(\BB - \BBhat){\gamma}^*\|_2^2} , \frac{n_2^{-1}\|\XX\subaa\supQ(\BB - \BBhat){\gamma}^*\|_2^2}{\|(\BB - \BBhat){\gamma}^*\|_2^2}\bigg\} \lesssim C_1 \right\},\\
        G_{10}(t) &= \left\{n_2^{-1}\|\XX\subaa\supQ\BBhat({\gamma}^* -\widehat{\gamma}\suba)\|_2^2 \lesssim 2\alpha + \|(\BBhat - \BB)\gamma^*\|_2^2 + \frac{4t^2L}{n_1}\right\},
    \end{align*}

where $w_1, w_2\in \R^{p}$ and $t>0$.

\begin{Lemma}
     Suppose that $\widehat{\beta}\subaa$ and $\BBhat = (\widehat{b}^{(1)},...,\widehat{b}^{(L)})$ satisfy Assumption \ref{assumption: b_1}. For $\widehat{\gamma}\suba$ defined in \eqref{eq: est for ini gamma}, $\widehat{\BB}_0$ defined in \eqref{eq: convex}, there exists $c>0$ such that,
    \begin{equation*}
        \begin{aligned}
           &\PP\left(G_4^j(w_1, w_2, t)\right) \geq 1 - 2\exp(-ct^2),\text{ for }j=1,2, \\
            &\PP(G_5(t)) \geq 1 - a_0\exp\left(-\frac{ct^2L_0}{4}\right),\,\,\, \PP(G_6) \geq 1 - a_0\exp\left(-\frac{cn\subaa}{4}\right), \\
            &\PP(G_7(t)) \geq 1 - a\exp\left(-\frac{c\cdot c_1^2t^2L}{16} \right) - a\exp\left(-\frac{cn_1}{4}\right), \\
            &\PP(G_8) \geq 1 - a\exp\left(-\frac{cn_2}{4}\right) - a\exp\left(-\frac{cn_1}{4}\right), \\
            &\PP(G_9) \geq 1-\exp(-cn_1)-\exp(-cn_2), \\ 
            &\PP(G_{10}(t)) \geq 1 - a\exp\left(-\frac{c\cdot c_1^2t^2L}{16} \right) - a\exp\left(-\frac{cn_1}{4}\right) - a\exp\left(-\frac{cn_2}{4}\right),
        \end{aligned}
    \end{equation*}
\label{lemma for lemma}
\end{Lemma}
where $a=3^{L}$, $a_0=3^{L_0}$, $w_1, w_2\in \R^{p}$ that are independent of $\epsilon\supQ\subaj$ and $X\supQ\subaj$.

For the event $G_3(t)$, we can decompose the left-hand side of inequality in $G_3(t)$ into:
    \begin{align*}
        \underbrace{\max_{\delta:= \gamma_1 - \gamma_2, \gamma_1,\gamma_2\in\Delta_{L_0}}\frac{\bigg|n_2^{-1}\innerproduct{\epsilon\subaa\supQ}{X\subaa\supQ \BBhat_0\delta}\bigg|}{n_2^{-1}\|X\subaa\supQ\BBhat_0 \delta\|_2}}_{\text{(I)}} \leq \underbrace{\max_{\delta:= \gamma_1 - \gamma_2, \gamma_1,\gamma_2\in\Delta_{L_0}}\frac{\bigg|n_2^{-1}\innerproduct{\epsilon\subaa\supQ}{X\subaa\supQ \BBhat_0\delta}\bigg|}{\frac{1}{\sqrt{n_2}}\|\BBhat_0 \delta\|_2}}_{\text{(II)}} \cdot \underbrace{\max_{\delta:= \gamma_1 - \gamma_2, \gamma_1,\gamma_2\in\Delta_{L_0}}\frac{\frac{1}{\sqrt{n_2}}\|\BBhat_0 \delta\|_2}{n_2^{-1}\|X\subaa\supQ\BBhat_0 \delta\|_2}}_{\text{(III)}}
    \end{align*}
    Therefore, we have
    \begin{equation*}
        \begin{aligned}
            \PP\left(\text{(I)}\geq t\sqrt{L}\right) & \leq \PP\left(\text{(II)} \cdot \text{(III)}\geq t\sqrt{L}\right) \leq \PP\left(\text{(II)} \geq \frac{c_1t\sqrt{L}}{2}\right) + \PP\left(\text{(III)} \geq \frac{2}{c_1}\right)
        \end{aligned}
    \end{equation*}
    According to the probability of event $G_5$ and $G_6$ in Lemma \ref{lemma for lemma}, $\PP\left(\text{(II)} \geq \frac{c_1t\sqrt{L}}{2}\right)$ and $\PP\left(\text{(III)} \geq \frac{2}{c_1}\right)$ can be upper bounded by $a_0\exp\left(-\frac{c\cdot c_1^2t^2L_0}{16} \right)$ and $a_0\exp\left(-\frac{cn_2}{4}\right)$. Therefore,  
    $$
    \PP\left(\text{(I)}\geq t\sqrt{L}\right) \leq a\exp\left(-\frac{c\cdot c_1^2t^2L_0}{16} \right) - a\exp\left(-\frac{cn_2}{4}\right).
    $$

    For event $G_2(t)$, we can decompose $\frac{1}{\sqrt{n_2}}\|\XX\subaa\supQ(\beta^* - \BBhat\widehat{\gamma}\suba)\|_2$ into:
    \begin{equation*}
        \begin{aligned}
            \frac{1}{\sqrt{n_2}}\|\XX\subaa\supQ(\beta^* - \BBhat\widehat{\gamma}\suba)\|_2 &\leq \frac{1}{\sqrt{n_2}}\|\XX\subaa\supQ(\BB{\gamma}^* - \BBhat\widehat{\gamma}\suba)\|_2 + \frac{1}{\sqrt{n_2}}\|\XX\subaa\supQ(\BB{\gamma}^* - \beta^*)\|_2 \\
            &\leq \underbrace{\frac{1}{\sqrt{n_2}}\|\XX\subaa\supQ\BBhat(\widehat{\gamma}\suba - {\gamma}^*)\|_2}_{\text{(IV)}} + \underbrace{\frac{1}{\sqrt{n_2}}\|\XX\subaa\supQ(\BB - \BBhat){\gamma}^*\|_2}_{\text{(V)}} + \underbrace{\frac{1}{\sqrt{n_2}}\|\XX\subaa\supQ(\BB{\gamma}^* - \beta^*)\|_2}_{\text{(VI)}}. \\
        \end{aligned}
    \end{equation*}
    According to event $G_4^2(w_1,w_2,\sqrt{n_2})$ where $w_1=w_2=\BB\gamma^*-\beta^*$, $\text{(VI)}^2$ can be upper bounded by
    \begin{equation}
        \begin{aligned}
                \text{(VI)}^2 &= n_2^{-1}\|\XX\subaa\supQ(\BB{\gamma}^* - \beta^*)\|_2^2 - \|\Sigma^{\frac{1}{2}}\subQ(\beta^* - \BB{\gamma}^*)\|_2^2 + \|\Sigma^{\frac{1}{2}}\subQ(\beta^* - \BB{\gamma}^*)\|_2^2 \\
                &\lesssim \left(1 + \frac{\sqrt{n_2}}{\sqrt{n_2}}\right)\|\Sigma^{\frac{1}{2}}\subQ(\beta^* - \BB{\gamma}^*)\|_2^2 \\
                & = 2\alpha
        \end{aligned}
    \end{equation}
    with probability at least $1 - 2\exp(-cn_2)$. The control for (IV) can refer to the probability of event $G_{10}(t)$, and (V) to $G_9$ in Lemma \ref{lemma for lemma}.
\end{proof}

\subsubsection{Proof of Lemma \ref{lemma for lemma}}
\begin{proof}
\begin{itemize}
    \item $G_4^j(w_1, w_2, t)$: \\
    The probability of event $G_4^j(w_1, w_2,t)$ can be re-written based on the Adam's law:
    \[
     \PP\left(G_4^j(w_1, w_2, t)\right) = \E_{w_1, w_2}\left[\PP\left(G_4^j(w_1, w_2, t)|w_1, w_2\right)\right] \,\, \text{ for }j=1,2.
    \]
    As long as $w_1$ and $w_2$ are independent of $\epsilon\subaj$ and $X\subaj\supQ$ in $G_4^j(w_1, w_2,t)$, we can treat them as constant vectors in the conditional probability. Leveraging the proof for Lemma 11 in \cite{cai2020semiinfhighdim}, by taking their events $G_4(w_1,w_2,t)$, we have that 
    \[
    \PP\left(G_4^j(w_1, w_2, t)|w_1, w_2\right) \geq 1 -  2\exp(-ct^2)\text{ for }j=1,2.
    \]
    \[
    \implies \PP\left(G_4^j(w_1, w_2, t)\right) \geq 1-2\exp(-ct^2)\text{ for }j=1,2.
    \]
    \item $G_5(t)$: \\
     Note that $\XX\subaa\supQ\BBhat_0\in\mathbb{R}^{n_j\times L_0}$. Denote $\Phi$ as the orthonormal bases for the column space of $\BBhat_0$, where $\Phi = (\phi_1,...,\phi_r)\in\mathbb{R}^{n_2\times L_0}$ and $\Phi\supT\Phi=I_{L_0}$. Hense, there exists $\nu\in\mathbb{R}^{L_0}$ such that $\BBhat_0\delta=\Phi\nu$. If we normalize $\Phi\nu$ by dividing its norm, $\frac{\Phi\nu}{\|\Phi\nu\|_2}$ can be further re-written as $\Phi\nu^* = \sum_{j=1}^r \Phi_j \nu_j^*$ and $\sum_{j=1}^r (\nu_j^*)^2=1$. Then for any $\delta = \gamma_1 - \gamma_2$ where $\gamma_1, \gamma_2\in\Delta_{L_0}$, we have
    \begin{equation}
        \frac{n_2^{-1}\innerproduct{\epsilon\subaa\supQ}{\XX\subaa\supQ\BBhat_0\delta}}{\|\BBhat_0\delta\|_2} =n_2^{-1}\innerproduct{\epsilon\subaa\supQ}{\XX\subaa\supQ\Phi\nu^*} \leq \max_{u\in\mathbb{R}^{L_0}: \|u\|_2=1} n_2^{-1}\innerproduct{\epsilon\subaa\supQ}{\XX\subaa\supQ\Phi u}.
        \label{eq: thm3 condition1}
    \end{equation}
    Since $\E(\epsilon_i\supQ|\XX_i\supQ)$=0, $n_2^{-1}\innerproduct{\epsilon\subaa\supQ}{\XX\subaa\supQ\Phi u}$ is a summation of independent, centered sub-exponential random variables. We apply the Bernstein inequality in Corollary 5.17 in \citet{vershynin2010}, and derive that
    \begin{equation}
        \PP\left(|n_2^{-1}\innerproduct{\epsilon\subaa\supQ}{\XX\subaa\supQ\Phi u}|\geq \frac{t}{\sqrt{n_2}}\right) \leq \exp(-ct^2)
        \label{eq: thm3 condition2}
    \end{equation}
     Denote $\mathcal{U} = \{u\in\mathbb{R}^{L_0}: \|u\|_2=1\}$, which is a unit Euclidean sphere with $\mathcal{N}_{\kappa}$ as its $\kappa$-net. We apply the property of the $\kappa$-net, $|\mathcal{N}_{\kappa}| = c_0^{L_0}$ with $1< c_0 \leq 1+1/\kappa$, and derive that
    \[
    \max_{u\in\mathbb{R}^{L_0}: \|u\|_2=1} n_2^{-1}\innerproduct{\epsilon\subaa\supQ}{\XX\subaa\supQ\Phi u} \leq \frac{1}{1-\kappa}\cdot\max_{x\in \mathcal{N}_{\kappa}} n_2^{-1}\innerproduct{\epsilon\subaa\supQ}{\XX\subaa\supQ\Phi x} 
    \]
    Therefore, combining the result in \eqref{eq: thm3 condition1} and \eqref{eq: thm3 condition2}, 
    \begin{equation}
        \begin{aligned}
             &\PP\left(\max_{\delta = \gamma_1 - \gamma_2, \gamma_1, \gamma_2\in\Delta_{L_0}}\bigg|\frac{n_2^{-1}\innerproduct{\epsilon\subaa\supQ}{\XX\subaa\supQ\BBhat_0\delta}}{\frac{1}{\sqrt{n_2}}\|\BBhat_0\delta\|_2}\bigg| \geq t\right) \\
             &\leq \PP\left(\max_{u\in\mathbb{R}^{L_0}: \|u\|_2=1} \bigg|n_2^{-1}\innerproduct{\epsilon\subaa\supQ}{\XX\subaa\supQ\Phi u}\bigg| \geq \frac{t}{\sqrt{n_2}}\right) \\
             &\leq \PP\left(\max_{x\in \mathcal{N}_{\kappa}} \bigg|n_2^{-1}\innerproduct{\epsilon\subaa\supQ}{\XX\subaa\supQ\Phi x}\bigg|  \geq \frac{(1-\kappa)t}{\sqrt{n_2}}\right) \\
             &\leq \sum_{x\in\mathcal{N}_{\kappa}} \PP\left(\bigg|n_2^{-1}\innerproduct{\epsilon\subaa\supQ}{\XX\subaa\supQ\Phi x}\bigg|  \geq \frac{(1-\kappa)t}{\sqrt{n_2}}\right) \\
             &\leq (1+1/\kappa)^{L_0} \exp(-c(1-\kappa)^2t^2)
        \end{aligned}
    \end{equation}
    By taking $\kappa = 1/2$, $a=3^{L_0}$, $t = \sqrt{L_0}t_0$ in the above inequality, we obtain that
    \[
    \PP\left(\max_{\delta = \gamma_1 - \gamma_2, \gamma_1, \gamma_2\in\Delta_{L_0}}\bigg|\frac{n_2^{-1}\innerproduct{\epsilon\subaj\supQ}{\XX\subaa\supQ\BBhat_0\delta}}{\frac{1}{\sqrt{n_2}}\|\BBhat_0\delta\|_2}\bigg| \geq t_0\sqrt{L_0} \right) \leq a\exp\left(-\frac{ct_0^2L_0}{4} \right)
    \]

    \item $G_6$: \\
    To show the probability of event $G_6$, note the estimation of $\BBhat_0$ only uses target data from $\mathcal{A}_1$ and source estimators. Therefore, we have $\BBhat_0$ independent of $\XX\subaa\supQ$. 
    Recall event $G_4^2(\BBhat_0u, \BBhat_0u, t)$ where $\|u\|_2 =1$ satisfies that 
    \[
    1 - \PP\left(G_4^2(\BBhat_0u, \BBhat_0u, t)\right) = \PP\left(\bigg|n_2^{-1}\|X\subaa\supQ\BBhat_0 u\|_2^2 - \|\Sigma\subQ^{1/2}\BBhat_0 u\|_2^2\bigg| \geq \frac{t}{\sqrt{n_2}}\|\Sigma\subQ^{1/2}\BBhat_0 u\|_2^2\right) \leq \exp(-ct^2).
    \]
   Thus, 
    \begin{equation}
            \frac{\|\BBhat_0 u\|_2^2}{\frac{1}{n_2}\|X\subaa\supQ\BBhat_0 u\|_2^2} = \frac{\|\BBhat_0 u\|_2^2}{\frac{1}{n_2}\|X\subaa\supQ\BBhat_0 u\|_2^2 - \|\Sigma\subQ^{1/2}\BBhat_0 u\|_2^2 + \|\Sigma\subQ^{1/2}\BBhat_0 u\|_2^2} \lesssim \frac{\|\BBhat_0 u\|_2^2}{ (1 - t/\sqrt{n_2})\|\Sigma\subQ^{1/2}\BBhat_0 u\|_2^2}.
        \label{eq: lemma3 cond3}
    \end{equation}
    Since we assume $\Sigma\subQ$ satisfies that $\lambda_{\min}(\Sigma\subQ)\geq c_1 > 0$ in Assumption \ref{assumption: a_1}, the right hand side of \eqref{eq: lemma3 cond3} can be bounded by $\frac{1}{(1 - t/\sqrt{n_2})c_1}$. Therefore,
        \begin{equation}
            \PP\left(\bigg|\frac{\|\BBhat_0 u\|_2^2}{\frac{1}{n_2}\|X\subaa\supQ\BBhat_0 u\|_2^2}\bigg| \geq \frac{1}{(1 - t/\sqrt{n_2})c_1}\right) \leq \exp(-ct^2).
        \label{eq: lemma3 cond4}
    \end{equation}
    Similar to the proof for event $G_5$, we control the max of $\delta$ over the entire $(L_0)$-sphere by a $\kappa$-net:
        \begin{equation}
        \begin{aligned}
             &\PP\left(\max_{\delta = \gamma_1 - \gamma_2, \gamma_1, \gamma_2\in\Delta_{L_0}}\frac{\|\BBhat_0 \delta\|_2^2}{\frac{1}{n_2}\|X\subaa\supQ\BBhat_0 \delta\|_2^2} \geq \frac{1}{(1 - t/\sqrt{n_2})c_1} \right) \\
             &\leq \PP\left(\max_{u\in\mathbb{R}^{L_0}: \|u\|_2=1} \frac{\|\BBhat_0 u\|_2^2}{n_2^{-1}\|X\subaa\supQ\BBhat_0 u\|_2^2} \geq \frac{1}{(1 - t/\sqrt{n_2})c_1} \right) \\
             &\leq \PP\left(\max_{x\in \mathcal{N}_{\kappa}} \frac{\|\BBhat_0 x\|_2^2}{\frac{1}{n_2}\|X\subaa\supQ\BBhat_0 x\|_2^2} \geq \frac{1-\kappa}{(1 - t/\sqrt{n_2})c_1} \right) \\
             &\leq \sum_{x\in\mathcal{N}_{\kappa}} \PP\left(\frac{\|\BBhat_0 x\|_2^2}{\frac{1}{n_2}\|X\subaa\supQ\BBhat_0 x\|_2^2} \geq \frac{1-\kappa}{(1 - t/\sqrt{n_2})c_1}\right) \\
             &\leq (1+1/\kappa)^{L_0} \exp(-c(1-\kappa)^2t^2)
        \end{aligned}
    \end{equation}
     By taking $\kappa = 1/2$, $t = \frac{\sqrt{n_2}}{2}$ and $a = 3^{L_0}$, we have 
    \[
    \PP\left(\max_{\delta = \gamma_1 - \gamma_2, \gamma_1, \gamma_2\in\Delta_{L_0}}\frac{\|\BBhat_0 \delta\|_2^2}{\frac{1}{n_2}\|X\subaa\supQ\BBhat_0 \delta\|_2^2} \geq \frac{2}{c_1}\right) \leq a\exp\left(-\frac{cn_2}{4}\right)
    \]

    \item $G_7(t)$: \\
    For the event $G_7(t)$, we can decompose the left-hand side of inequality in $G_7(t)$ into:
    \begin{align*}
        \underbrace{\max_{\delta:= \gamma_1 - \gamma_2, \gamma_1,\gamma_2\in\Delta_{L}}\frac{\bigg|\frac{1}{n_1}\innerproduct{\epsilon\suba\supQ}{X\suba\supQ \BBhat\delta}\bigg|}{\frac{1}{n_1}\|X\suba\supQ\BBhat \delta\|_2}}_{\text{(I)}} \leq \underbrace{\max_{\delta:= \gamma_1 - \gamma_2, \gamma_1,\gamma_2\in\Delta_{L}}\frac{\bigg|\frac{1}{n_1}\innerproduct{\epsilon\suba\supQ}{X\suba\supQ \BBhat\delta}\bigg|}{\frac{1}{\sqrt{n_1}}\|\BBhat \delta\|_2}}_{\text{(II)}} \cdot \underbrace{\max_{\delta:= \gamma_1 - \gamma_2, \gamma_1,\gamma_2\in\Delta_{L}}\frac{\frac{1}{\sqrt{n_1}}\|\BBhat \delta\|_2}{\frac{1}{n_1}\|X\suba\supQ\BBhat \delta\|_2}}_{\text{(III)}}
    \end{align*}
    Then we can decompose the probability of (I) into:
    \begin{equation*}
        \begin{aligned}
            \PP\left(\text{(I)}\geq t_0\sqrt{L}\right) & \leq \PP\left(\text{(II)} \cdot \text{(III)}\geq t_0\sqrt{L}\right) \leq \PP\left(\text{(II)} \geq \frac{c_1t_0}{2}\sqrt{L}\right) + \PP\left(\text{(III)} \geq \frac{2}{c_1}\right) 
        \end{aligned}
    \end{equation*}
    The control over (II) can be proved similar to the proof for event $G_5$, since we have $\BBhat\in\R^{L}$ that is independent of $\epsilon\suba\supQ$ and $\XX\suba\supQ$ in (II), while in $G_5$ it is $\BBhat_0$ independent of $\epsilon\subaa\supQ$ and $\XX\subaa\supQ$. As long as the independence holds, the proof for $G_5$ can be directly applied to (II). Likewise, the proof for the probability of $G_6$ can be applied to (III). Therefore, 
    $$\PP\left(\text{(I)}\geq t_0\sqrt{L}\right) \leq a\exp\left(-\frac{c\cdot c_1^2t_0^2L}{16} \right) + a\exp\left(-\frac{cn_1}{4}\right).$$

    \item $G_8$: \\
    We first decompose the left-hand side of $G_8$ into
    \[
    \underbrace{\max_{\delta:= \gamma_1 - \gamma_2, \gamma_1,\gamma_2\in\Delta_{L}}\frac{n_2^{-1}\|\XX\subaa\supQ\BBhat \delta\|_2^2}{\frac{1}{n_1}\|\XX\suba\supQ\BBhat \delta\|_2^2}}_{\text{(I)}} \leq \underbrace{\max_{\delta:= \gamma_1 - \gamma_2, \gamma_1,\gamma_2\in\Delta_{L}}\frac{\|\BBhat \delta\|_2^2}{\frac{1}{n_1}\|\XX\suba\supQ\BBhat \delta\|_2^2}}_{\text{(II)}} \cdot \underbrace{\max_{\delta:= \gamma_1 - \gamma_2, \gamma_1,\gamma_2\in\Delta_{L}}\frac{n_2^{-1}\|\XX\subaa\supQ\BBhat \delta\|_2^2}{\|\BBhat \delta\|_2^2}}_{\text{(III)}}
    \]
     The control for (II) is similar to the strategy we use in event $G_6$, with replacement of $\widehat{\BB}_0$ to $\widehat{\BB}$ and $\XX\subaa\supQ$ to $\XX\suba\supQ$, which does not affect the independence between $\XX\suba\supQ$ and $\widehat{\BB}$. Similarly, (III) can be bounded by:
    \[        
    \frac{n_2^{-1}\|\XX\subaa\supQ\BBhat u\|_2^2}{\|\BBhat u\|_2^2} = \frac{n_2^{-1}\|\XX\subaa\supQ\BBhat u\|_2^2 - \|\Sigma\subQ^{1/2}\BBhat u\|_2^2 + \|\Sigma\subQ^{1/2}\BBhat u\|_2^2}{\|\BBhat u\|_2^2} \lesssim \frac{ (1 + t/\sqrt{n_2})\|\Sigma\subQ^{1/2}\BBhat u\|_2^2}{\|\BBhat u\|_2^2}.
    \]
    In order to control the max of $\delta$ over the entire $(L_0)$-sphere, we again introduce the $1/2$-net and take $t = \frac{n_2}{2}$, and eventually prove that,
    \begin{equation}
        \begin{aligned}
             \PP\left(\max_{\delta = \gamma_1 - \gamma_2, \gamma_1, \gamma_2\in\Delta_{L}} \frac{n_2^{-1}\|\XX\subaa\supQ\BBhat \delta\|_2^2}{\|\BBhat \delta\|_2^2} \geq \frac{3}{2}C_1\right) \leq a\exp\left(-\frac{cn_2}{4}\right)
        \end{aligned}
    \end{equation}
    As a consequence,
    \begin{align*}
        \PP\left(\text{(I)} \geq \frac{3C_1}{c_1}\right) 
        & \leq \PP\left(\text{(II)} \geq \frac{2}{c_1}\right) + \PP\left(\text{(III)} \geq \frac{3C_1}{2}\right)  \leq a\exp\left(-\frac{cn_1}{4}\right) + a\exp\left(-\frac{cn_2}{4}\right)
    \end{align*}

    \item $G_9$: \\
    Similar to the proof for (III) in event $G_8$, we can first bound $\frac{n_2^{-1}\|\XX\subaa\supQ(\BBhat - \BB)\gamma^*\|_2^2}{\|(\BBhat-\BB)\gamma^*\|_2^2}$ by:
    \begin{equation*}
        \begin{aligned}
            \frac{n_1^{-1}\|\XX\suba\supQ(\BBhat - \BB)\gamma^*\|_2^2}{\|(\BBhat-\BB)\gamma^*\|_2^2} &= \frac{n_1^{-1}\|\XX\suba\supQ(\BBhat - \BB)\gamma^*\|_2^2 - \|\Sigma\subQ^{1/2}(\BBhat - \BB)\gamma^*\|_2^2 + \|\Sigma\subQ^{1/2}(\BBhat - \BB)\gamma^*\|_2^2}{\|(\BBhat - \BB)\gamma^*\|_2^2} \\
            &\lesssim \frac{\|\Sigma\subQ^{1/2}(\BBhat - \BB)\gamma^*\|_2^2}{\|(\BBhat - \BB)\gamma^*\|_2^2} \\
            &\leq C_1,
        \end{aligned}
    \end{equation*}
    with probability at least $1-\exp(-cn_1)$. Likewise, we can show that $\frac{n_2^{-1}\|\XX\subaa\supQ(\BBhat - \BB)\gamma^*\|_2^2}{\|(\BBhat-\BB)\gamma^*\|_2^2} \lesssim C_1$ with probability at least $1-\exp(-cn_2)$. As a result, 
    \[
    \max\bigg\{\frac{n_1^{-1}\|\XX\suba\supQ(\BB - \BBhat){\gamma}^*\|_2^2}{\|(\BB - \BBhat){\gamma}^*\|_2^2} , \frac{n_2^{-1}\|\XX\subaa\supQ(\BB - \BBhat){\gamma}^*\|_2^2}{\|(\BB - \BBhat){\gamma}^*\|_2^2}\bigg\} \lesssim C_1
    \]
    with probability at least $1-\exp(-cn_1)-\exp(-cn_2)$.
    
    \item $G_{10}(t)$: \\
    We first aim to control $\frac{1}{n_1}\|\XX\suba\BBhat({\gamma}^* - \widehat{\gamma}\suba)\|_2^2$. Since $\widehat{\gamma}\suba$ is the minimizer of $\frac{1}{n_1}\|\YY\suba\supQ - \XX\suba\supQ \BBhat\gamma\|_2^2$ over any $\gamma\in\Delta^L$, we have
        \begin{align*}
            \frac{1}{n_1}\|\YY\suba\supQ - \XX\suba\supQ \BBhat{\gamma}^*\|_2^2 &\geq \frac{1}{n_1}\|\YY\suba\supQ - \XX\suba\supQ \BBhat\widehat{\gamma}\suba\|_2^2  \\
            \iff \frac{1}{n_1}\|\YY\suba\supQ - \XX\suba\supQ \BBhat{\gamma}^*\|_2^2 &\geq \frac{1}{n_1}\|\YY\suba\supQ - \XX\suba\supQ\BBhat{\gamma}^* + \XX\suba\supQ\BBhat({\gamma}^* - \widehat{\gamma}\suba)\|_2^2 \\
            \iff \frac{1}{n_1}\|\XX\suba\supQ\BBhat({\gamma}^* - \widehat{\gamma}\suba)\|_2^2 &\leq \frac{2}{n_1}\innerproduct{\YY\suba\supQ - \XX\suba\supQ\BBhat{\gamma}^*}{\XX\suba\supQ\BBhat(\widehat{\gamma}\suba - {\gamma}^*)} \\
            &= \underbrace{\frac{2}{n_1}\innerproduct{\XX\suba\supQ(\beta^* - \BBhat{\gamma}^*)}{\XX\suba\supQ\BBhat(\widehat{\gamma}\suba - {\gamma}^*)}}_{\text{(I)}} +
            \underbrace{\frac{2}{n_1}\innerproduct{\epsilon\suba}{\XX\suba\supQ\BBhat(\widehat{\gamma}\suba - {\gamma}^*)} }_{\text{(II)}}. \stepcounter{equation}\tag{\theequation}\label{eq: G3 proof 1}
        \end{align*}
    According to the Cauchy-Schwarz inequality, 
    \begin{equation}
        \text{(I)}\leq \frac{2}{n_1}\|\XX\suba\supQ(\beta^* - \BBhat{\gamma}^*)\|_2\|\XX\suba\supQ\BBhat({\gamma}^* - \widehat{\gamma}\suba)\|_2.    
         \label{eq: G3 proof 2}
    \end{equation}
    Using the result for the event $G_7\left(t\right)$, when we take $\delta = \widehat{\gamma}\suba - {\gamma}^*$, with probability at least $1 - a\exp\left(-\frac{c\cdot c_1^2t^2L}{16} \right) - a\exp\left(-\frac{cn_1}{4}\right)$, (II) can be re-written as
    \begin{equation}
        \begin{aligned}
            \frac{2}{n_1}\innerproduct{\epsilon\suba}{\XX\suba\supQ\BBhat(\widehat{\gamma}\suba - {\gamma}^*)} &\lesssim  \frac{2t\sqrt{L}}{\sqrt{n_1}}\cdot \frac{1}{\sqrt{n_1}}\|X\supQ\suba\BBhat({\gamma}^* - \widehat{\gamma}\suba)\|_2
        \end{aligned}
         \label{eq: G3 proof 3}
    \end{equation}
    Combining \eqref{eq: G3 proof 1}, \eqref{eq: G3 proof 2} and \eqref{eq: G3 proof 3}, 
    \begin{equation}
        \begin{aligned}
            \frac{1}{n_1}\|\XX\suba\supQ\BBhat({\gamma}^* - \widehat{\gamma}\suba)\|_2^2 &\lesssim 
            \frac{1}{n_1}\|\XX\suba\supQ(\beta^* - \BBhat{\gamma}^*)\|_2\|\XX\suba\supQ\BBhat({\gamma}^* - \widehat{\gamma}\suba)\|_2 +  \\
            &\;\;\;\;\;\frac{2t\sqrt{L}}{\sqrt{n_1}} \cdot \frac{1}{\sqrt{n_1}}\|X\supQ\suba\BBhat({\gamma}^* - \widehat{\gamma}\suba)\|_2 \\
            \iff \frac{1}{\sqrt{n_1}}\|X\supQ\suba\BBhat({\gamma}^* - \widehat{\gamma}\suba)\|_2 &\lesssim \frac{1}{\sqrt{n_1}}\|\XX\suba\supQ(\beta^* - \BBhat{\gamma}^*)\|_2 + \frac{2t\sqrt{L}}{\sqrt{n_1}} \\
            &\leq \underbrace{\frac{1}{\sqrt{n_1}}\|\XX\suba\supQ(\beta^* - \BB{\gamma}^*)\|_2}_{\text{(III)}} + \underbrace{\frac{1}{\sqrt{n_1}}\|\XX\suba\supQ(\BB - \BBhat){\gamma}^*\|_2}_{\text{(IV)}} + \frac{2t\sqrt{L}}{\sqrt{n_1}}
        \end{aligned}
        \label{eq: G3 proof 4}
    \end{equation}
    Applying the result for event $G_4(w_1,w_2, \sqrt{n_1})$ again with $w_1=w_2=\beta^* - \BB\gamma^*$, with probability at least $1-2\exp(-cn_1)$,
    \begin{equation}
        \text{(III)}^2 \lesssim \left(1 + \frac{\sqrt{n_1}}{\sqrt{n_1}}\right) (\beta^* - \BB{\gamma}^*)\supT\Sigma\subQ(\beta^* - \BB{\gamma}^*) = 2\alpha
        \label{eq: G3 proof 5}
    \end{equation}
   The upper bound for (IV) can refer to event $G_9$. To sum up, with probability at least $1 - 2\exp(-cn_1) - a\exp\left(-\frac{c\cdot c_1^2t_0^2L}{16} \right) - a\exp\left(-\frac{cn_1}{4}\right)$,
    \begin{equation}
        \frac{1}{n_1}\|\XX\suba\supQ\BBhat({\gamma}^* - \widehat{\gamma}\suba)\|_2^2 \lesssim 2\alpha + \|(\BBhat - \BB)\gamma^*\|_2^2 + \frac{4t^2L}{n_1}.
        \label{eq: G3 proof 6}
    \end{equation}
    According to event $G_8$, by taking $\delta = \gamma^* - \widehat{\gamma}\suba$, we can derive that  
    \[
    n_2^{-1}\|\XX\subaa\supQ\BBhat({\gamma}^* - \widehat{\gamma}\suba)\|_2^2 \lesssim \frac{3C_1}{c_1}\cdot\frac{1}{n_1}\|\XX\suba\supQ\BBhat({\gamma}^* - \widehat{\gamma}\suba)\|_2^2 \lesssim  2\alpha + \|(\BBhat - \BB)\gamma^*\|_2^2 + \frac{4t^2L}{n_1}
    \]
    \end{itemize}
    
\end{proof} 

\subsubsection{Proof of Corollary \ref{Cor: low-dim}}
\begin{proof}
    In this proof we aim to derive that $\|(\BB - \BBhat)\gamma^*\|_2^2$ in \eqref{eq: thm 3} has an upper bound $\sum_{l=1}^L (\gamma_l^*)^2\frac{p}{N_l}\sigma_l^2$ with a high probability in the low dimension case. Based on the Markov inequality, for any $t>0$,
    \[
    \PP\left(\|(\BB - \BBhat)\gamma^*\|_2^2 \geq t\E\left(\|(\BB - \BBhat)\gamma^*\|_2^2\right)\right) \leq \frac{1}{t}.
    \]
    Recall that the OLS estimator under the linear model \eqref{eq: source model} has the explicit form $$\widehat{b}\supl = \left((\XX\supl)\supT \XX\supl\right)^{-1}(\XX\supl)\supT \YY\supl.$$ 
    Thus, $\E\left(\|(\BB - \BBhat)\gamma^*\|_2^2\right)$ can be further expanded as:
    \begin{align*}
        \E\left(\|(\BB - \BBhat)\gamma^*\|_2^2\right) &=\E\left(\bigg\|\sum_{l=1}^L\gamma_l^*(\widehat{b}\supl - b\supl)\bigg\|_2^2\right) \\
        &= \sum_{l=1}^L(\gamma_l^*)^2 \E\left(\|\widehat{b}\supl - b\supl\|_2^2\right)  \\
        &= \sum_{l=1}^L(\gamma_l^*)^2 tr\left(Var(\left((\XX\supl)\supT \XX\supl\right)^{-1}(\XX\supl)\supT \YY\supl)\right)  \\
        &= \sum_{l=1}^L(\gamma_l^*)^2 \sigma_l^2 tr\left(\E(\left((\XX\supl)\supT \XX\supl\right)^{-1})\right)  \\
        &= \sum_{l=1}^L(\gamma_l^*)^2 \frac{\sigma_l^2}{N_l}\E\left(tr(\left((\XX\supl)\supT \XX\supl/N_l\right)^{-1})\right)  \\
    \end{align*}
Denote $\widehat{\Sigma}\supl := \frac{1}{N_l}(\XX\supl)\supT \XX\supl$ and its $j$-th eigenvalue as $\lambda_j^{-1}(\widehat{\Sigma}\supl)$. We have that $$tr((\widehat{\Sigma}\supl)^{-1}) = \sum_{j=1}^p \lambda_j^{-1}(\widehat{\Sigma}\supl) \leq p\lambda_{\min}^{-1}(\widehat{\Sigma}\supl).$$
According to Theorem 5.39 in \citep{vershynin2010}, with independent rows in $\XX\supl$, the minimum singular value of its sample covariance matrix satisfies $s_{\min}(\widehat{\Sigma}\supl) = \sqrt{\lambda_{\min}(\widehat{\Sigma}\supl)} \geq \sqrt{N_l} - C\sqrt{p} - t_0$ with probability at least $1 - 2\exp(-ct_0^2)$ for every $t_0>0$. Taking $t_0 = \sqrt{p}$ and assuming $N_l\geq (C+1)p + 1$, we apply this lower bound for $\lambda_{\min}(\widehat{\Sigma}\supl)$ and eventually obtain the upper bound for $\|(\BB - \BBhat)\gamma^*\|_2^2$:
\[
\PP\left(\|(\BB - \BBhat)\gamma^*\|_2^2 \geq \sum_{l=1}^L(\gamma_l^*)^2 \sigma_l^2 \frac{tp}{N_l} \right) \leq \frac{1}{t} + 2\exp(-cp).
\]

\end{proof}

\subsubsection{Proof of Corollary \ref{Cor: high-dim}}
\begin{proof}
       In this proof we aim to derive that $\|(\BB - \BBhat)\gamma^*\|_2$ in \eqref{eq: thm 3} has an upper bound $\sum_{l=1}^L \gamma_l^*\sqrt{\frac{k_l\log p}{N_l}\sigma_l^2}$ with a high probability in the high dimension case. We first reformulate $(\BB-\BBhat){\gamma}^*$ as a convex combination of source coefficients and upper bound it based on the triangle inequality:
    \begin{equation*}
        \begin{aligned}
            \|(\BB - \BBhat){\gamma}^*\|_2 = \|\sum_{l=1}^L (b^{(l)} - \widehat{b}^{(l)}){\gamma_l}^*\|_2 \leq \sum_{l=1}^L {\gamma_l}^*\|b^{(l)} - \widehat{b}^{(l)}\|_2 
        \end{aligned}
    \end{equation*}
    According to assumption \ref{assumption: b_1}, $\|b^{(l)} - \widehat{b}^{(l)}\|_2 \lesssim \sqrt{\frac{k_l\log p}{N_l}\sigma_l^2} := \sqrt{g(N_l)}$ for $l=1,...,L$. Therefore, 
    \begin{equation*}
        \begin{aligned}
            \PP\left(\|(\BB - \BBhat){\gamma}^*\|_2 > \sum_{l=1}^L \gamma_l^*\sqrt{g(N_l)}\right) &\leq \PP\left(\sum_{l=1}^L \gamma_l^*\|b^{(l)} - \widehat{b}^{(l)}\|_2 > \sum_{l=1}^L \gamma_l^*\sqrt{g(N_l)}\right)  \\
            &\leq \PP\left(\bigcup_{l=1}^L \left\{\|b^{(l)} - \widehat{b}^{(l)}\|_2 > \sqrt{g(N_l)}\right\}\right) \\
            &\leq \sum_{l=1}^L \PP\left(\|b^{(l)} - \widehat{b}^{(l)}\|_2 > \sqrt{g(N_l)}\right) \\
            &\leq \sum_{l=1}^L \delta_l(N_l).
        \end{aligned}
    \end{equation*}
   Therefore, we derive that $\|(\BB - \BBhat){\gamma}^*\|_2^2 \lesssim (\sum_{l=1}^L \gamma_l^*\sqrt{g(N_l)})^2$ with probability at least $1 - \sum_{l=1}^L \delta_l(N_l)$.
    
\end{proof}

\subsubsection{Proof of Corollary \ref{corol: ideal case}}
In the ideal case where the $L^*$ sources share the same model as the target, and under the low-dimensional setting, the optimal weight vector $\gamma^*$ defined in \eqref{eq: best source selection} simplifies to an inverse-variance form: $\gamma_l^* = \frac{I(l\in\mathcal{A}0)\cdot N_l}{\sum{j\in\mathcal{A}_0} N_j}$. As a result, the second term in the minimand of \eqref{eq: cor low dim rate} can be rewritten as
$$\sum_{l=1}^L (\gamma_l^*)^2 \frac{p}{N_l} = \sum_{l\in\mathcal{A}_0} \big(\frac{N_l}{\sum_{j\in\mathcal{A}_0}N_j}\big)^2 \cdot \frac{p}{N_l} = \frac{p}{\sum_{l\in\mathcal{A}_0}N_l}.$$
Note that $\min\{\frac{p}{n}, \frac{p}{\sum_{l\in\mathcal{A}_0} N_l}\}$ in the rate of GART and $\frac{p}{\sum_{l\in\mathcal{A}_0}N_l + n}$ in that of TransLasso are in the same order as $\frac{p}{\sum_{l\in\mathcal{A}_0}N_l}$, given that $n\ll\sum_{l\in\mathcal{A}_0}N_l$. The only difference lies in $L/n$ and $\log L/n$. When $L$ is bounded, such a difference is ignorable.
    
Under the high-dimension case, as shown in Corollary \ref{Cor: high-dim}, with the same optimal inverse-variance weighting $\gamma^*$, the second term inside the minimand of the GART's rate becomes:
\begin{equation}
    \begin{aligned}
    \left(\sum_{l=1}^L \gamma_l^* \sqrt{\frac{k_l\log p}{N_l}}\right)^2  = k\log p\cdot \left( \frac{\sum_{l\in\mathcal{A}_0}\sqrt{N_l}}{\sum_{l\in\mathcal{A}_0}N_l}\right)^2 \leq \frac{k\log p}{\frac{1}{L^*}\sum_{l\in\mathcal{A}_0}N_l}.
    \end{aligned}
\label{eq: L* more}
\end{equation}

\subsubsection{Proof of Proposition \ref{prop: relationship between alpha and h}}
\label{sec: proof of prop 2}
Define $\gamma^h$ with $\gamma^h_l = \mathbb{I}(l\in\mathcal{A}_h)/|\mathcal{A}_h|$ for $l=1,...,L$. For a given $h$ satisfying that $|\mathcal{A}_h|>0$,
    \begin{align*}
        \alpha &= \min_{\gamma\in\Delta_L} (\beta^*-\BB\gamma)\trans\Sigma\subQ(\beta^*-\BB\gamma^*) \\
        & \leq (\beta^*-\BB\gamma^h)\trans\Sigma\subQ(\beta^*-\BB\gamma^h) \\
        & \leq \lambda_{\max}(\Sigma\subQ)\|\beta^*-\BB\gamma^h\|_2^2 \\
        & \leq \lambda_{\max}(\Sigma\subQ)\|\beta^*-\BB\gamma^h\|_1^2 \\
        & \leq \lambda_{\max}(\Sigma\subQ) h^2.
    \end{align*}
    Since $\lambda_{\max}(\Sigma\subQ)$ is smaller than $C_1$ based on our Assumption 2, $\alpha$ has an upper bound in the same order of $h^2$.

\section{Additional Discussion on Initial Estimators}
\subsection{Other Possible Estimators}
\label{supp: initial estimator}
One consequence of including $\widehat{\beta}\subaj$ in $\BBhat_{0\mathcal{A}_j}$ is that $\widehat{\beta}\subinit\supweighted$ is a consistent estimator of $\beta^*$, whereas $\widehat{\beta}\supsource\subinit$ becomes biased in terms of $\beta^*$ when the source mixing assumption is violated. Though a consistent initial is preferable if aiming for transferability, an initial shifting towards the source hull could be more appealing when prior knowledge indicates that future data relates more to the existing sources, and we focus on generalizability. To balance the two needs, we design a mixed initial that lies in between $\widehat{\beta}\subinit\supsource$ and $\widehat{\beta}\subinit\suplinear$:
\begin{equation}
    \widehat{\beta}\subinit\supmixST(\omega) = \argmin_{\beta} n^{-1}\|\YY\supmixST - \XX\supQ\beta\|_2^2, \,\,\, \YY\supmixST=[\XX\supQ_{\mathcal{A}_{\omega}}\widehat{\beta}\subinit\suplinear, \XX\supQ_{\mathcal{A}^c_{\omega}}\widehat{\beta}\subinit\supsource],
        \label{eq: def of mixST init beta}
\end{equation}
where $\mathcal{A}_{\omega}$ contains $\lceil \omega n \rceil$ randomly sampled rows and $\mathcal{A}_{\omega}^c = \{1,...,n\}/\mathcal{A}_{\omega}$. $\omega\in[0,1]$ quantifies the need for transferability over generalizability towards the source mixture. Another possible initial mixes $\widehat{\beta}\subinit\suplinear$ and $\mathbf{0}$, serving to balance transferability and overall generalizability toward zero:
\begin{equation}
    \widehat{\beta}\subinit\supmixT0(\omega) = \argmin_{\beta} n^{-1}\|\YY\supmixT0 - \XX\supQ\beta\|_2^2, \,\,\, \YY\supmixT0=[\XX\supQ_{\mathcal{A}_{\omega}}\widehat{\beta}\subinit\suplinear, \XX\supQ_{\mathcal{A}^c_{\omega}}\mathbf{0}].
    \label{eq: def of mixT0 init beta}
\end{equation}

\subsection{Theoretical Results of Choosing a More Accurate Initial}
In the following theorem, we shall demonstrate the benefits of choosing a more informative initial estimator based on the available target data compared to the other less accurate estimators, e.g., the zero initial estimator. 
\setcounter{Theorem}{3}
\begin{Theorem} Consider the linear models \eqref{eq: source model} and \eqref{eq: target model} and suppose that Assumptions \ref{assumption: a_1, source}, \ref{assumption: a_1} and \ref{assumption: b_1} hold. Additionally, assume $\tau\geq\min\left\{\frac{k\log p}{n}\sigma\subQ^2, \alpha + \left(\sum_{l=1}^L \gamma_l^* \sqrt{\frac{k_l\log p}{N_l}\sigma_l^2}\right)^2\right\}$. Consider two different initial estimators $\widehat{\beta}\subinit^0$ and $\widehat{\beta}\subinit^1$, and their corresponding GART estimator $\widehat{\beta}\subGART^0:=\widehat{\beta}\subGART(\widehat{\beta}\subinit^0, \tau)$ and $\widehat{\beta}\subGART^1:=\widehat{\beta}\subGART(\widehat{\beta}\subinit^1, \tau)$. Define the difference in their estimation error as $\delta(\widehat{\beta}\subinit^0, \widehat{\beta}\subinit^1; \XX, \tau) := \frac{1}{n}\|\XX(\widehat{\beta}_G^0-\beta^*)\|_2^2 - \frac{1}{n}\|\XX(\widehat{\beta}_G^1 -\beta^*)\|_2^2$. Then there exists positive constants $C>0$ and $c>0$ such that, with probability at least {$1-\exp(-c(t^2\cdot L+n)) - \delta(n)-\sum_{l=1}^L \delta_l(N_l)$} for $t\geq C$,
    \begin{equation}
        -\frac{\sqrt{\tau}}{\sqrt{n_2}}\|\XX\subaa\supQ(\widehat{\beta}\subinit^1-\beta^*)\|_2 \lesssim \delta(\widehat{\beta}\subinit^0, \widehat{\beta}\subinit^1; \XX\subaa, \tau)
        \lesssim \frac{\sqrt{\tau}}{\sqrt{n_2}}\|\XX\subaa\supQ(\widehat{\beta}\subinit^0-\beta^*)\|_2.
    \label{eq: thm4 bound for est error}
    \end{equation}
    Also, $\delta(\widehat{\beta}\subinit^0, \widehat{\beta}\subinit^1; \XX, \tau)$ is strictly positive when the following condition is met:
    \begin{equation}
        \frac{1}{\sqrt{n_2}}\|\XX\subaa\supQ(\widehat{\beta}_G^0-\widehat{\beta}_G^1)\|_2 > 2\left(\frac{1}{\sqrt{n_2}}\|\XX\subaa\supQ(\widehat{\beta}\suba-\beta^*)\|_2  + \frac{1}{\sqrt{n_2}}\|\XX\subaa\supQ(\widehat{\beta}\subinit^1-\beta^*)\|_2\right).
        \label{eq: thm4 strict positive condition}
    \end{equation}
\label{thm: second guidance}
\end{Theorem}

The proof has been shown in Section \ref{sec: proof for thm initial}. When $\tau$ is approaching zero, the upper and lower bounds in \eqref{eq: thm4 bound for est error} converge to zero, indicating the performance among different GART estimators is similar to each other for $\tau\rightarrow 0.$ For a particularly small $\tau$, the guidance from the constraint set $\widehat{\mathcal{S}}(\tau)$ will dominate the guidance from the initial estimators. On the other hand, when $\tau$ is relatively large and $\widehat{\beta}\subinit^1$ is a consistent estimator, the lower bound in \eqref{eq: thm4 bound for est error} converges to zero while the upper bound can diverge with an inaccurate initial estimator $\widehat{\beta}\subinit^0$. Thus, the estimation error will likely decrease by using a more accurate initial estimator $\widehat{\beta}\subinit^1$. 

 To justify the feasibility of the condition that guarantees the improved estimation error by using $\widehat{\beta}\subinit^1$ as an initial compared to $\widehat{\beta}\subinit^0$, we first examine the term $\frac{1}{\sqrt{n_2}}\|\XX\subaa\supQ(\widehat{\beta}_G^0-\widehat{\beta}_G^1)\|_2$ on the left-hand side of \eqref{eq: thm4 strict positive condition}. This term can be relatively large, or even converge to a positive constant, when $\widehat{\beta}\subinit^0$ is an inconsistent estimator of $\beta^*$ (e.g., zero initial). In contrast, the right-hand side of \eqref{eq: thm4 strict positive condition} converges to zero given $\widehat{\beta}\subinit^1$ is consistent. Therefore, in this scenario, a better initial estimator leads to a strict improvement compared to using a less informative initial estimator. 

\def\bE{\mathbf{E}}
\section{Additional Simulations}
\label{sec: appendix simulation}
\subsection{Additional Setting Details}
\label{sec: appendix simulation desc}
Across all simulation studies, we generate $\XX_i\supl = \mu\subPl^X + \bE_i\supl$ with $\bE_i\supl \sim N(0, \Sigma\subPl)$, $\XX_i\supQ = \bE_i\supQ$ with $\bE_i\supQ \sim N(0, \Sigma\subQ)$, where $(\Sigma\subPl)_{k,k'} = 0.6^{|k-k'|} + I(k=k')\times N(0, 0.01)$, $(\Sigma_{\Tar})_{k,k'} = 0.7^{|k-k'|}$, 
and $\mu\subPl^X \sim \mbox{exponential}(1) + N(0, 0.01)$. We generate the outcomes from linear models with different choices of $\{b\supl, l=1,...,L\}$ and $\beta^*$ under different settings, but generate error distributions from $\epsilon_i\supl\sim N(0, \sigma_{\mathbb{P}}^2)$ and $\epsilon_i\supQ\sim N(0, \sigma_{\Tar}^2)$ across all settings, where $\sigma\subP^2=0.5$ and $\sigma\subQ = 1$. To evaluate the performance of different estimators,  we generate independent validation datasets $\{(Y_i,\XX_i), i=1,...,N\subvalid\}$ with $Y_i = \XX_i\trans\beta\subvalid + \bE_i$, where $\XX_i$ and $\bE_i$ conform to the same distribution as $\XX_i\supQ$ and $\bE_i\supQ$, respectively.

\subsubsection{Setting 2}
In {\bf setting 2}, we let
$b^{(1)} = [0.2, 0.05, 0.4, -0.15, -0.6, 0, 0.2, 0, 0, 0, 0.3, 0, 0, 0, 0.4, 0, 0, 0, -1, 0, 0, 0.2,\allowbreak 0, 0, 0, 0.3, 0, 0, 0, 0.4, 0, 0, 0, -1, 0]$; $b^{(2)} = [0.3, 0.2, 0.6, -0.3, -0.6, 0, 0, 0.2, 0, 0, 0, 0.3, 0, 0, 0, 0.4,\allowbreak 0, 0, 0, 1,0, 0, 0.2, 0, 0, 0, 0.3, 0, 0, 0, 0.4, 0, 0, 0, 1]$; $b^{(3)} = [-0.2, -0.1, -0.5, 0.2, 0.6, 0,
    0, 0, 0.2, 0, 0, 0, 0.3,\allowbreak 0, 0, 0, 0.4, 0, 0, 0, 0, 0, 0, 0.2, 0, 0, 0, 0.3, 0, 0, 0, 0.4, 0, 0, 0]$; $b^{(4)} = [0.3, 0.1, 0.5, -0.2, -0.7, 0.2, 0, 0, 0, 0.3,\allowbreak 0, 0, 0, 0.4, 0, 0, 0, 1, 0, 0, 0.2, 0, 0, 0, 0.3, 0, 0, 0, 0.4, 0, 0, 0, 1, 0, 0]$.
We then define $\beta^*$ as a mixture of the first 3 sources plus a deviation in a specified direction $\bold{a}$, which increases with parameter $\kappa$:
$\beta^* = \frac{1}{3}\left(b^{(1)} + b^{(2)} + b^{(3)}\right) + \frac{\kappa}{\sqrt{35}} \bold{a}$. The perturbation vector $\bold{a}$ consists of $1$ or $-1$ as its elements: $\bold{a} = [-1,1,-1,-1,1,-1,1,-1,-1,1,-1,1,-1,-1,1,-1,1,-1,-1,1,-1,1,-1,-1,\allowbreak1,-1,1,-1,-1,1,-1,1,-1,-1,1]$. We vary $\kappa$ from $0.006$ to $3.25$ to show the model performance when the target effect deviates from the combination of sources. 

\subsubsection{Setting 3}
\label{sec: appendix sim highd}
In setting 3, we aim to show the model performance with different sparsity levels. $p=307, L=10, n=100$, $\beta^* = (0.3, 0.1, 0.5, -0.2, -0.7, 0, 0, \underbrace{0,...,0}_{\times 150})$. Each $b\supl$ has the form:
\[
b\supl = (0.3, 0.1, 0.5, -0.2, -0.7, 0, 0, \underbrace{(-1)^{a_1}\frac{\kappa}{s_0},...,(-1)^{a_{s_0}}\frac{\kappa}{s_0}}_{\times s_0}, \underbrace{0,...,0}_{\times 150-s_0}).
\]
$\kappa$ controls the distance between the target and each source. Also, $s_0$ and $\kappa$ control the $l_0$ and $l_1$-level sparsity of $b\supl$, respectively. $s_0$ is fixed as $50$ and $\kappa$ varies from 1 to 35.

\subsection{Additional Simulation Results}
\label{sec: additional simu result}

\subsubsection{More Results Using Multi-task Learning Methods}
\label{sec: multi-task}
\begin{figure}[H]
    \centering
\includegraphics[width=1\linewidth]{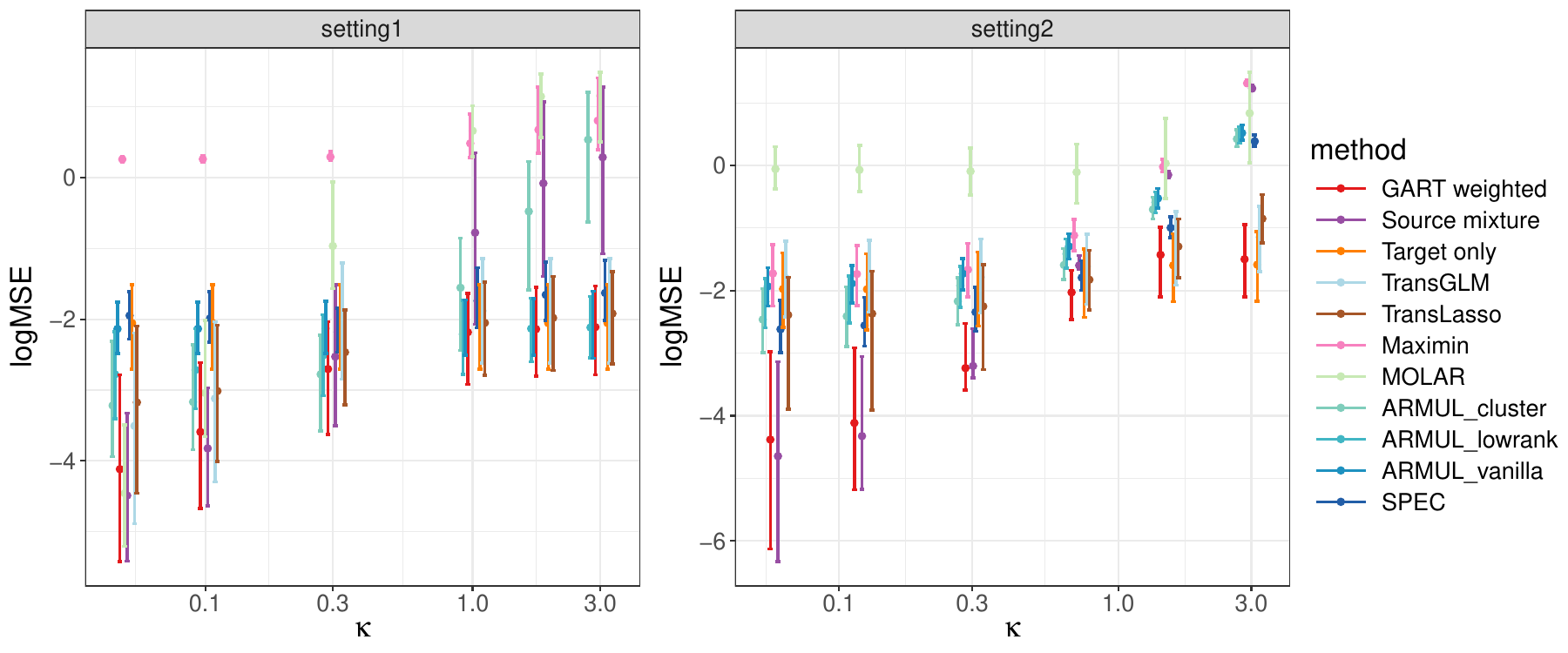}
    \caption{New Figure 3 including multi-task learning benchmarks}
    \label{fig: new fig3}
\end{figure}

We included comparison between GART and more multi-task learning benchmarks in this section. New benchmarks include MOLAR (Median-based Multitask Estimator for Linear Regression) \citep{huang2025optimal}, three versions (vanilla, clustered, low-rank) of ARML (Adaptive and robust multi-task learning) \citep{duan2023adaptive}, and
SPEC (Spectral multi-task learning) \citep{tian2025learningsimilarlinearrepresentations}. These results provide a clearer empirical assessment of how GART compares to multi-task learning methods that borrow strength through a shared latent center or sparse deviations across tasks. In setting 1 (Figure \ref{fig: new fig3}, left subplot), when the source tasks closely resemble the target (e.g. source-target discrepancy parameter is 0), multitask learning methods such as MOLAR outperforms the target-only estimator and achieve performance comparable to GART. However, as source heterogeneity increases, their performance deteriorates quickly and often becomes similar to  or worse than the target-only estimator. In setting 2 (Figure \ref{fig: new fig3}, right subplot) where each source differs from the target but the overall source-mixing discrepancy remains small, multi-task exhibit large MSE, especially for MOLAR, whereas GART continues to effectively leverage source information and achieves a much smaller MSE. 

In the model-generalizability settings (Settings 4, Figure \ref{fig: new fig4} below), where the validation population differs from the training target population, multitask learning methods tend to outperform the target-only estimator when the distribution shift is large (large $\nu$), except for MOLAR, which exhibits a much worse behavior that we choose not to show here. When the shift is small (small $\nu$), multitask learning methods are often worse than the target-only model. Overall, these results suggest that while multi-task learning can be competitive under strong similarity assumptions, GART provides more stable performance across varying levels of heterogeneity and distribution shift, supporting its robustness as a practical transfer learning strategy.

\begin{figure}[H]
    \centering
\includegraphics[width=.6\linewidth]{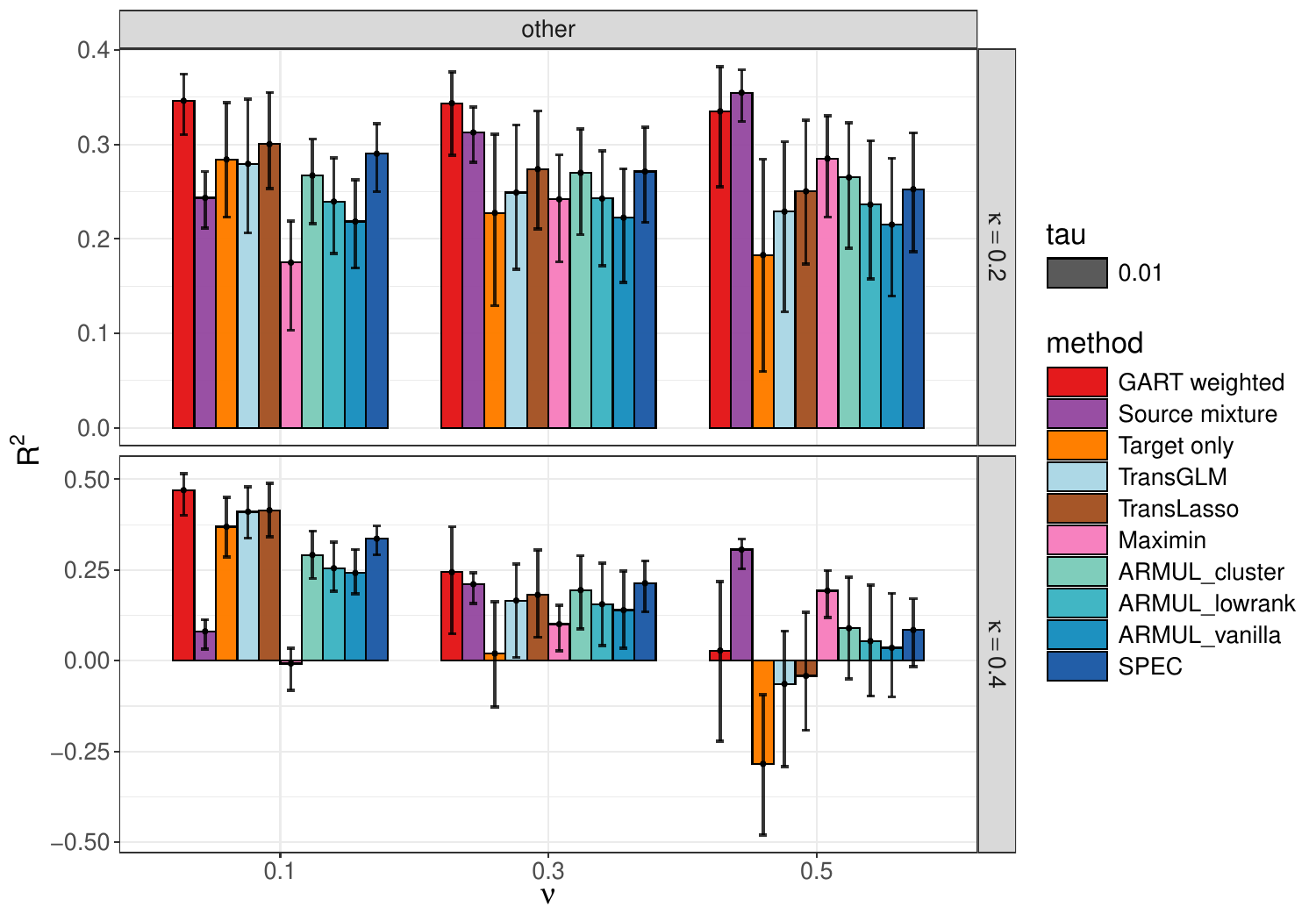}
    \caption{New Figure 4 including multi-task learning benchmarks}
    \label{fig: new fig4}
\end{figure}

\subsubsection{Setting 3}
\begin{figure}[H]
    \centering
    \includegraphics[width=.5\linewidth, page=1]{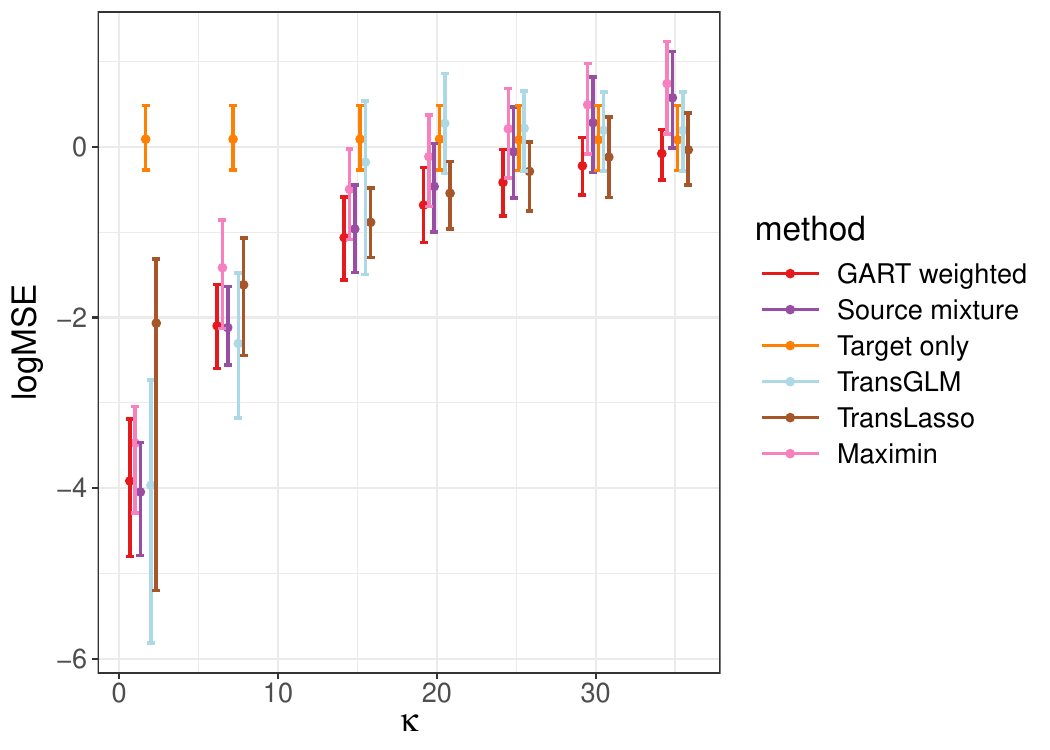}
    \caption{Empirical mean and 95\% confidence interval of $\log \text{MSE}$ when $n=200$ for setting 3. A growing $\kappa$ indicates a larger distance between each source and the target.}
    \label{fig: simulation setting3}
\end{figure}

In setting 3, 50 variables have either positive or negative effects on the outcome among sources, but have null effects for the target outcome. A smaller $s_1$ not only represents a diminishing gap between $\beta^*$ and the source convex hull, but also a closer distance between $\beta^*$ and each $b\supl$. That is, $\kappa$ in this setting is similar to $\kappa$ in setting 1, but in a high-dimension case. As a result, Figure \ref{fig: simulation setting3} resembles the left one in Figure \ref{fig: simulation setting1-2 new}. With a smaller $s_1$ (i.e., all sources are transferrable), all methods show superior performance compared to the target-only estimator. When $s_1$ grows, higher mse for TransGLM and TransLasso indicates the insufficient usage of source data, while the source mixture and the Maximin estimator experience a performance drop due to the violation of the source mixing assumption. On the other hand, GART with the weighted combination initial estimator achieves the lowest MSE especially for non-zero $\kappa$'s, suggesting a robust model performance in the high-dimension case.

\subsubsection{Setting 4.2}
\begin{figure}[t]
    \centering
 \includegraphics[width=1\linewidth, page=2]{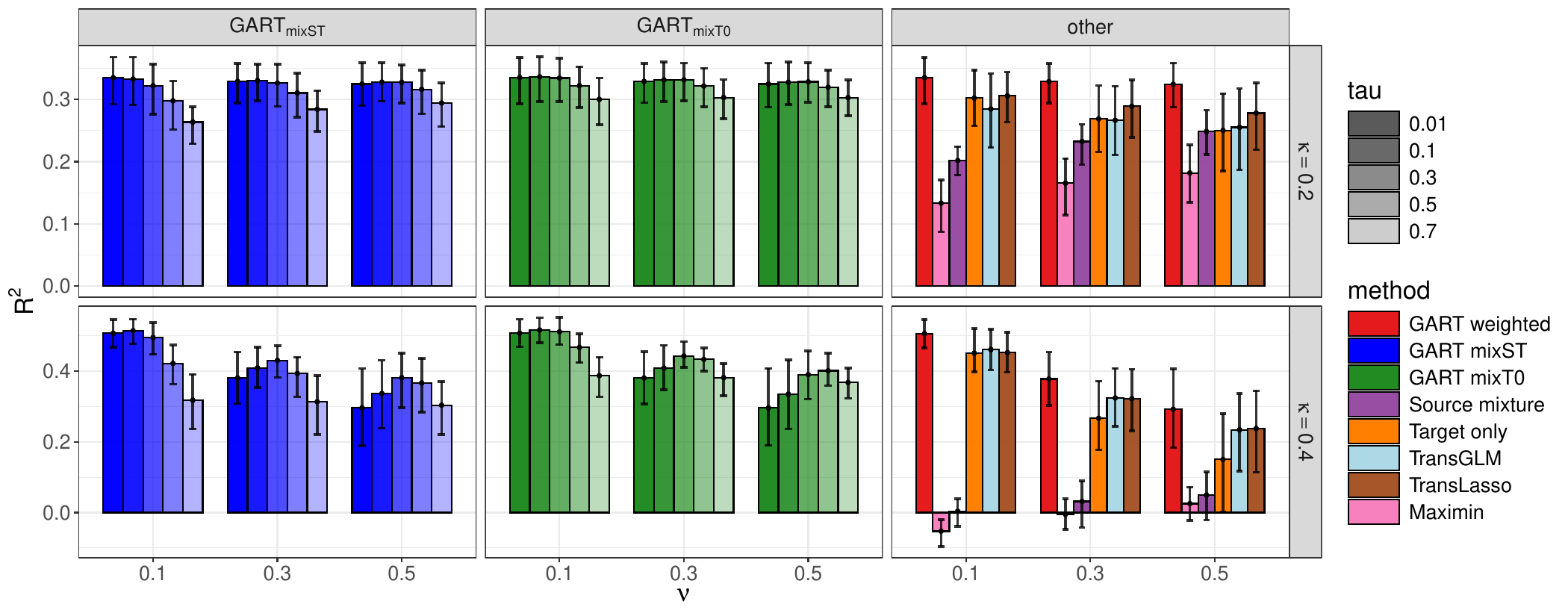}
    \caption{Empirical mean and 95\% confidence interval of $R^2$ of GART estimators and other benchmarks for settings 5 with $\kappa = 0.2$ and $\kappa = 0.4$ on different future target populations (indexed by $\nu$). GART is constructed using different $\tau$ and initial estimators: weighted, $\betahat\subinit\supmixST$ (mixST), $\betahat\subinit\supmixT0$ (mixT0).}
          \label{fig: simulation initial 5}
\end{figure}

Figure~\ref{fig: simulation initial 5} further evaluates the robustness of GART in setting 4.2, where the future target populations deviate from the observed target with increasing severity indexed by $\nu$, and $\beta^*$ lies moderately ($\kappa = 0.2$) or substantially ($\kappa = 0.4$) outside the source convex hull. When $\kappa = 0.2$, GART estimators across all initialization choices consistently outperform alternative methods, and their performance remains stable as $\nu$ increases. As $\kappa$ increases to $0.4$, GART initialized with $\betahat\subinit\supmixST$ or $\betahat\subinit\supmixT0$ exhibits stronger generalization to distant future populations, particularly when trained with larger $\tau$, at the cost of a slight loss when $\nu$ is small. While other methods display similar trends to those in setting 4, both Maximin and Source Mixture perform substantially worse than the GART estimators initialized with $\betahat\subinit\supmixST$ or $\betahat\subinit\supmixT0$, further highlighting the advantage of GART’s constraint-guided formulation when improving model generalizability.

\subsubsection{Impact of Initial Estimator}
\label{sec: appendix compare ini}
\begin{figure}[H]
    \centering
    \includegraphics[width=.8\linewidth, page=1]{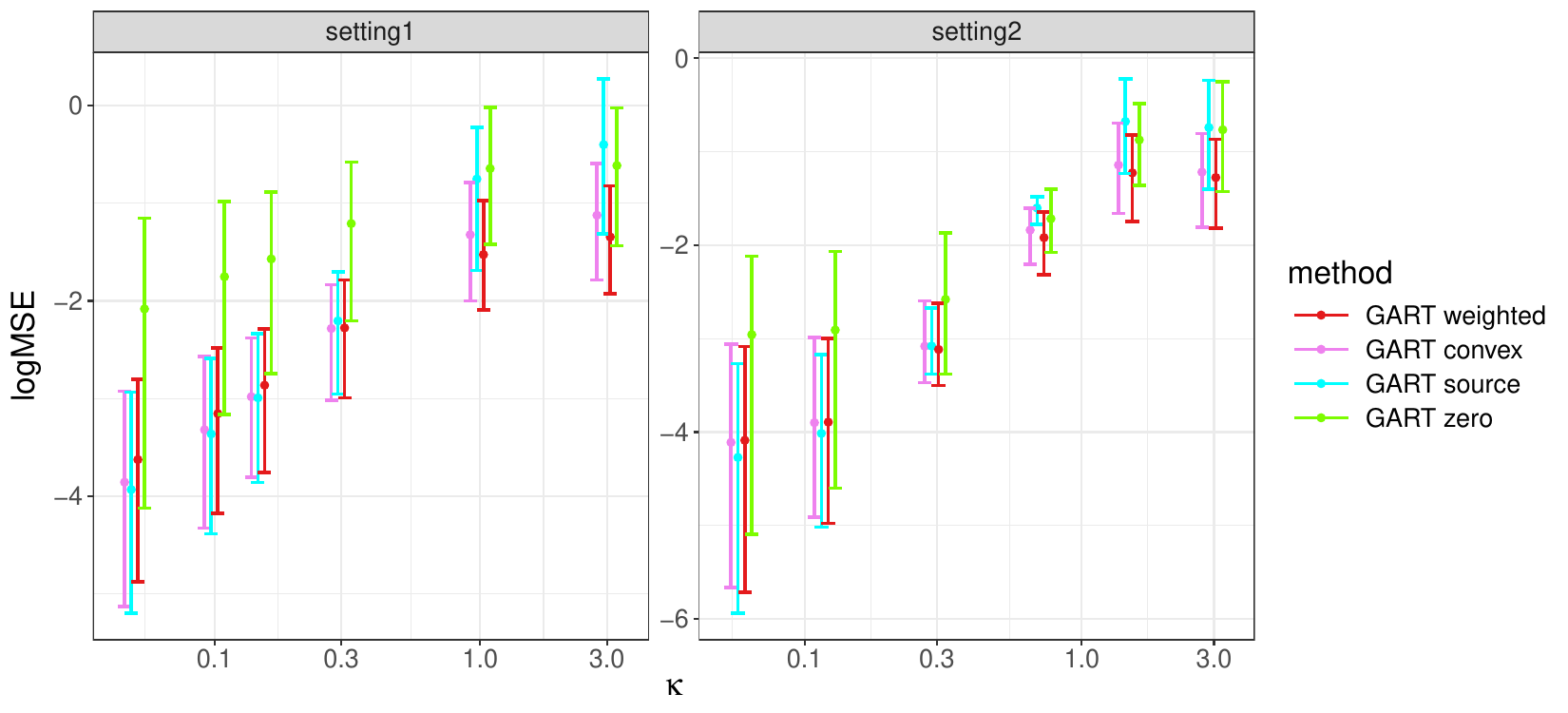}
    \caption{Empirical mean and 95\% confidence interval of $\log \text{MSE}$ of GART estimators constructed with the initial estimator chosen as weighted, convex combination, source combination, or zero, in settings 1 and 2. The prediction performance was assessed across a spectrum of observed target populations, indexed by $\kappa$, with the source mixing assumption progressively failing as $\kappa$ increases.}
    \label{fig: simulation in compare}
\end{figure}

We next summarize the impact of initial estimator on the GART performance under various settings. We first evaluate the performance of $\betahat\subGART$ on the observed target in settings 1 and 2 with varying $\kappa$, where the source mixing assumptions gradually fails as we increase $\kappa$. larger value of $\kappa$ ind. To optimize the performance on the observed target, we choose a small of $\tau = 0.01$ under which the initial estimators  $\betahat\subinit\supmixST$,  $\betahat\subinit\supmixT0$ and $\betahat\subinit\supweighted$ result in very similar performance. Thus we only present results in Figure \ref{fig: simulation in compare} for initial estimators chosen as $0$, source mixture $\betahat\subinit\supsource$ in \eqref{eq: def of source mix init beta}, $\betahat\subinit\supweighted$ in \eqref{eq: def of weighted init beta}, and $\betahat\subinit\supconvex$ as the convex combination of sources and the target with positive weights:
\begin{equation}
    \widehat{\beta}\subinit\supconvex = \left(\BBhat_{0\mathcal{A}_1}\widehat{\gamma}_{\mathcal{A}_1}\supconvex + \BBhat_{0\mathcal{A}_2}\widehat{\gamma}_{\mathcal{A}_2}\supconvex\right) / 2
    \label{eq: def of convex init beta}
\end{equation}
with $\widehat{\gamma}_{\mathcal{A}_j}\supconvex = \argmin_{\gamma\in\Delta_{L+1}} n\subajc^{-1}\|\YY\subajc\supQ - \XX\subajc\supQ\BBhat_{0\mathcal{A}_j}\gamma\|_2^2$. GART, when initialized with $\betahat\subinit\supweighted$, generally achieved the most robust performance, while GART initialized with $\betahat\subinit=0$ consistently performed the poorest. The source mixture initialization is effective only when the source mixing assumptions are approximately met, whereas the convex combination initialization is less adaptable across different settings compared to $\betahat\subinit\supweighted$.

\begin{figure}[H]
    \centering
\begin{subfigure}[b]{.9\linewidth}\centering
    \includegraphics[width=.9\linewidth, page=1]{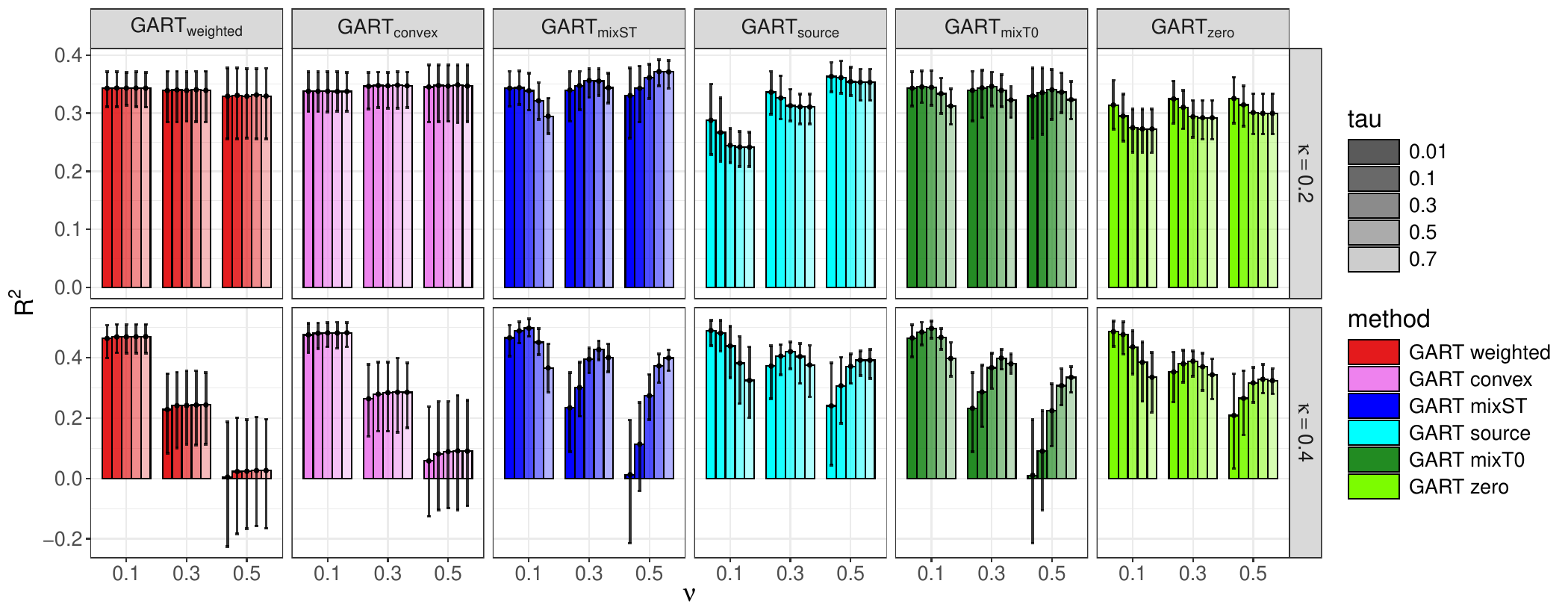}
    \caption{Setting 4.1}
    \end{subfigure}
\begin{subfigure}[b]{.9\linewidth}\centering
    \includegraphics[width=0.9\linewidth, page=2]{supp_figure/change_valid_diffbase_R2.pdf}
    \caption{Setting 4.2}
    \end{subfigure}
    \label{fig: simulation setting4.1 tau}
    \caption{Average $R^2$ of GART estimators for settings 4.1 and 4.2 with $\kappa = 0.2$ and $\kappa = 0.4$ on different future target populations (indexed by $\nu$) constructed using different $\tau$ and initial estimators: weighted, convex combination, $\betahat\subinit\supmixST$ (mixST), linear combination of source (source), $\betahat\subinit\supmixT0$ (mixT0), and zero. }
    \label{fig: simulation initial 4.1+4.2}
\end{figure}

Figure \ref{fig: simulation initial 4.1+4.2}  summarizes the performance of GART equipped with different initials and different choices of $\tau$ in setting 4.1 and 4.2, where we focus more on model generalizability. In both settings, when $\nu$ is small ($\le 0.1$) yielding to future target reasonably close to the observed target, the GART with weighted initial consistently attain high accuracy. As $\nu$ increases, GART initialized with $\betahat\subinit\supmixST$ or  $\betahat\subinit\supmixT0$ can attain more robust performance especially when $\kappa =0.4$, corresponding to the case when the observed target is also away from the source hull. GART with linear source can attain better performance in setting 4.1 when the future target can be well approximated by source mixing yet the observed target cannot. On the other hand, when the future target is also away from the source hull as in setting 4.2, GART with $\betahat\subinit\supmixT0$ can attain high generalizability by choosing a larger $\tau$. Thus, in practice, we recommend choosing a slightly larger $\tau$ and adopt $\betahat\subinit\supmixT0$ as the iitial when the goal is to generalize to a future target that is substantially different from the observed target.

\newpage
\section{Additional Real Data Analysis}
\label{sec: more real data analysis}
\begin{figure}[h]
    \centering
    \includegraphics[width=1\linewidth]{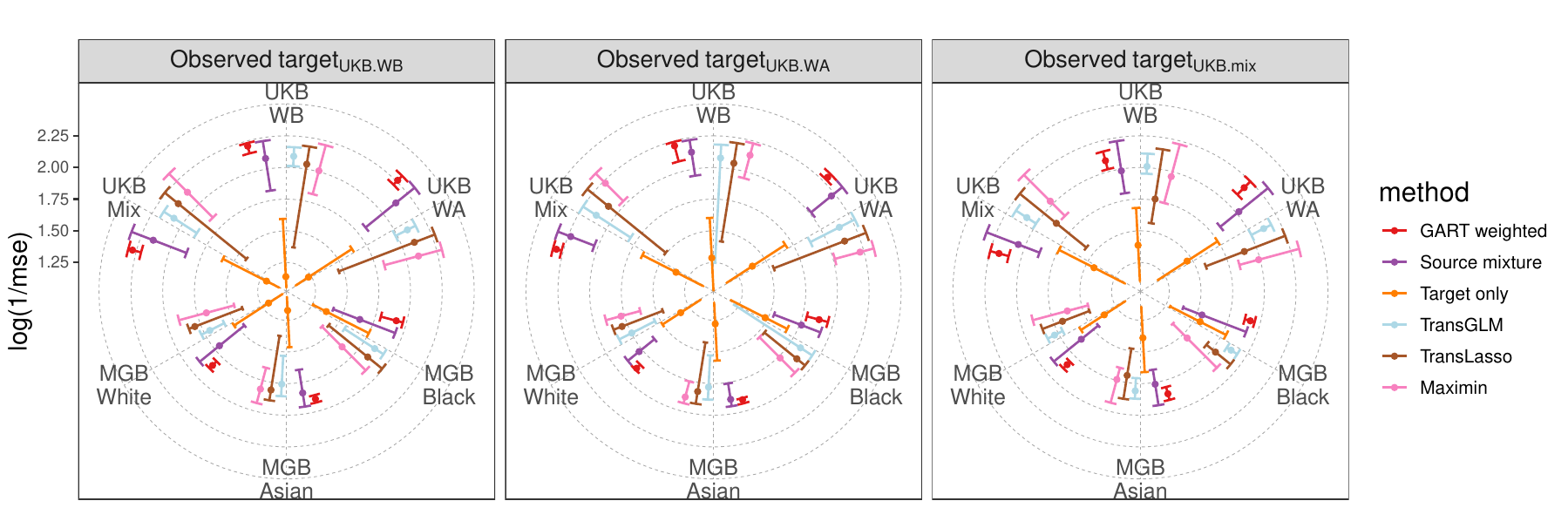}
    \caption{Prediction error $\log (1/\text{mse})$ on different validation populations when trained with UKB WB, UKB WA, or UKB WB being the target population and $n=50$.}
    \label{fig: radar 50}
\end{figure}

\end{document}